\documentclass[twoside,11pt]{article}

\usepackage[accepted]{melba}

\usepackage{amsmath,amsfonts}

\usepackage[toc,page]{appendix}
\usepackage{booktabs}
\usepackage{catchfilebetweentags}
\usepackage{caption}
\usepackage{commath}
\usepackage{float}
\usepackage{graphicx}
\usepackage[acronym]{glossaries-extra}
\usepackage{macros}
\usepackage{makecell}
\usepackage{mathabx}
\usepackage{mathtools}
\usepackage[super]{nth}
\usepackage[binary-units=true]{siunitx}
\usepackage{subcaption}
\usepackage{xfrac}
\usepackage{xspace}

\usepackage{cleveref}

\loadglsentries{glossary.sty}

\newcommand{\nochains}{two}
\newcommand{\nosamples}{\numprint{500}}
\newcommand{\nosamplesburnin}{\numprint{100000}}
\newcommand{\nosamplestotal}{one million}

\melbaheading{2021:016}{ https://www.melba-journal.org/papers/2021:016.html}{2021}{1-25}{01/2021}{10/2021}{Daniel Grzech, Mohammad Farid Azampour, Huaqi Qiu, Ben Glocker, Bernhard Kainz and Lo\"{i}c Le Folgoc}{Uncertainty for Safe Utilization of Machine Learning in Medical Imaging (UNSURE) 2020}{Christian Baumgartner, Adrian Dalca, Carole Sudre, Ryutaro Tanno, Sandy Wells}

\ShortHeadings{Uncertainty quantification in non-rigid image registration via SG-MCMC}{Grzech, Azampour, Qiu, Glocker, Kainz and Le Folgoc}
\firstpageno{1}

\title{Uncertainty quantification in non-rigid image registration\\via stochastic gradient Markov chain Monte Carlo}

\author{\name Daniel Grzech \email d.grzech17@imperial.ac.uk \\
        \addr Department of Computing, Imperial College London, London, UK
        \AND
	    \name Mohammad Farid Azampour \email f.azampour20@imperial.ac.uk \\
	    \addr Department of Computing, Imperial College London, London, UK \\
	    \addr Computer Aided Medical Procedures, Technische Universit\"at M\"unchen, Munich, Germany \\
	    \addr Department of Electrical Engineering, Sharif University of Technology, Tehran, Iran
	    \AND
	    \name Huaqi Qiu \email huaqi.qiu15@imperial.ac.uk \\
	    \addr Department of Comuting, Imperial College London, London, UK
	    \AND
    	\name Ben Glocker \email b.glocker@imperial.ac.uk \\
    	\addr Department of Computing, Imperial College London, London, UK
    	\AND
    	\name Bernhard Kainz \email b.kainz@imperial.ac.uk \\
    	\addr Department of Computing, Imperial College London, London, UK \\
    	\addr Friedrich-Alexander-Universit\"at Erlangen-N\"urnberg, Erlangen, Germany
    	\AND
    	\name Lo\"{i}c Le Folgoc \email l.le-folgoc@imperial.ac.uk \\
    	\addr Department of Computing, Imperial College London, London, UK
}

\begin{document}

% top matter
\maketitle

% abstract
\begin{abstract}%   <- trailing '%' for backward compatibility of .sty file
    We develop a new Bayesian model for non-rigid registration of three-dimensional medical images, with a focus on uncertainty quantification. Probabilistic registration of large images with calibrated uncertainty estimates is difficult for both computational and modelling reasons. To address the computational issues, we explore connections between the \emph{Markov chain Monte Carlo by backpropagation} and the \emph{variational inference by backpropagation} frameworks, in order to efficiently draw samples from the posterior distribution of transformation parameters. To address the modelling issues, we formulate a Bayesian model for image registration that overcomes the existing barriers when using a dense, high-dimensional, and diffeomorphic transformation parametrisation. This results in improved calibration of uncertainty estimates. We compare the model in terms of both image registration accuracy and uncertainty quantification to VoxelMorph, a state-of-the-art image registration model based on deep learning.
\end{abstract}

% keywords

%---------------------------------------------------------------
% Introduction (or first section)
\section{Introduction}
\label{sec:intro}

Image registration is the problem of aligning images into a common coordinate system such that the discrete pixel locations have the same semantic information. It is a common pre-processing step for many applications, e.g. the statistical analysis of imaging data and computer-aided diagnosis through comparison with an atlas. Image registration methods based on deep learning tend to incorporate task-specific knowledge from large datasets, whereas traditional methods are more general purpose. Many established models are based on the iterative optimisation of an energy function consisting of task-specific similarity and regularisation terms, which has to be done independently for every pair of images in order to calculate the deformation field \citep{Schnabel2001, Klein2009, Avants2014}.

DLIR \citep{DeVos2019} and VoxelMorph \citep{Balakrishnan2018, Balakrishnan2019, Dalca2018, Dalca2019} changed this paradigm by learning a function that maps a pair of input images to a deformation field. This gives a speed-up of several orders of magnitude at inference time and maintains an accuracy comparable to traditional methods. An overview of state-of-the-art models for image registration based on deep learning can be found in \cite{Lee2019}.

Due to the perceived conceptual difficulty and computational overhead, Bayesian methods tend to be shunned when designing medical image analysis algorithms. However, in order to fully explore the parameter space and lessen the impact of ad-hoc hyperparameter choices, it is desirable to use Bayesian models. In addition, with help of open-source libraries with automatic differentiation like PyTorch, the implementation of even complex Bayesian models for image registration is very similar to that of non-probabilistic models.

In this paper, we make use of the \gls{SG-MCMC} algorithm to design an efficient posterior sampling algorithm for 3D non-rigid image registration. \gls{SG-MCMC} is based on the idea of stochastic gradient descent interpreted as a stochastic process with a stationary distribution centred on the optimum and whose covariance can be used to approximate the posterior distribution \citep{Chen2016,Mandt2017}. \gls{SG-MCMC} methods have been useful for training generative models on very large datasets ubiquitous in computer vision, e.g. \cite{Du2019, Nijkamp2019, Zhang2020}. We show that they are also applicable to image registration.

%<*extension1>
This work is an extended version of \cite{Grzech2020}, where we first proposed use of the \gls{SG-MCMC} algorithm for non-rigid image registration. The code to reproduce the results is available in a public repository: \url{https://github.com/dgrzech/ir-sgmcmc}. %<*extension3>
The following is a summary of the main contributions of the previous work:
\begin{enumerate}
    \item We proposed a computationally efficient \gls{SG-MCMC} algorithm for three-dimensional diffeomorphic non-rigid image registration;
    \item We introduced a new regularisation loss, which allows to carry out inference of the regularisation strength when using a transformation parametrisation with a large number of degrees of freedom;
    \item We evaluated the model both qualitatively and quantitatively by analysing the output transformations, image registration accuracy, and uncertainty estimates on inter-subject brain \gls{MRI} data from the UK Biobank dataset.
\end{enumerate} %</extension3> %<*extension2>
In this version, we extend the previous work: \begin{itemize}
    \item[--] %<*extension4>
    We provide more details on the Bayesian formulation, including a comprehensive analysis of the learnable regularisation loss, as well as a more in-depth analysis of the model hyperparameters and hyperpriors; %</extension4>
    \item[--] %<*extension5>
    We conduct additional experiments in order to compare the uncertainty estimates output by \gls{VI}, \gls{SG-MCMC}, and VoxelMorph qualitatively, as well as quantitatively by analysing the Pearson correlation coefficient between the displacement and label uncertainties;
    \item[--] We analyse the differences between uncertainty estimates when the \gls{SG-MCMC} algorithm is initialised to different transformations and when using different parametrisations of the transformation, including non-parametric \glspl{SVF} and \glspl{SVF} based on B-splines; and %</extension5>
    \item[--] We include a detailed evaluation of the computational speed and of the output transformation smoothness.
\end{itemize}
%</extension2>
%</extension1>

%---------------------------------------------------------------
\section{Related work}
\label{sec:related-work}

%<*related-work>
The problem of uncertainty quantification in non-rigid image registration is controversial because of ambiguity regarding the definition of uncertainty as well as the accuracy of uncertainty estimates \citep{Luo2019}. Uncertainty quantification in probabilistic image registration relies either on variational Bayesian methods \citep{Simpson2012, Simpson2013, Wassermann2014}, which are fast and approximate, and popular within models based on deep learning \citep{Dalca2018, Liu2019, Schultz2019}, or \gls{MCMC} methods, which are slower but enable asymptotically exact sampling from the posterior distribution of the transformation parameters. The latter include e.g. Metropolis-Hastings used for intra-subject registration of brain \gls{MRI} scans \citep{Risholm2010, Risholm2013} and estimating delivery dose in radiotherapy \citep{Risholm2011}, reversible-jump \gls{MCMC} used for cardiac \gls{MRI} \citep{LeFolgoc2017}, and Hamiltonian Monte Carlo used for atlas building \citep{Zhang2013}.
\par
Uncertainty quantification for image registration has also been done via kernel regression \citep{Zollei2007, Firdaus2012} and deep learning \citep{Dalca2018, Krebs2019, Heinrich2019, Sedghi2019}. More generally, Bayesian frameworks have been used e.g. to characterize image intensities \citep{Hachama2012} and anatomic variability \citep{Zhang2014}.
\par
One of the main obstacles to a more widespread use of \gls{MCMC} methods for uncertainty quantification is the computational cost. This was recently tackled by embedding \gls{MCMC} in a multilevel framework \citep{Schultz2018}. \gls{SG-MCMC} was previously used for \emph{rigid} image registration \citep{Karabulut2017}. It has also been employed in the context of unsupervised non-rigid image registration based on deep learning, where it allowed to sample from the posterior distribution of the network weights, rather than directly the transformation parameters \citep{Khawaled2020}.
\par
Previous work on data-driven regularisation focuses on transformation parametrisations with a relatively low number of degrees of freedom, e.g. B-splines \citep{Simpson2012} and a sparse parametrisation based on Gaussian \glspl{RBF} \citep{LeFolgoc2017}. Limited work exists also on spatially-varying regularisation, again with B-splines \citep{Simpson2015}. Deep learning has been used for spatially-varying regularisation learnt using more than one image pair \citep{Niethammer2019}. \cite{Shen2019} introduced a related model which could be used for learning regularisation strength based on a single image pair but suffered from non-diffeomorphic output transformations and slow speed.
%</related-work>

%---------------------------------------------------------------
% A methodological, model, or similar section often comes here.
\section{Registration model}

We denote an image pair by $\dataset = (F, M)$, where $F \colon \Omega_F \to \mathbb{R}$ and $M \colon \Omega_M \to \mathbb{R}$ are a fixed and a moving image respectively. The goal of image registration is to align the underlying domains $\Omega_F$ and $\Omega_M$ with a transformation $\varphi \left(w \right) \colon \Omega_F \to \Omega_M$, i.e. to calculate parameters $w$ such that $F \simeq M(w) \coloneqq M \circ \varphi^{-1} (w)$. The transformation is often expected to possess desirable properties, e.g. diffeomorphic transformations are smooth and invertible, with a smooth inverse.

We parametrise the transformation using the \gls{SVF} formulation \citep{Arsigny2006, Ashburner2007}, which we briefly review below. The \gls{ODE} that defines the evolution of the transformation is given by:
\begin{equation}
    \fracpartial{\varphi^{(t)}}{t} = w \left( \varphi^{(t)} \right)
    \label{eq:SVF}
\end{equation}
where $\varphi^{(0)}$ is the identity transformation and $t \in [ 0, 1 ]$. If the velocity field $w$ is spatially smooth, then the solution to \Cref{eq:SVF} is a diffeomorphic transformation. Numerical integration is done by scaling and squaring, which uses the following recurrence relation with $2^T$ steps \citep{Arsigny2006}:
\begin{equation}
    \varphi^{(1 / {2}^{t-1})} = \varphi^{(1 / {2}^t)} \circ \varphi^{(1 / {2}^t)}
    \label{eq:scaling-and-squaring}
\end{equation}

The Bayesian image registration framework that we present is not limited to \glspl{SVF}. Moreover, there is a very limited amount of research on the impact of the transformation parametrisation on uncertainty quantification. Previous work on uncertainty quantification in image registration characterised uncertainty using a single transformation parametrisation, e.g. a small deformation model using B-splines in \cite{Simpson2012}, the \gls{FE} method in \cite{Risholm2013}, and multi-scale Gaussian \glspl{RBF} in \cite{LeFolgoc2017}, or a \gls{LDDMM} in \cite{Wassermann2014}.

To help understand the potential impact of the transformation parametrisation on uncertainty quantification, we also implement \glspl{SVF} based on cubic B-splines \citep{Modat2012}. In this case, the \gls{SVF} consists of a grid of B-spline control points, with regular spacing $\delta \geq 1$ voxel. The dense \gls{SVF} at each point is a weighted combination of cubic B-spline basis functions \citep{Rueckert1999}. To calculate the transformation based on the dense velocity field, we again use the scaling and squaring algorithm in \Cref{eq:scaling-and-squaring}.

%---------------------------------------------------------------
\subsection{Likelihood model}
\label{sec:likelihood}

The likelihood $p \left( \dataset \mid w \right)$ specifies the relationship between the data and the transformation parameters by means of a similarity metric. In probabilistic image registration, it usually takes the form of a Boltzmann distribution \citep{Ashburner2007}:
\begin{equation}
    \log p \left( \dataset \mid w; \mathcal{H} \right) \varpropto -\mathcal{E}_{\text{data}} \left( \mathcal{D}, w; \mathcal{H} \right)
\end{equation}
where $\mathcal{E}_{\text{data}}$ is the similarity metric and $\mathcal{H}$ an optional set of hyperparameters.

%<*similarity-metric>
\Gls{LCC}, which is invariant to linear intensity scaling, is a popular similarity metric but not meaningful in a probabilistic context. For this reason, instead of the sum of the voxel-wise product of intensities, like in standard \gls{LCC}, we opt for the sum of voxel-wise squared differences of images standardised to zero mean and unit variance inside a local neighbourhood of five voxels. This way, we can benefit from robustness under linear intensity transformations, as well as desirable properties of a \gls{GMM} of intensity residuals, i.e. robustness to outlier values caused by acquisition artefacts and misalignment over the course of registration \citep{LeFolgoc2017}. %</similarity-metric>

Let $\overline{F}$ and $\overline{M(w)}$ be respectively the fixed and the warped moving image with intensities standardised to zero mean and unit variance inside a neighbourhood of five voxels. For each voxel, the intensity residual $r_i = \overline{F} - \overline{M(w)}$, $i \in \{1, \dots, N^3\}$, is assigned to the $l$-th component of the mixture, $1 \leq l \leq L$, if the categorical variable $c_i \in \{ 1, \dots, L\}$ is equal to $l$, in which case it follows a normal distribution $\mathcal{N} \left( 0, \beta_l^{-1} \right)$\footnote{In order to reduce the notation clutter we omitted the voxel index for the fixed and moving images.}. The component assignment $c_i$ follows a categorical distribution and takes value $l$ with probability $\varrho_l$. %<*GMM-neighbourhood>
We use the same \gls{GMM} of intensity residuals on a global basis rather than per neighbourhood. %</GMM-neighbourhood>
In all experiments it has $L=4$ components, which we determine to be sufficient for a good model fit.

We also use the scalar virtual decimation factor $\alpha$ to account for the fact that voxel-wise residuals are not independent. This prevents over-emphasis on the data term and allows to better calibrate uncertainty estimates \citep{Groves2011,Simpson2012}. The full expression of the image similarity term is given by:
\begin{equation}
    \mathcal{E}_{\text{data}} \left( \mathcal{D}, w; \beta, \varrho \right) = -\alpha \times \sum \limits_{i=1}^{N^3} \log \sum \limits_{l=1}^L \varrho_l \sqrt{\frac{\beta_l}{2 \pi}} \exp \left( -\frac{\beta_l}{2} r_i^2 \right)
\end{equation}

%---------------------------------------------------------------
\subsection{Transformation priors}
\label{sec:reg}

In Bayesian models, the transformation parameters are typically regularised with use of a multivariate normal prior that ensures smoothness:
\begin{equation}
    \log p\left( w; \lambda_{\text{reg}} \right) \varpropto -\frac{1}{2} \lambda_{\text{reg}} \left( \mathrm{L}w \right)^\intercal \mathrm{L}w
    \label{eq:L2_reg}
\end{equation}
where $\lambda_{\text{reg}}$ is a scalar parameter that controls the regularisation strength, and $\mathrm{L}$ is the matrix of a differential operator. Here we assume that $\mathrm{L}$ represents the gradient operator, which penalises the magnitude of the \nth{1} derivative of a velocity field. Note that $\left( \mathrm{L}w \right)^\intercal \mathrm{L}w = \Vert \mathrm{L}w \Vert^2 \coloneqq \chi^2$.

The regularisation weight $\lambda_{\text{reg}}$ can either be fixed or estimated from data. The latter has been done successfully only for transformation parametrisations with a relatively low number of degrees of freedom, e.g. B-splines \citep{Simpson2012} and a sparse parametrisation \citep{LeFolgoc2017}. In case of an \gls{SVF}, where the number of degrees of freedom is orders of magnitude higher, the problem is more difficult. However, a reliable method to adjust regularisation strength based on data is crucial, as both the output transformation and registration uncertainty are highly sensitive to regularisation. %<*regularisation>
In order to infer the regularisation strength, we specify a log-normal prior on the scalar regularisation energy $\chi^2 \sim \text{Lognormal} \left( \mu_{\chi^2}, \sigma_{\chi^2}^2 \right)$, and derive a prior on the underlying \gls{SVF}:
\begin{align}
    \log p\left( \chi^2 \right) &\varpropto -\log \chi^2 - \log \sigma_{\chi^2} - \frac{\left( \log \chi^2 - \mu_{\chi^2} \right)^2}{2 \sigma_{\chi^2}^2} \\
    \log p(w) &\varpropto -\left( \frac{\nu}{2} - 1 \right) \log \chi^2 + \log p \left(\chi^2 \right)
\end{align}
where $\nu = 3 N^3$ is the number of degrees of freedom, i.e. the count of transformation parameters in all three directions. Given (semi-)informative hyperpriors on $\mu_{\chi^2}$ and $\sigma^2_{\chi^2}$, which we discuss in the next section, we can adjust the regularisation strength to the input images. The full expression of the regularisation term is given by: \begin{equation}
    \mathcal{E}_{\text{reg}} \left( w \right) = \frac{\nu}{2} \log \chi^2 + \log \sigma_{\chi^2} + \frac{\left( \log \chi^2 - \mu_{\chi^2} \right)^2}{2 \sigma_{\chi^2}^2}
\end{equation}

It is worth noting that the traditional $L_2$ regularisation with a fixed regularisation weight in \Cref{eq:L2_reg} actually belongs to this family of regularisation losses. If we specify a gamma prior instead of a log-normal prior on the scalar regularisation energy $\chi^2 \sim \Gamma \left( \sfrac{\nu}{2}, \sfrac{\lambda_{\text{reg}}}{2} \right)$, we get:
\begin{alignat}{3}
    \log p \left( \chi^2 \right) &\varpropto &&\left( \frac{\nu}{2} -1 \right) \log \chi^2 &&- \frac{1}{2} \lambda_{\text{reg}} \cdot \chi^2 \\
    \log p\left( w \right) &\varpropto &&\left( \frac{\nu}{2} - 1 \right) \log \chi^2 &&+ \left( \frac{\nu}{2} - 1 \right) \log \chi^2 - \frac{1}{2} \lambda_{\text{reg}} \cdot \chi^2 \\
    &\varpropto &&-\frac{1}{2} \lambda_{\text{reg}} (\mathrm{L}w)^\intercal \mathrm{L}w &&
\end{alignat}%</regularisation>

%---------------------------------------------------------------
\subsection{Hyperpriors}
\label{sec:hyperpriors}

We set the likelihood \gls{GMM} hyperpriors similarly to \cite{LeFolgoc2017}, with the mixture precision parameters $\beta = \left( \beta_1, \dots, \beta_L \right)$ assigned independent log-normal priors $\beta_l \sim \text{Lognormal} \left( \mu_{\beta_l}, \sigma^2_{\beta_l} \right)$ and the mixture proportions $\varrho = \left( \varrho_1, \dots, \varrho_L \right)$ with an uninformative Dirichlet prior $\varrho \sim \text{Dir} \left( \kappa \right)$, where $\kappa = \left( \kappa_1, \dots, \kappa_L \right)$.

%<*regularisation-hyperpriors>
Regularisation parameters require informative priors due to the difficulty of learning the regularisation strength based on a single image pair. Because of a gamma prior on the regularisation energy $\exp \left( \mu_{\chi^2} \right) \sim \Gamma \left( \sfrac{\nu}{2}, \sfrac{\lambda_{\text{init}}}{2} \right)$, we can rely on the familiar regularisation weight $\lambda_{\text{init}}$ to initialise the logarithm of the regularisation energy $\mu_{\chi^2}$ to the expected value of the logarithm of the gamma distribution, i.e. $\mathbb{E} \left[ \mu_{\chi^2} \right] = \psi \left( \sfrac{\nu}{2} \right) - \log \left( \sfrac{\lambda_{\text{init}}}{2} \right)$, where $\psi$ is the digamma function. The value of this expression is sharply peaked if the number of degrees of freedom $\nu$ is large, which yields a very informative prior on $\mu_{\chi^2}$. More details on how to calculate the expected value of the logarithm of a gamma distribution can be found in \Cref{app:reg}.
\par
The choice of a hyperprior on the scale parameter $\sigma_{\chi^2}$, which controls the amount of deviation of $\log \chi^2$ from the location parameter $\mu_{\chi^2}$, is more intuitive. Here we use a log-normal prior $\sigma_{\chi^2}^2 \sim \text{Lognormal} \left( \eta, \varsigma^2 \right)$.
%</regularisation-hyperpriors>

%---------------------------------------------------------------
\section{Variational inference}
\label{sec:VI}

Image registration methods often rely on \gls{VI} for uncertainty quantification. We only use \gls{VI} to initialise the \gls{SG-MCMC} algorithm, which also lets us compare the uncertainty estimates output by approximate and asymptotically exact methods for sampling from the posterior distribution of transformation parameters.

We assume that the posterior distribution $p \left( w \mid \dataset \right)$ is a multivariate normal distribution $q \left( w \right) \sim \mathcal{N} \left(\mu_w, \Sigma_w\right)$. To find parameters $\mu_w$ and $\Sigma_w$, we maximise the \gls{ELBO} \citep{Jordan1999}:
\begin{equation}
    \mathcal{L}(q) = \mathbb{E}_q \left[ \log p( \dataset \mid w) \right] - \kld{q}{p} = -\big \langle \mathcal{E}_{\text{data}} + \mathcal{E}_{\text{reg}} \big \rangle_q + H(q)
    \label{eq:ELBO}
\end{equation}
where $\kld{q}{p}$ is the \gls{KL-divergence} between the approximate posterior $q$ and the prior $p$. Like in traditional image registration, the energy function consists of the sum of similarity and regularisation terms, with an additional term for the entropy of the posterior distribution $H(q)$. We show how to calculate this term in \Cref{app:KL}.

It is not possible to calculate every element of the covariance matrix $\Sigma_w$ due to high dimensionality of the problem. Instead, we approximate the covariance matrix as a sum of diagonal and low-rank parts, i.e. $\Sigma_w \approx \text{diag} \left( \sigma_w^2 \right) + u_w u_w^\intercal$, with $\sigma_w \in \mathbb{R}^{3N^3 \times 1}$ and $u_w \in \mathbb{R}^{3N^3 \times R}$, where $R$ is a hyperparameter which determines the parametrisation rank. Using a multivariate normal distribution as the approximate posterior distribution of transformation parameters is common in probabilistic image registration. The key difference in our work is the diagonal + low-rank parametrisation of the covariance matrix. Most recent image registration models with an \gls{SVF} transformation parametrisation are based on deep learning and make the assumption of a diagonal covariance matrix \citep{Dalca2018, Krebs2019}.

We use the reparametrisation trick with two samples per update to backpropagate with respect to the variational posterior parameters:
\begin{align}
    w         &= \mu_{w} \pm \left( \text{diag} \left( \sigma_w \right) \cdot \epsilon + u_{w}  \cdot x\right) \\ 
    \epsilon    &\sim \mathcal{N} \left( 0, I_{3N^3} \right),\ x \sim \mathcal{N} \left( 0, I_R \right) \nonumber
\end{align}

In order to make the optimisation less susceptible to undesired local maxima of the \gls{ELBO}, we take advantage of Sobolev gradients \citep{Neuberger1997}. Samples from the posterior are convolved with a Sobolev kernel. We approximate the 3D kernel by three separable 1D kernels to lower the computational overhead. Using the notation in \cite{Slavcheva2018a}, we set the kernel width to $s_{H^1}=7$ and the smoothing parameter to $\lambda_{H^1} = 0.5$.

%<*SAEM>
The \gls{GMM} and regularisation hyperparameters are fit using the \gls{SAEM} algorithm \citep{Richard2009, Zhang2013}. The mixture precision hyperparameters $\beta$ and proportion hyperparameters $\varrho$ are updated by solving the optimisation problem:
\begin{equation}
    \beta^{(k)}, \varrho^{(k)} = \argmax_{\beta, \varrho} \mathbb{E}_q \left[ \log p \left( \dataset, w; \beta, \varrho \right) \mid \beta^{(k-1)}, \varrho^{(k-1)} \right] + \log p \left( \beta \right) + \log p \left( \varrho \right)
    \label{eq:SAEM}
\end{equation}

This is done at each step of the iterative optimisation algorithm. We update the regularisation hyperparameters in an analogous way. Even though the hybrid \gls{VI} and \gls{SAEM} approach is computationally efficient and requires minimal implementation effort, it disregards the uncertainty caused by hyperparameter variability.
%</SAEM>

%---------------------------------------------------------------
\section{Stochastic gradient Markov chain Monte Carlo}
\label{sec:SG-MCMC}

%<*SGLD-init1>
Image registration algorithms based on \gls{VI} restrict parametrisation of the posterior distribution to a specific family of probability distributions, which may not include the true posterior. To avoid this problem and sample the transformation parameters in an efficient way, we use \gls{SGLD} \citep{Besag1993,Welling2011}.  %</SGLD-init1>
The update equation is given by:
\begin{equation}
    w_{k+1} \leftarrow w_k + \tau A \nabla \log \pi \left( w_k \right) + \sqrt{2 \tau A} \xi_k
    \label{eq:update}
\end{equation}
where $\tau$ is the step size, %<*preconditioning-matrix>
$A$ is an optional preconditioning matrix, %</preconditioning-matrix>
$\nabla \log \pi \left( w_k \right)$ is the gradient of the logarithm of the posterior \gls{PDF}, and $\xi_k \sim \mathcal{N} \left( 0, I_{3N^3} \right)$. %<*SGLD-init2>
\gls{SGLD} does not require a particular initialisation, so %<*preconditioning-matrix2>
we study several different possibilities, including a sample $w_0 \sim \mathcal{N} \left( \mu_w, \Sigma_w \right)$ from the approximate variational posterior, in which case we set $A = \text{diag} \left( \sigma^2_w \right)$. %</SGLD-init2>
The preconditioning helps with the \gls{MCMC} mixing time in case the target distribution is strongly anisotropic. %</preconditioning-matrix2>

It is worth noting that, except for the preconditioning matrix $A$ and the noise term $\xi_k$, \Cref{eq:update} is equivalent to a gradient descent update when minimising the \gls{MAP} objective function $- \log p \left( w \mid \dataset \right) = -\log p \left( \dataset \mid w \right) -\log p \left(w \right)$. When drawing samples from \gls{SGLD}, we continue to update the \gls{GMM} as well as regularisation hyperparameters like in \Cref{eq:SAEM}, except that the expected value is calculated with respect to the new posterior $\pi \left( w \right)$.

In the limit as $k \to \infty$ and $\tau \to 0$, \gls{SGLD} can be used to draw exact samples from the posterior of the transformation parameters without Metropolis-Hastings accept-reject tests, which are computationally expensive. Indeed, these costs prevent the use of other \gls{MCMC} algorithms for the registration of large 3D images. In practice, the step size needs to be adjusted to avoid high autocorrelation between samples yet remain smaller than the width of the most constrained direction in the local energy landscape \citep{Neal2011}. The step size can also be used to control the trade-off between accuracy and computation time. We can quantify uncertainty either quickly in a coarse manner or slowly, with more detail. 

Despite the fact that the term $\nabla \log \pi \left( w \right)$ allows to traverse the energy landscape in an efficient way, \gls{SGLD} suffers from high autocorrelation and slow mixing between modes \citep{Hill2020}. However, simplicity of the formulation makes it better suited than other \gls{MCMC} methods for high-dimensional problems like three-dimensional image registration.

%---------------------------------------------------------------
\section{Experiments}

\subsection{Setup}
\label{sec:setup}

The model is implemented in PyTorch. For all experiments we use three-dimensional T2-FLAIR \gls{MRI} brain scans and subcortical structure segmentations from the UK Biobank dataset \citep{Sudlow2015}. Input images are pre-registered with the affine component of \textit{drop2} \citep{Glocker2008}\footnote{\url{https://github.com/biomedia-mira/drop2}} and resampled to $N = 128$ isotropic voxels of length \SI{1.82}{\mm} along every dimension.

We use $2^{12}$ steps to integrate \glspl{SVF}. In order to start optimisation with small displacements, $\mu_w$ is initialised to zero, which corresponds to an identity transformation, $\sigma_w$ to half a voxel length in each direction and $u_w$ to a tenth of a voxel length in each direction. We are mainly interested in the approximate variational posterior in order to initialise the \gls{SG-MCMC} algorithm, so the rank parameter is set to $R = 1$. We use the Adam optimiser with a step size of $1 \times 10^{-2}$ for the approximate variational posterior parameters $\mu_w$, $\log \sigma_w^2$, and $u_w$. Training is run until the \gls{ELBO} value stops to increase, which requires approximately \numprint{1024} iterations.

%<*hyperpriors-params-values>
In the likelihood model, we use $\kappa = 0.5$ for an uninformative Jeffreys prior on the mixture proportions, while the mixture precision hyperparameters are set to $\mu_{\beta_l} = 0.0$ and $\sigma_{\beta_l} = 2.3$.
%<*diffeomorphisms>
The model is much less sensitive to the value of the likelihood hyperparameters than the regularisation hyperparameters, which are calibrated to guarantee diffeomorphic transformations sampled from the approximate variational posterior $q \left( w \right) \sim \mathcal{N} \left( \mu_w, \Sigma_w \right)$. The local transformation is diffeomorphic only in locations where the Jacobian determinant is positive \citep{Ashburner2007}, so we aim to keep the number of voxels where the Jacobian determinant is non-positive $\vert \det J_{\varphi^{-1}} \vert \leq 0$ close to zero. %</diffeomorphisms>
We calibrate the location hyperparameter $\lambda_{\text{init}}$ in every experiment, while the scale hyperparameters are set to $\eta = 2.8$ and $\varsigma = 5.0$.
%</hyperpriors-params-values>

%<*SAEM-convergence>
\gls{SAEM} convergence is known to be conditional on decreasing step sizes \citep{Delyon1999}. For this reason, we use a small step size decay of $1 \times 10^{-3}$ and the Adam optimiser with a step size of $2 \times 10^{-1}$ for the GMM hyperparameters $\log \beta^{-0.5}$ and $\log \varrho$, and $1 \times 10^{-2}$ for the regularisation hyperparameters $\mu_{\chi^2}$ and $\log \sigma_{\chi^2}$. In case of parameters whose value is constrained to be positive, we state the step size used on the logarithms. In practice, we did not observe the result to be dependent on these step sizes.
%</SAEM-convergence>

%---------------------------------------------------------------
\subsection{Regularisation strength}
\label{sec:exp12}

%<*regularisation-comparison>
First we evaluate the proposed regularisation. We compare it to a gamma prior on $\lambda$, i.e. $\lambda \sim \Gamma \left( s, r \right)$, where $s$ and $r$ are the shape and the rate parameters respectively, set to uninformative values $s = r = \sfrac{\nu}{2}$ \citep{Simpson2012}. %</regularisation-comparison>

\begin{figure}[!htb]
    \centering
    \minipage[c]{0.25\linewidth}
        \caption*{\centering\footnotesize Fixed image $F$ and mean displacement}
    \endminipage
    \hfill
    \minipage[c]{0.65\linewidth}
        \begin{subfigure}{0.22\textwidth}
            \centering
            \includegraphics[trim={17cm 1.25cm 17cm 1.25cm},clip,width=\textwidth]{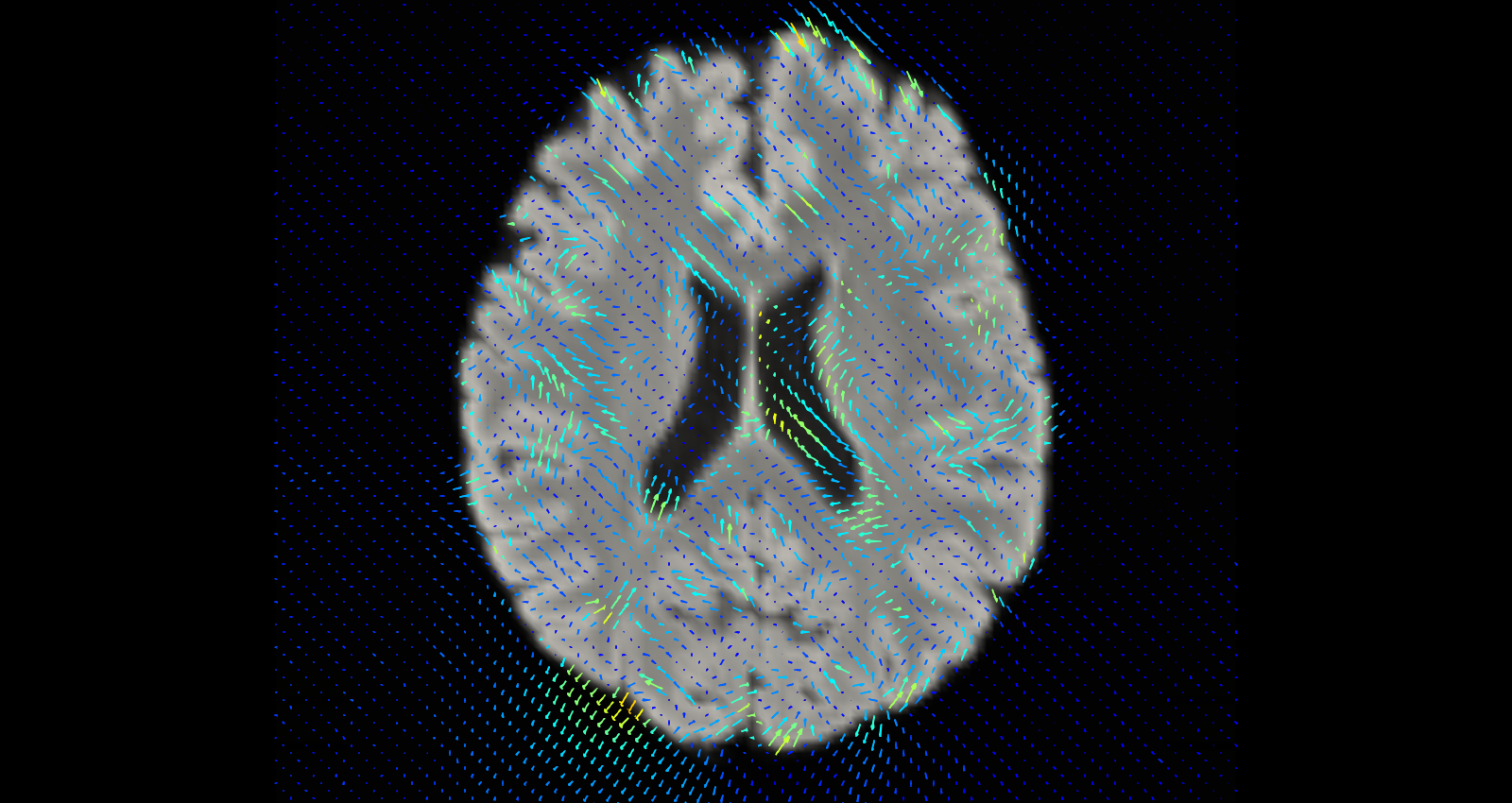}
        \end{subfigure}
        \begin{subfigure}{0.22\textwidth}
            \centering
            \includegraphics[trim={17cm 1.25cm 17cm 1.25cm},clip,width=\textwidth]{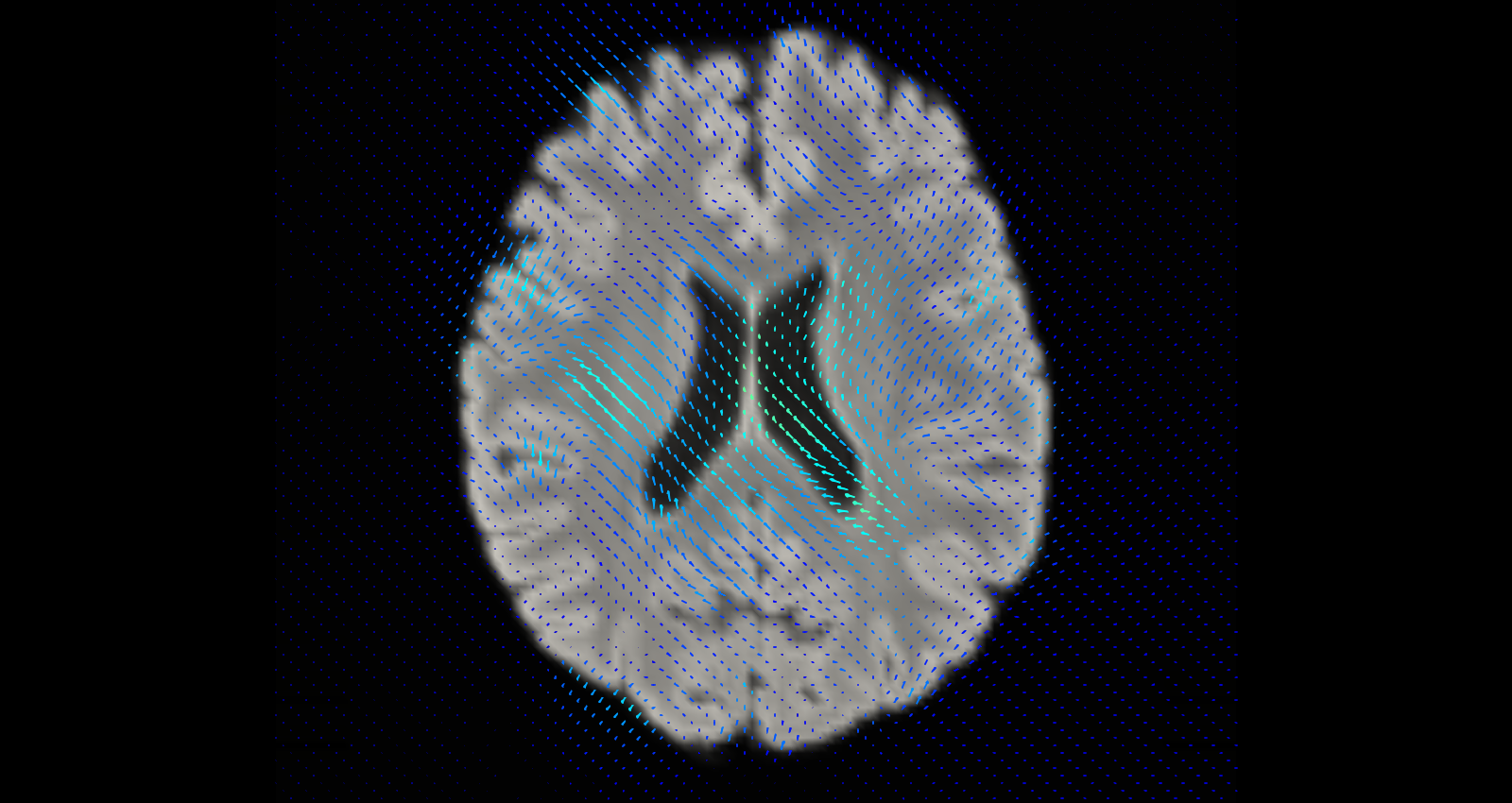}
        \end{subfigure}
        \begin{subfigure}{0.22\textwidth}
            \centering
            \includegraphics[trim={17cm 1.25cm 17cm 1.25cm},clip,width=\textwidth]{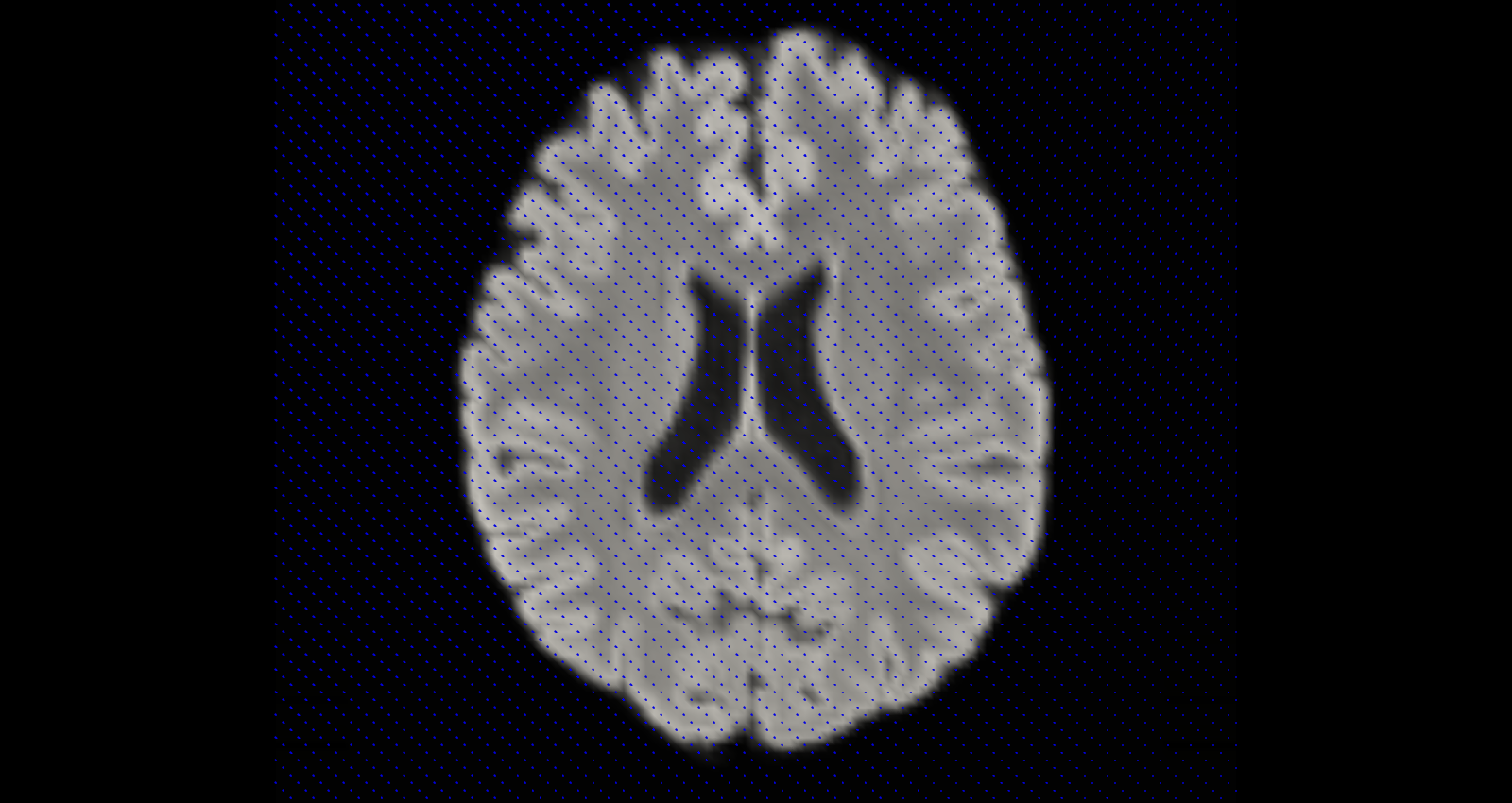}
        \end{subfigure}
        \begin{subfigure}{0.22\textwidth}
            \centering
            \includegraphics[trim={17cm 1.25cm 17cm 1.25cm},clip,width=\textwidth]{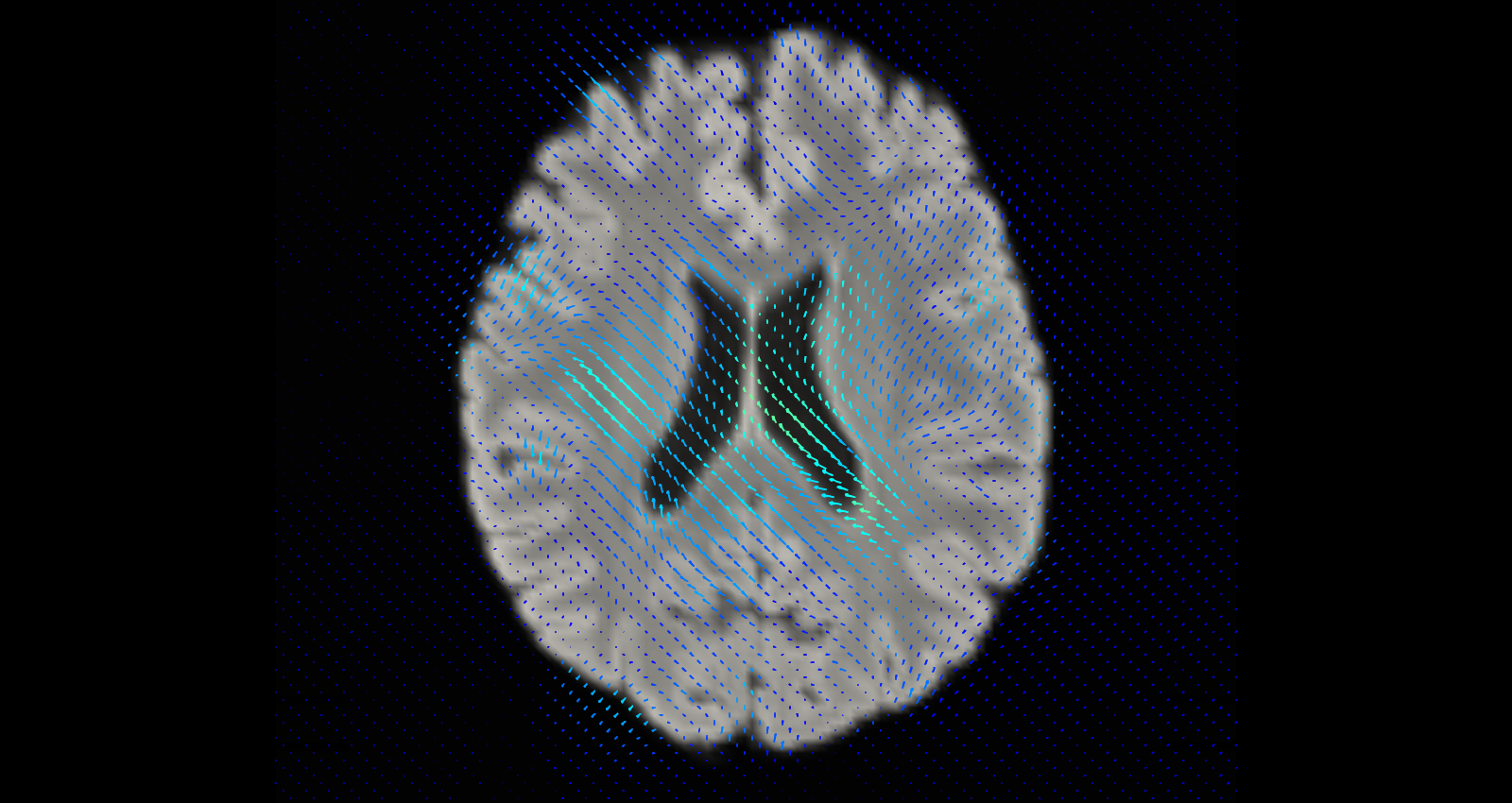}
        \end{subfigure}
        \begin{subfigure}{0.08\textwidth}
            \centering
            \includegraphics[trim={36cm 5cm 12cm 5cm},clip,height=2.5cm]{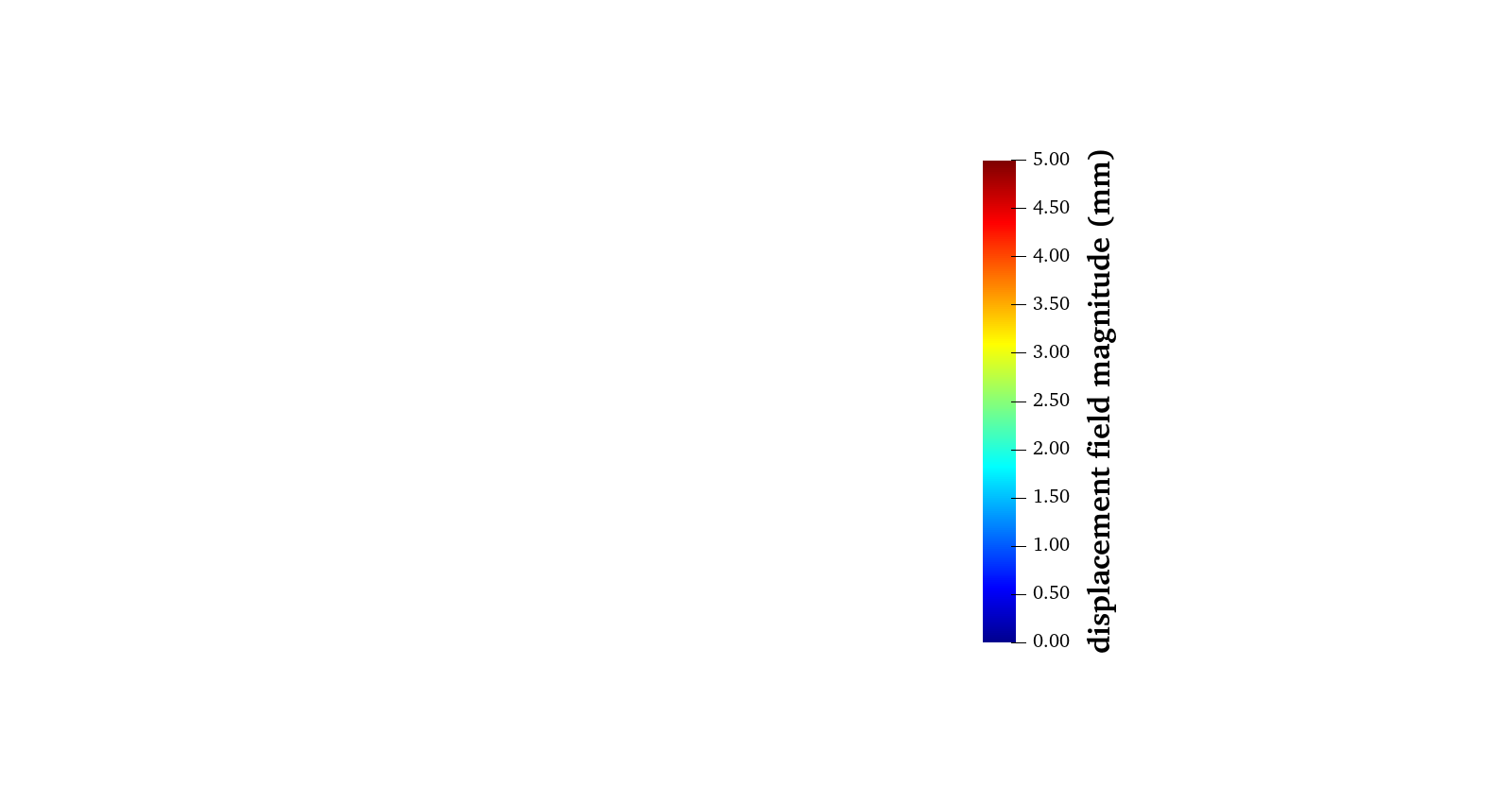}
        \end{subfigure}
    \endminipage
    
    \minipage[c]{0.25\linewidth}
        \caption*{\centering \footnotesize Warped moving image $M\left( \mu_w \right)$}
    \endminipage
    \hfill
    \minipage[c]{0.65\linewidth}
        \begin{subfigure}{0.22\textwidth}
            \centering
            \includegraphics[trim={17cm 1.25cm 17cm 1.25cm},clip,width=\textwidth]{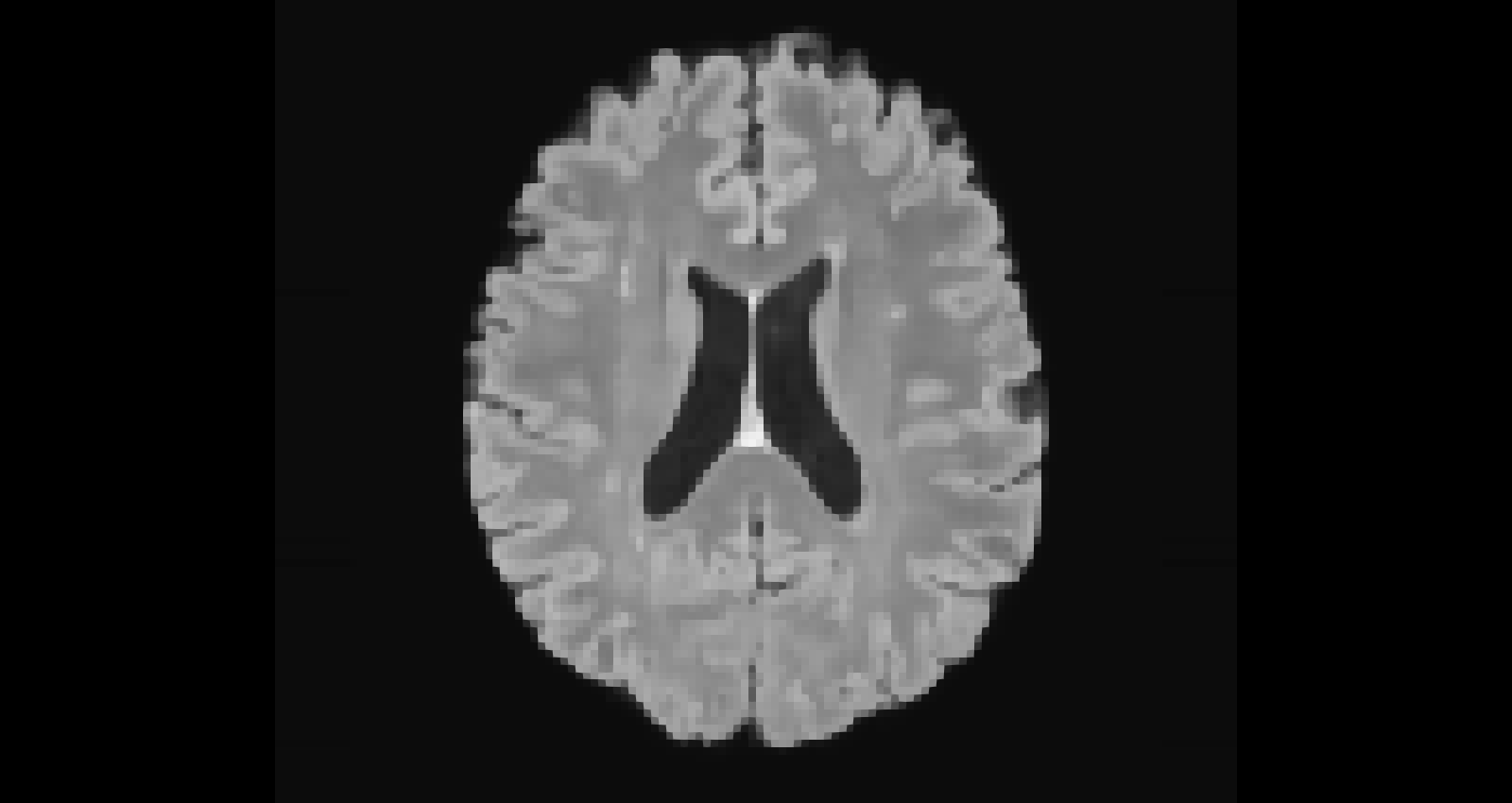}
            \caption{$\lambda_{\text{reg}}=0.1$}
        \end{subfigure}
        \begin{subfigure}{0.22\textwidth}
            \centering
            \includegraphics[trim={17cm 1.25cm 17cm 1.25cm},clip,width=\textwidth]{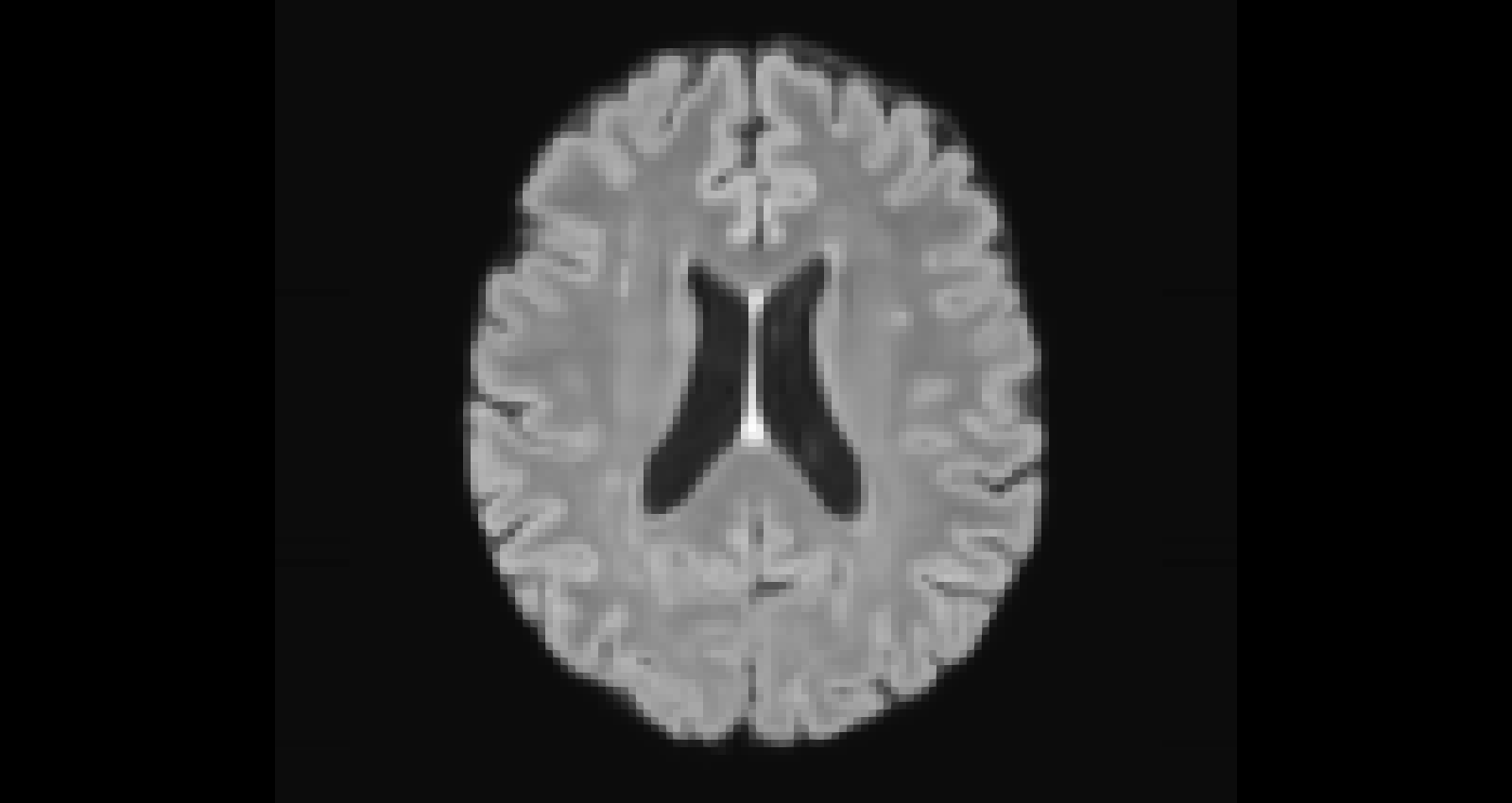}
            \caption{$\lambda_{\text{reg}}=1.2$}
            \label{fig1:lambda_12}
        \end{subfigure}
        \begin{subfigure}{0.22\textwidth}
            \centering
            \includegraphics[trim={17cm 1.25cm 17cm 1.25cm},clip,width=\textwidth]{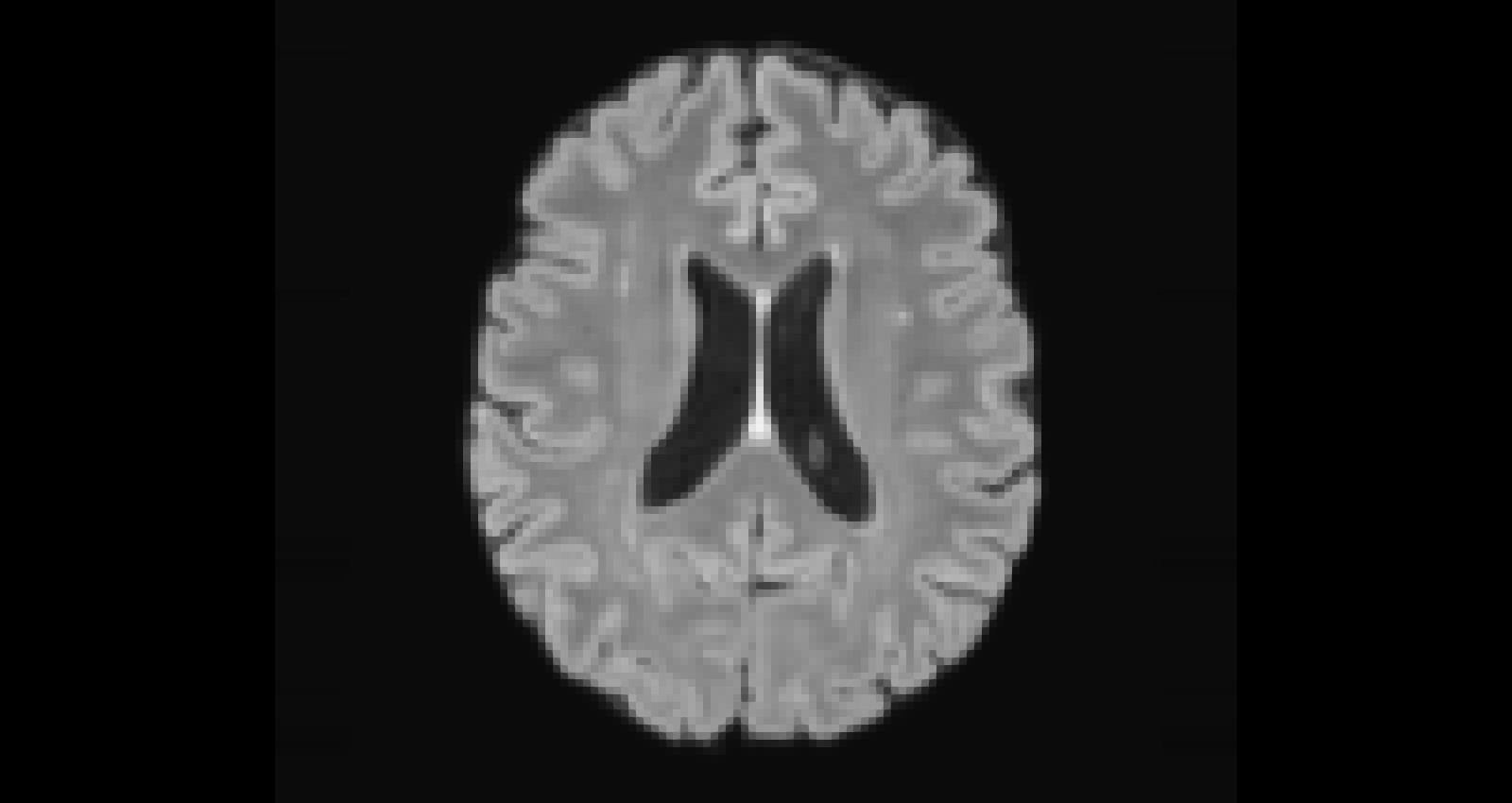}
            \caption{Baseline}
        \end{subfigure}
        \begin{subfigure}{0.22\textwidth}
            \centering
            \includegraphics[trim={17cm 1.25cm 17cm 1.25cm},clip,width=\textwidth]{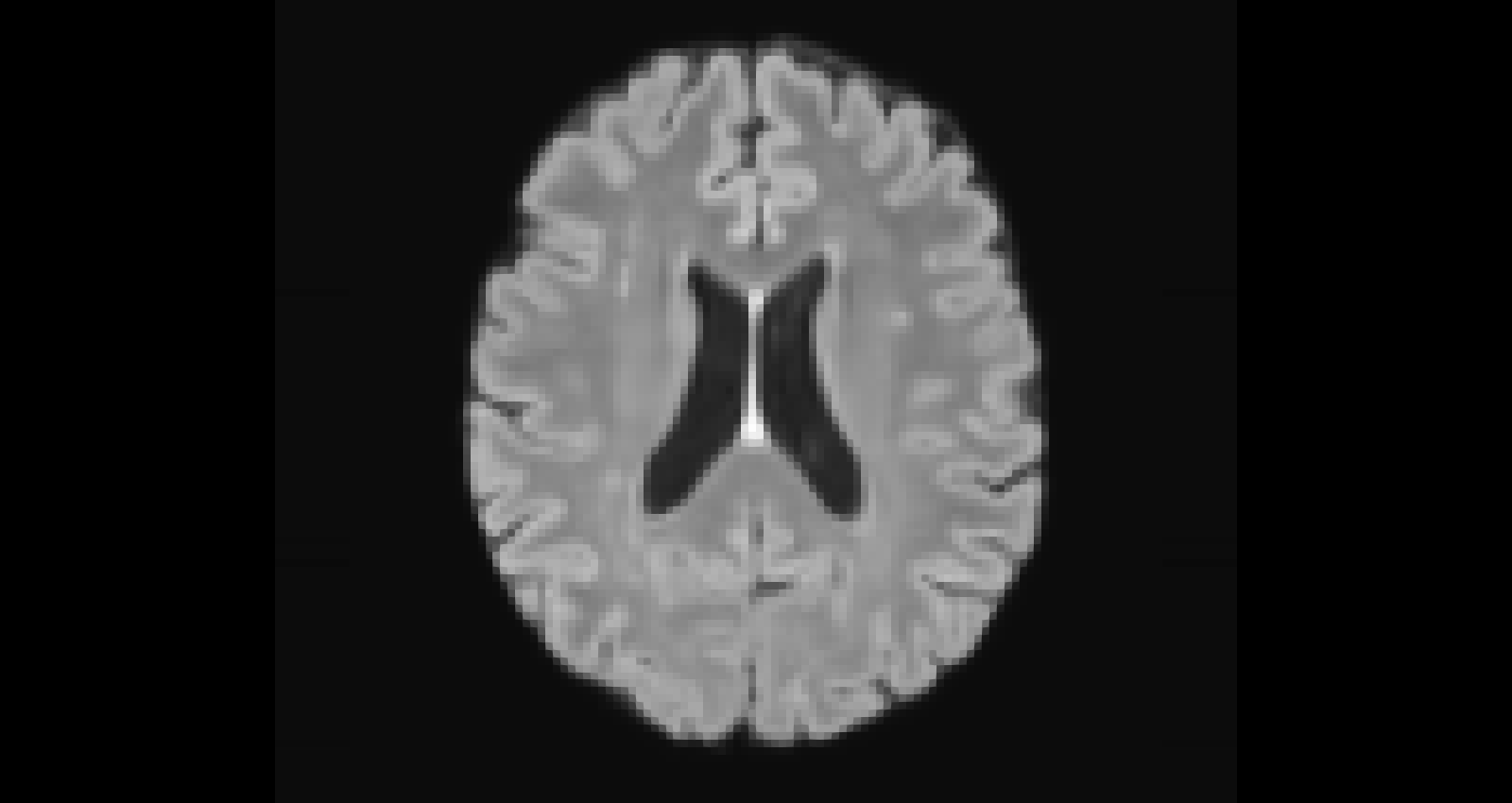}
            \caption{Proposed}
            \label{fig1:ours}
        \end{subfigure}
        \begin{subfigure}{0.08\textwidth}
            \centering
            \includegraphics[width=0.5\textwidth,height=2.5cm]{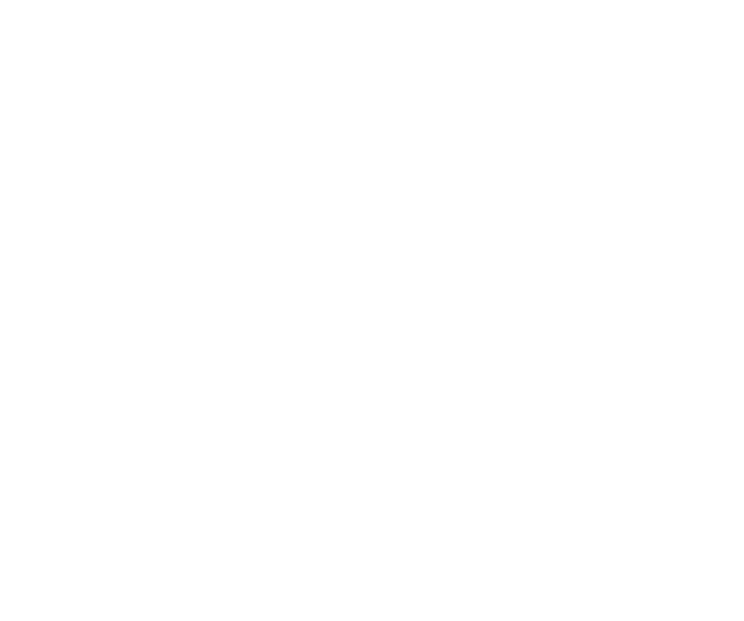}
            \caption*{}
        \end{subfigure}
    \endminipage

    \caption{Output when using two fixed regularisation weights $\lambda_{\text{reg}} \in \{0.1, 1.2\}$, the baseline method for learnable regularisation strength, and the proposed learnable regularisation loss with $\lambda_{\text{init}} = 1.2$. For fixed regularisation weight $\lambda_{\text{reg}} = 0.1$, the sampled transformations are not diffeomorphic. In case of the baseline method, the learnt regularisation strength is too high, which effectively prevents the alignment of images. When using the proposed learnable regularisation loss, we strike a balance between the baseline method and fixed regularisation weight $\lambda_{\text{reg}} = 0.1$, where the regularisation strength is too low. The figure shows the middle axial slice of 3D images.}
    \label{fig:reg}
\end{figure}

We compare the output of \gls{VI} when using fixed regularisation weights $\lambda_{\text{reg}} \in \{0.1, 1.2\}$, the baseline method for learnable regularisation strength, and our regularisation loss. The result on a sample pair of input images is shown in \Cref{fig:reg}. For the baseline method, the learnt regularisation strength is too high, which effectively prevents the alignment of images. This indicates that previous schemes for inference of regularisation strength from data are inadequate when the transformation parametrisation involves a very large number of degrees of freedom. In case of $\lambda_{\text{reg}} = 0.1$, the resulting transformation is not diffeomorphic. The output when using our regularisation loss with $\lambda_{\text{init}} = 1.2$ strikes a balance between the baseline and $\lambda_{\text{reg}} = 0.1$, where there is an overemphasis on the data term.

\begin{figure}[!htb]
    \centering
    
    \minipage[c]{0.65\linewidth}
        \begin{subfigure}{0.21\textwidth}
            \centering
            \includegraphics[width=\textwidth]{figures/empty.png}
        \end{subfigure}
        \begin{subfigure}{0.21\textwidth}
            \centering
            \includegraphics[trim={17cm 1.25cm 17cm 1.25cm},clip,width=\textwidth]{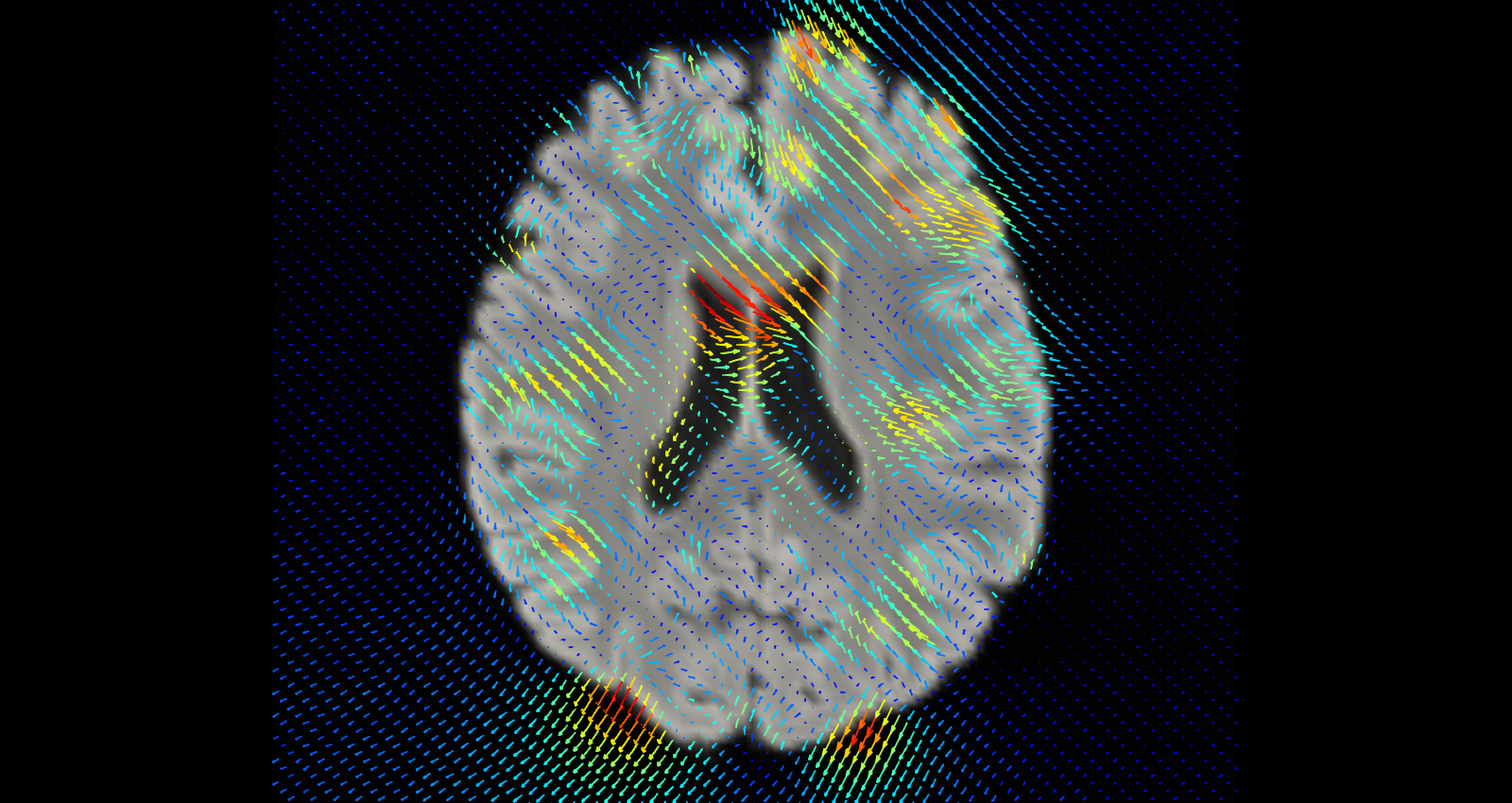}
        \end{subfigure}
        \begin{subfigure}{0.21\textwidth}
            \centering
            \includegraphics[trim={17cm 1.25cm 17cm 1.25cm},clip,width=\textwidth]{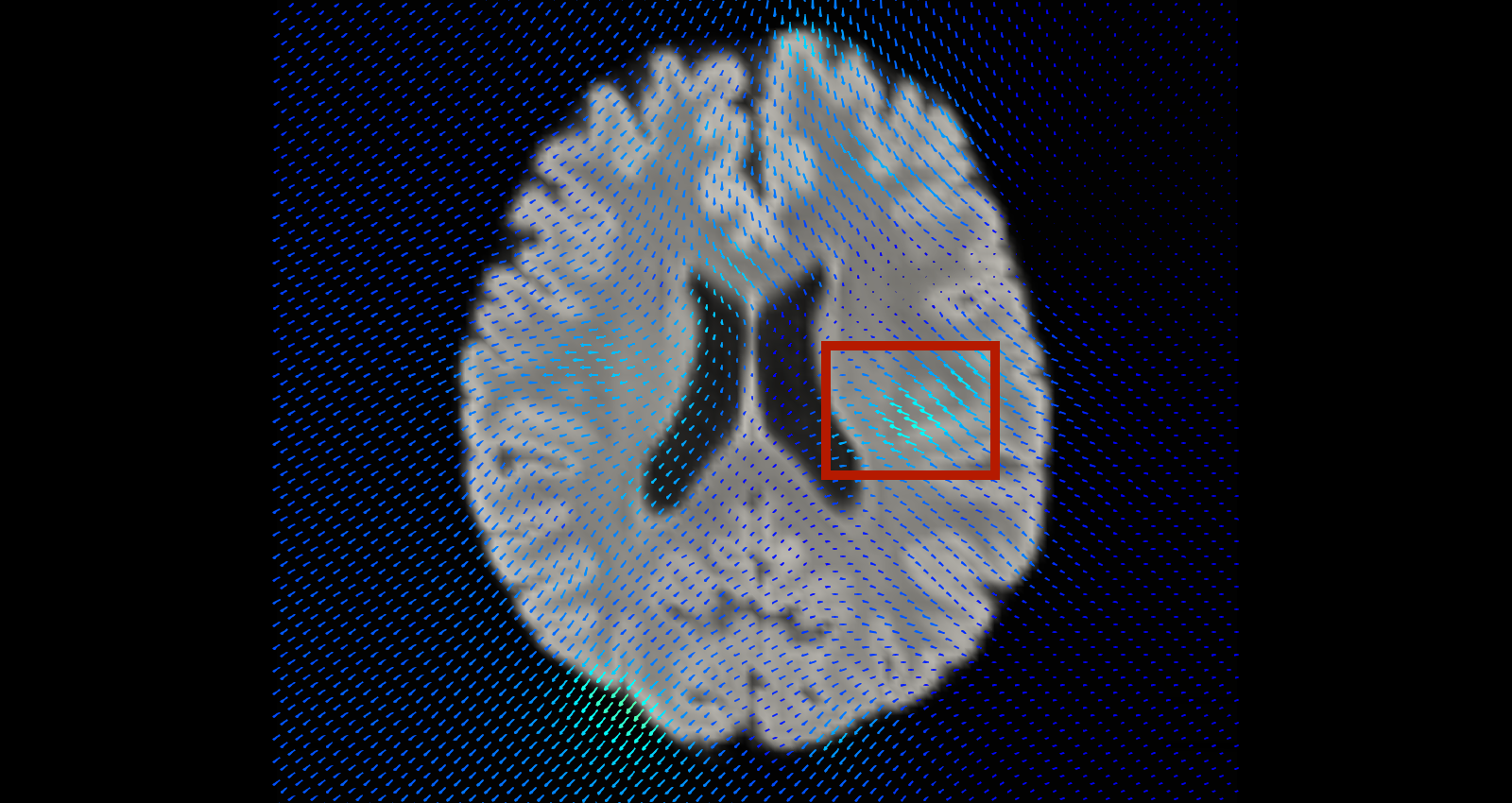}
        \end{subfigure}
        \begin{subfigure}{0.21\textwidth}
            \centering
            \includegraphics[trim={17cm 1.25cm 17cm 1.25cm},clip,width=\textwidth]{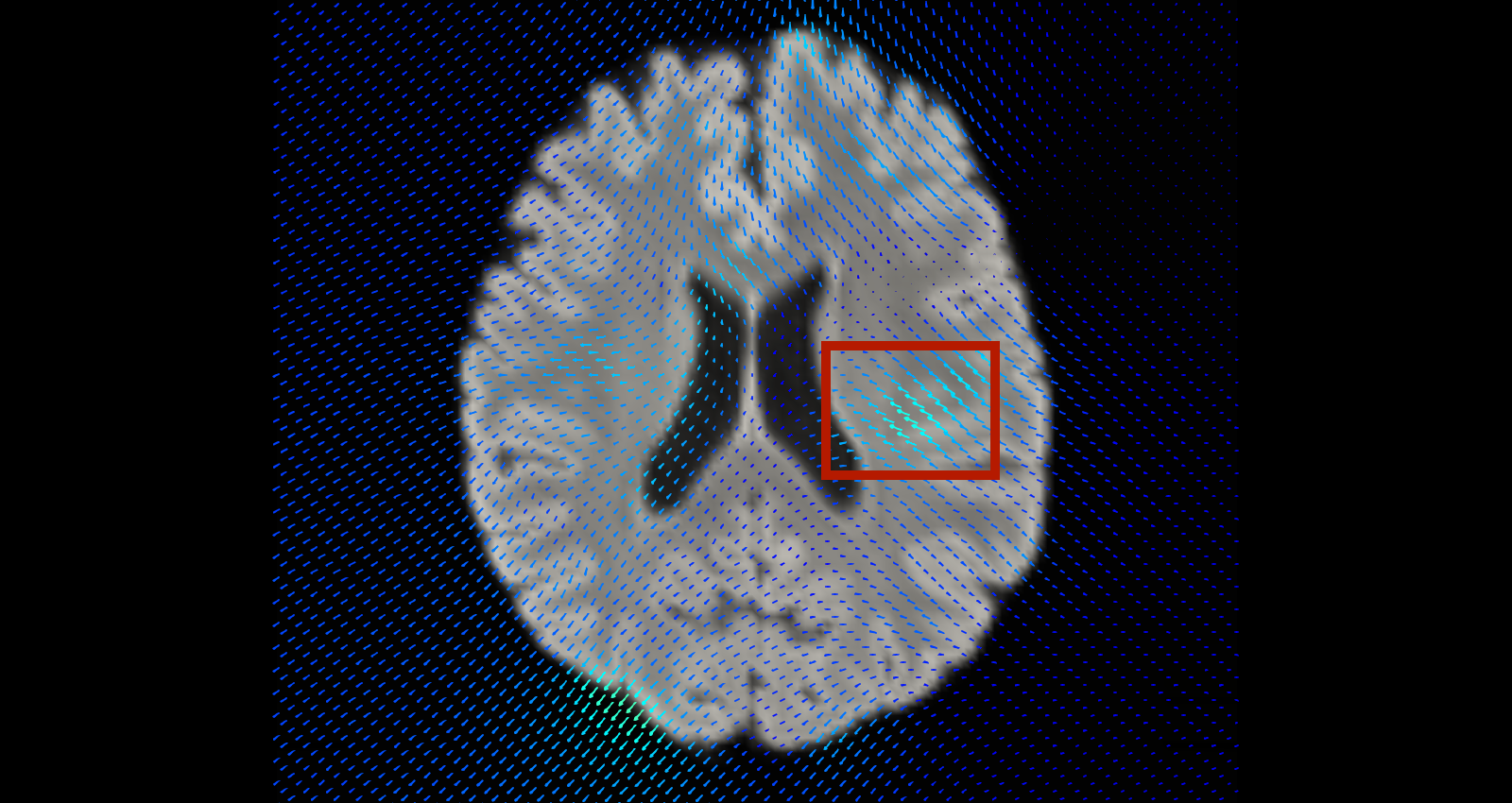}
        \end{subfigure}
        \begin{subfigure}{0.08\textwidth}
            \centering
            \includegraphics[trim={36cm 5cm 12cm 5cm},clip,height=2.5cm]{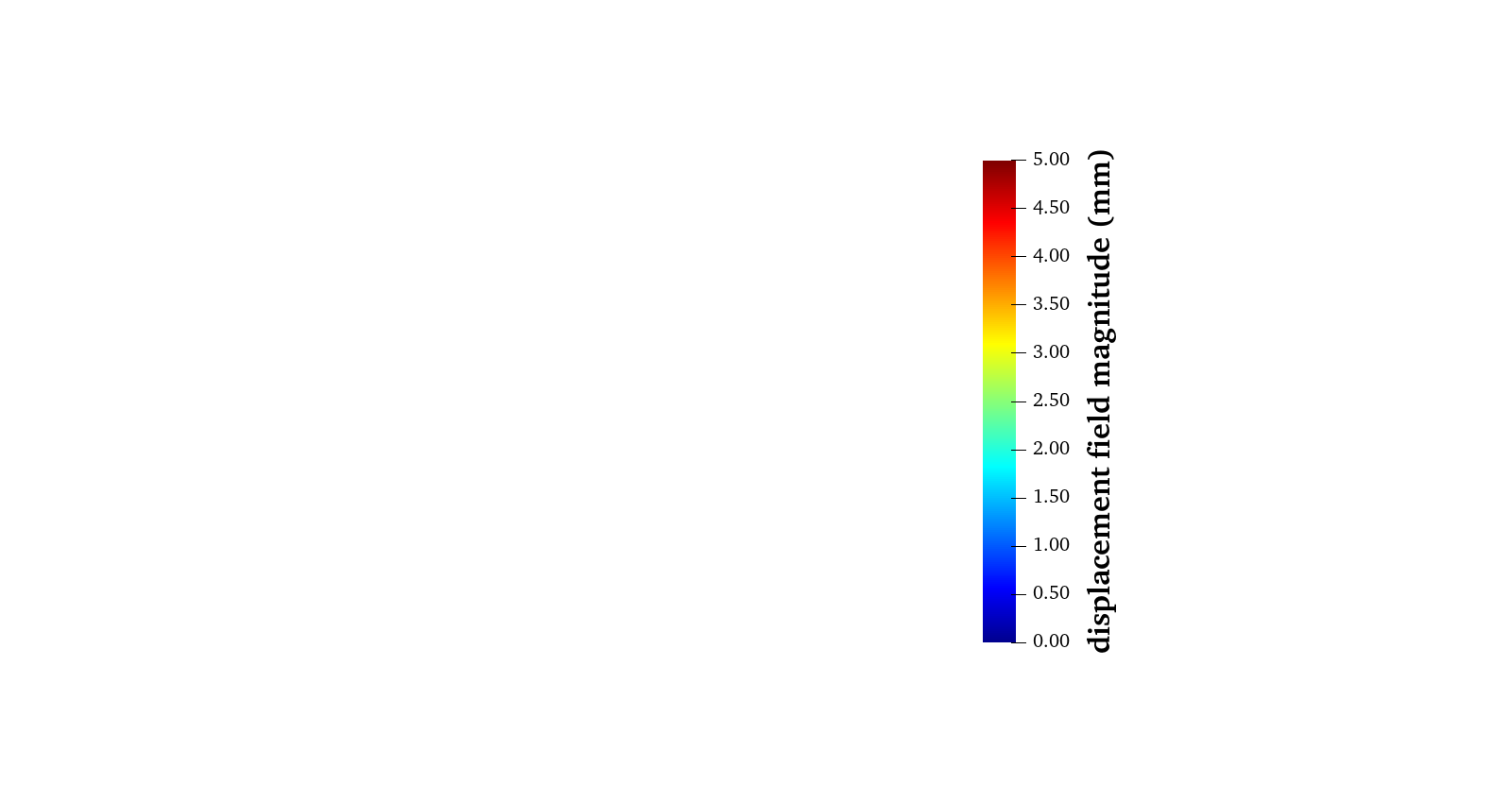}
        \end{subfigure}
    \endminipage
    \minipage[c]{0.15\linewidth}
        \caption*{\centering\footnotesize Fixed image $F$ and mean displacement}
    \endminipage
    
    \minipage[c]{0.65\linewidth}
        \begin{subfigure}{0.21\textwidth}
            \centering
            \includegraphics[width=\textwidth]{figures/empty.png}
        \end{subfigure}
        \begin{subfigure}{0.21\textwidth}
            \centering
            \includegraphics[width=\textwidth]{figures/empty.png}
        \end{subfigure}
        \begin{subfigure}{0.21\textwidth}
            \centering
            \includegraphics[width=\textwidth]{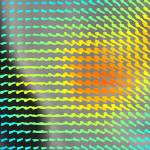}
        \end{subfigure}
        \begin{subfigure}{0.21\textwidth}
            \centering
            \includegraphics[width=\textwidth]{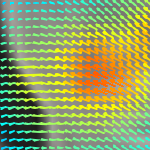}
        \end{subfigure}
        \begin{subfigure}{0.08\textwidth}
            \centering
            \includegraphics[trim={36cm 5cm 12cm 5cm},clip,height=2.5cm]{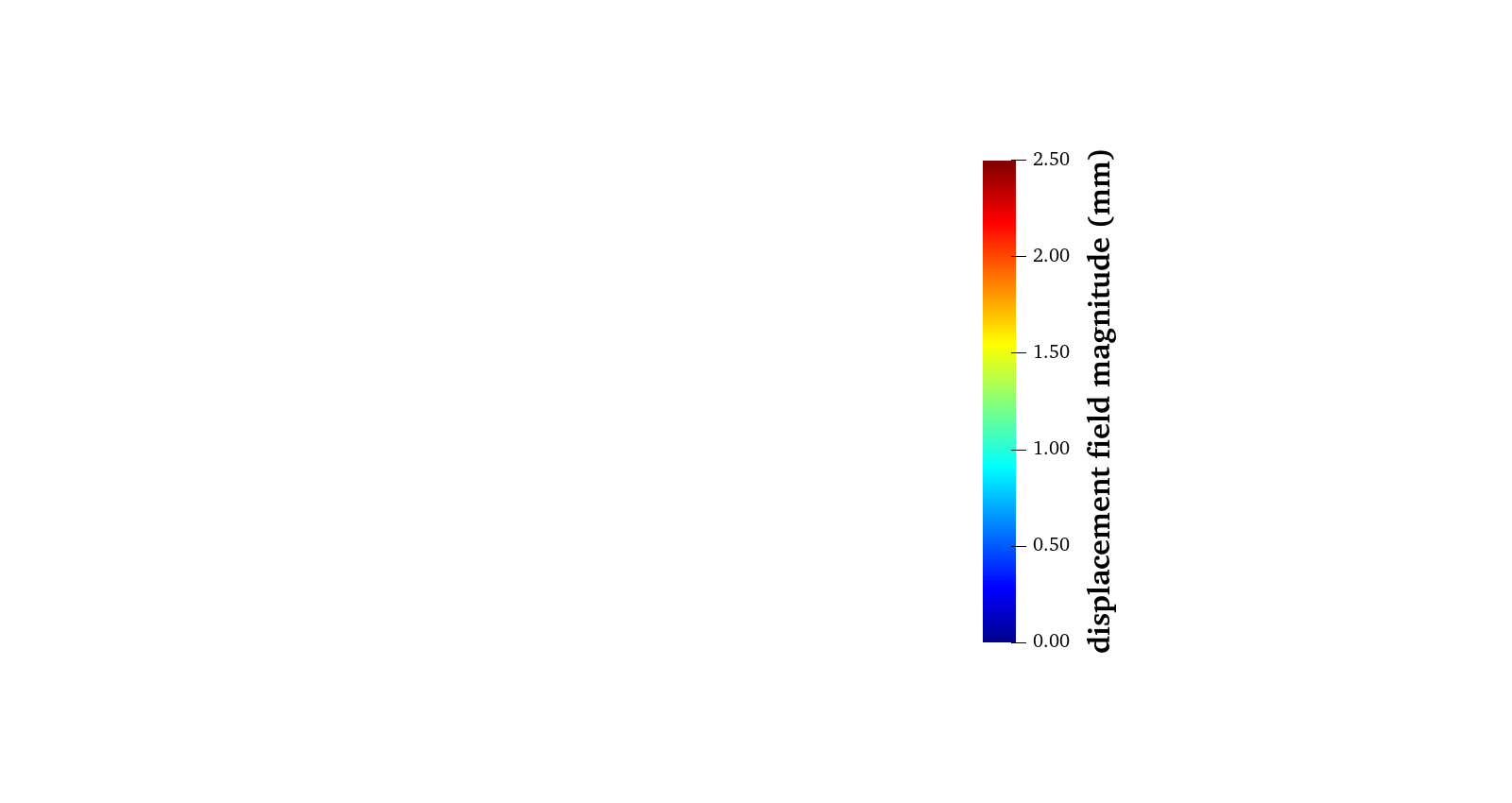}
        \end{subfigure}
    \endminipage
    \minipage[c]{0.15\linewidth}
        \caption*{\centering\footnotesize Zoom on the lateral ventricle area}
    \endminipage

    \minipage[c]{0.65\linewidth}
        \begin{subfigure}{0.21\textwidth}
            \centering
            \includegraphics[trim={17cm 1.25cm 17cm 1.25cm},clip,width=\textwidth]{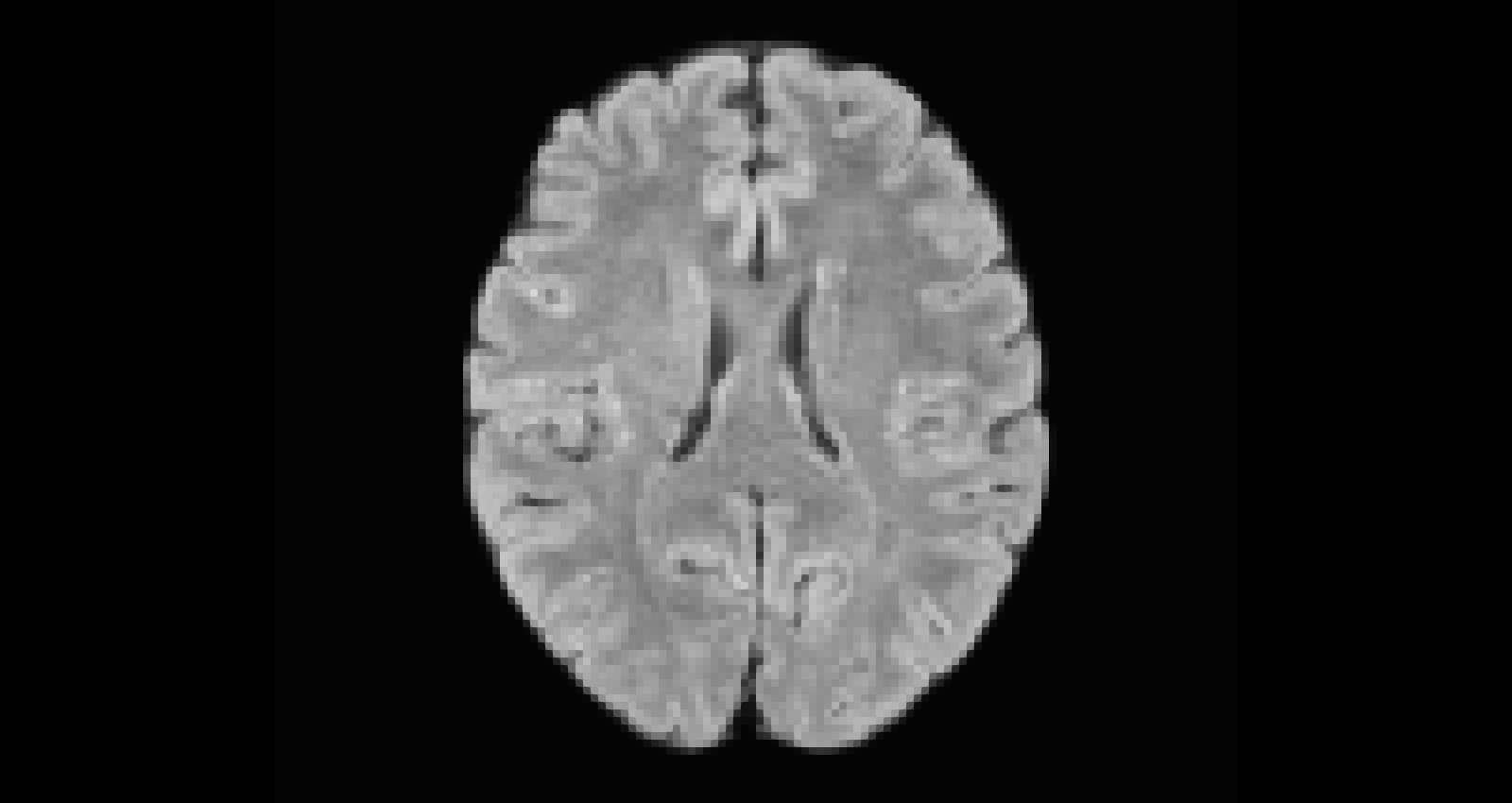}
            \caption*{$M_1$}
        \end{subfigure}
        \begin{subfigure}{0.21\textwidth}
            \centering
            \includegraphics[trim={17cm 1.25cm 17cm 1.25cm},clip,width=\textwidth]{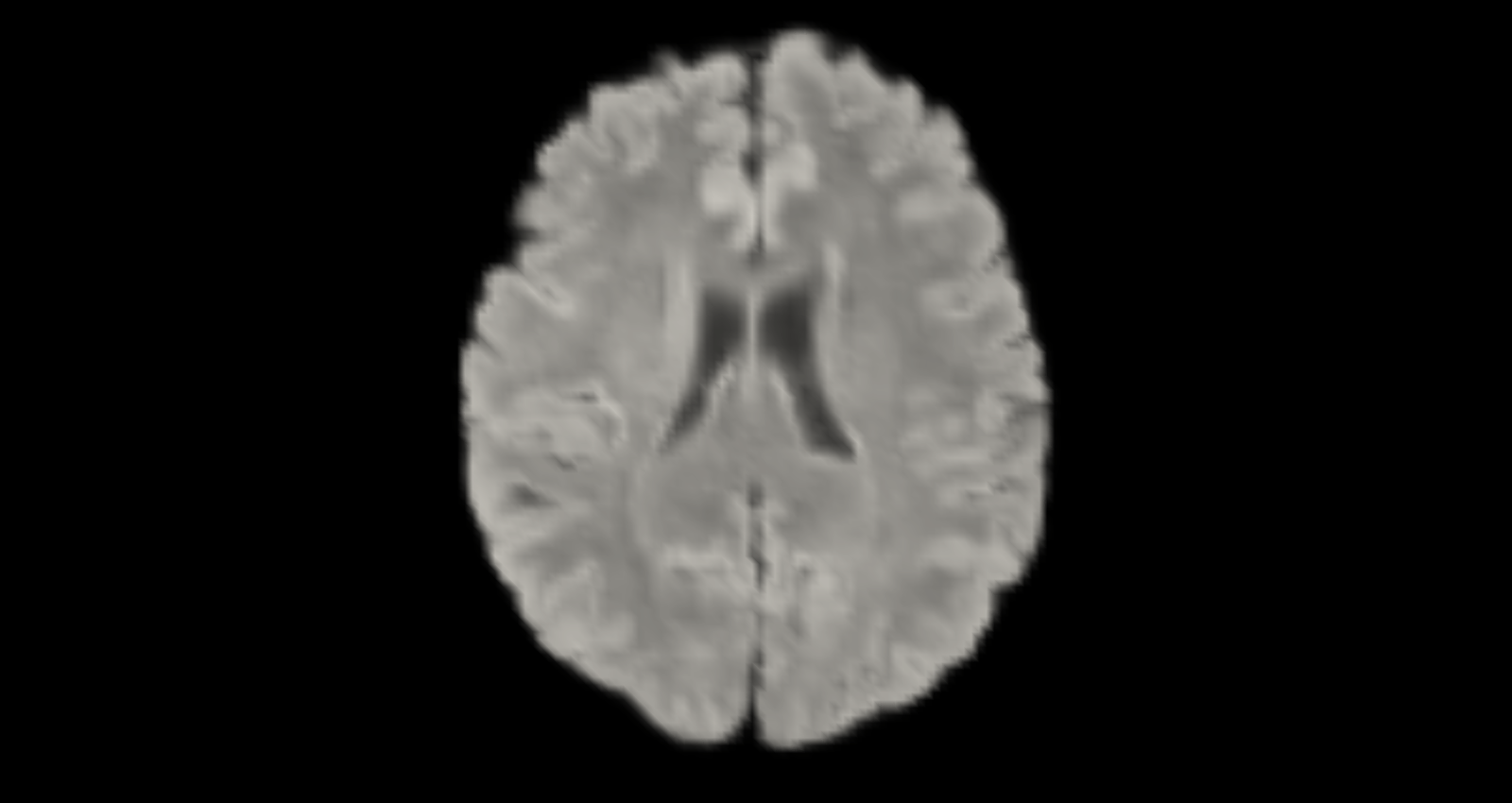}
            \caption{$\lambda_{\text{reg}} = 0.2$}
        \end{subfigure}
        \begin{subfigure}{0.21\textwidth}
            \centering
            \includegraphics[trim={17cm 1.25cm 17cm 1.25cm},clip,width=\textwidth]{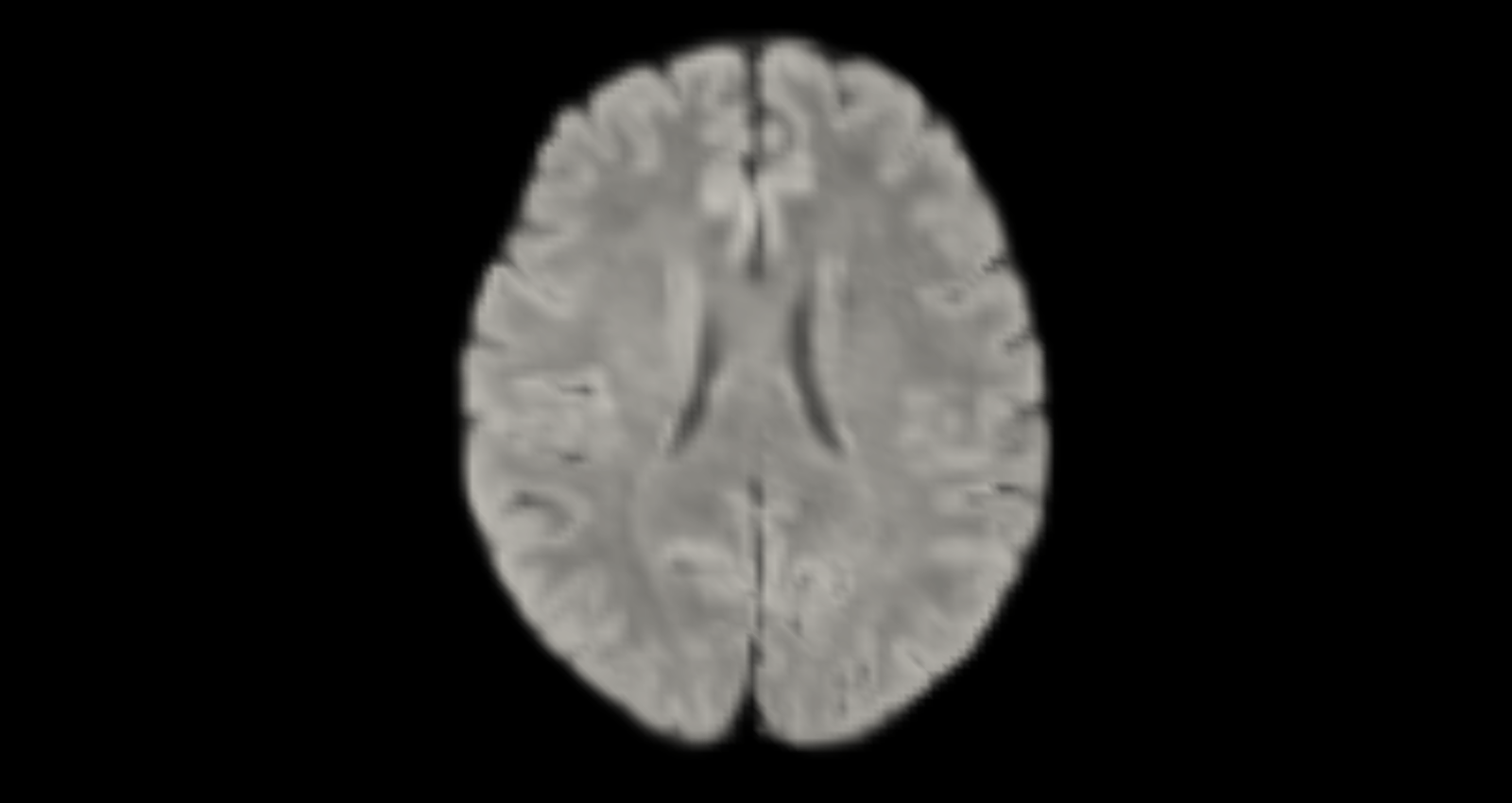}
            \caption{$\lambda_{\text{reg}} = 2.0$}
        \end{subfigure}
        \begin{subfigure}{0.21\textwidth}
            \centering
            \includegraphics[trim={17cm 1.25cm 17cm 1.25cm},clip,width=\textwidth]{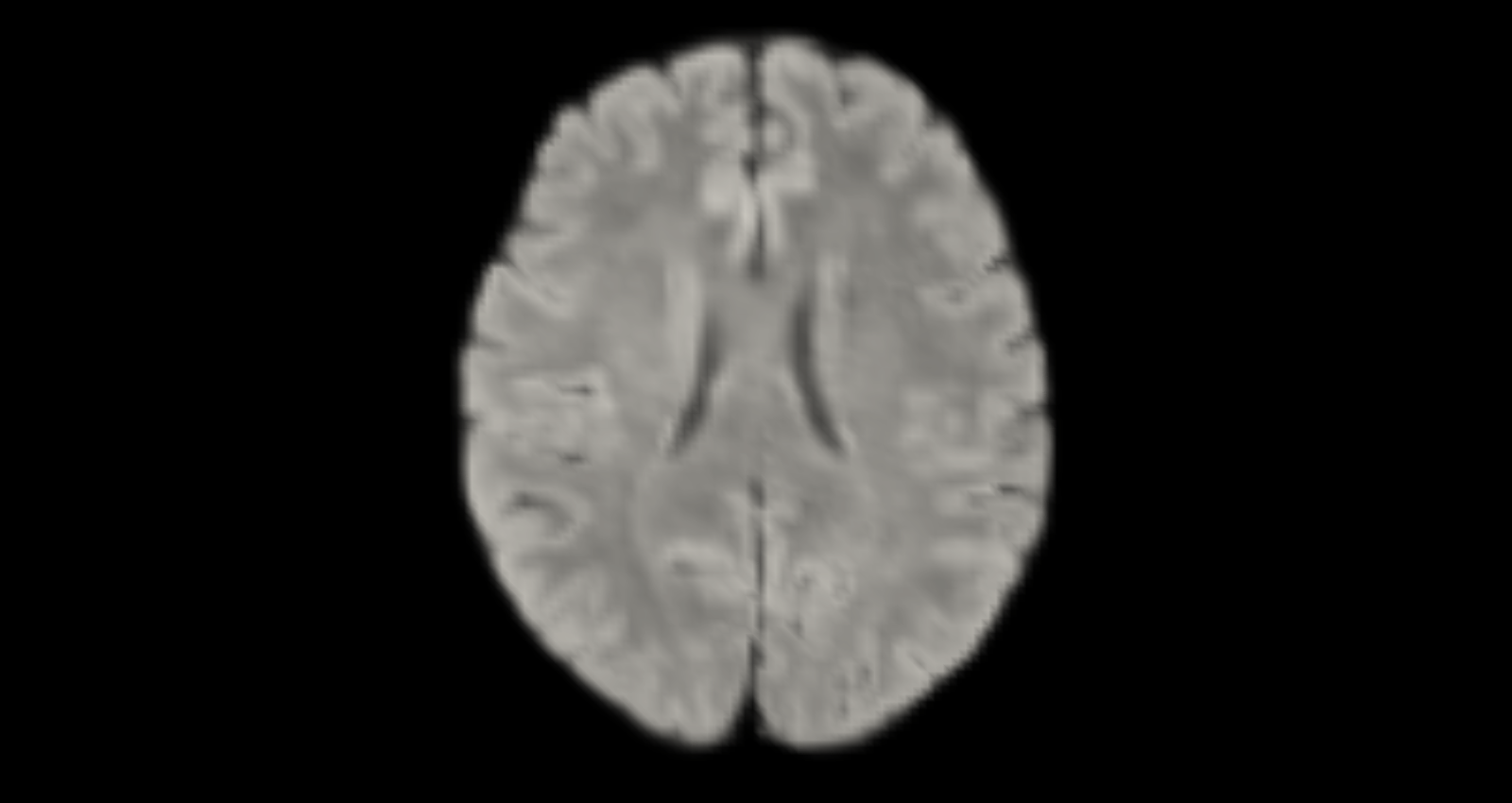}
            \caption{Proposed}
        \end{subfigure}
    \endminipage
    \minipage[c]{0.15\linewidth}
        \caption*{\centering\footnotesize Warped moving image $M_1\left( \mu_w \right)$}
    \endminipage

    \minipage[c]{0.65\linewidth}
        \begin{subfigure}{0.21\textwidth}
            \centering
            \includegraphics[width=\textwidth]{figures/empty.png}
        \end{subfigure}
        \begin{subfigure}{0.21\textwidth}
            \centering
            \includegraphics[trim={17cm 1.25cm 17cm 1.25cm},clip,width=\textwidth]{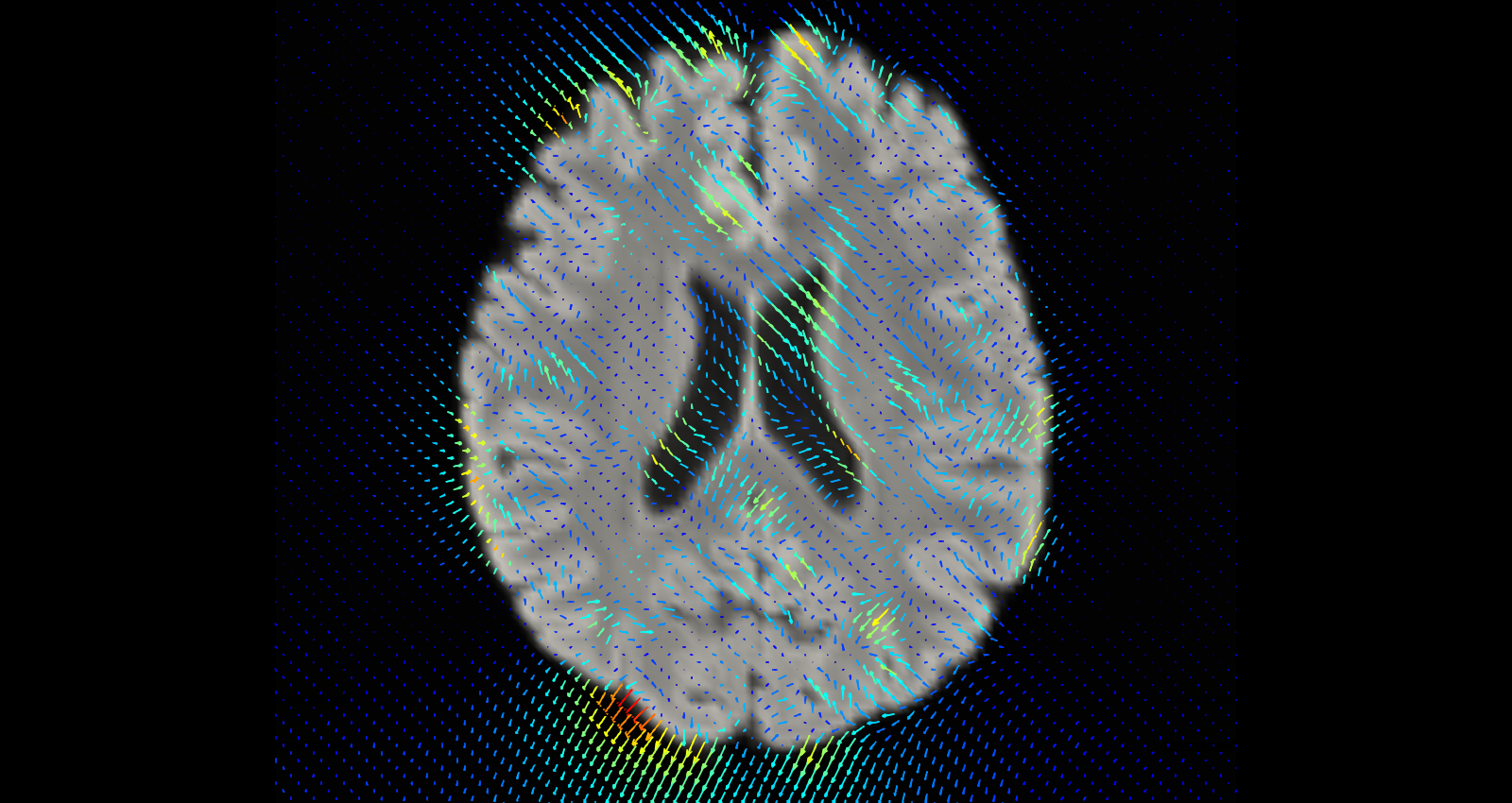}
        \end{subfigure}
        \begin{subfigure}{0.21\textwidth}
            \centering
            \includegraphics[trim={17cm 1.25cm 17cm 1.25cm},clip,width=\textwidth]{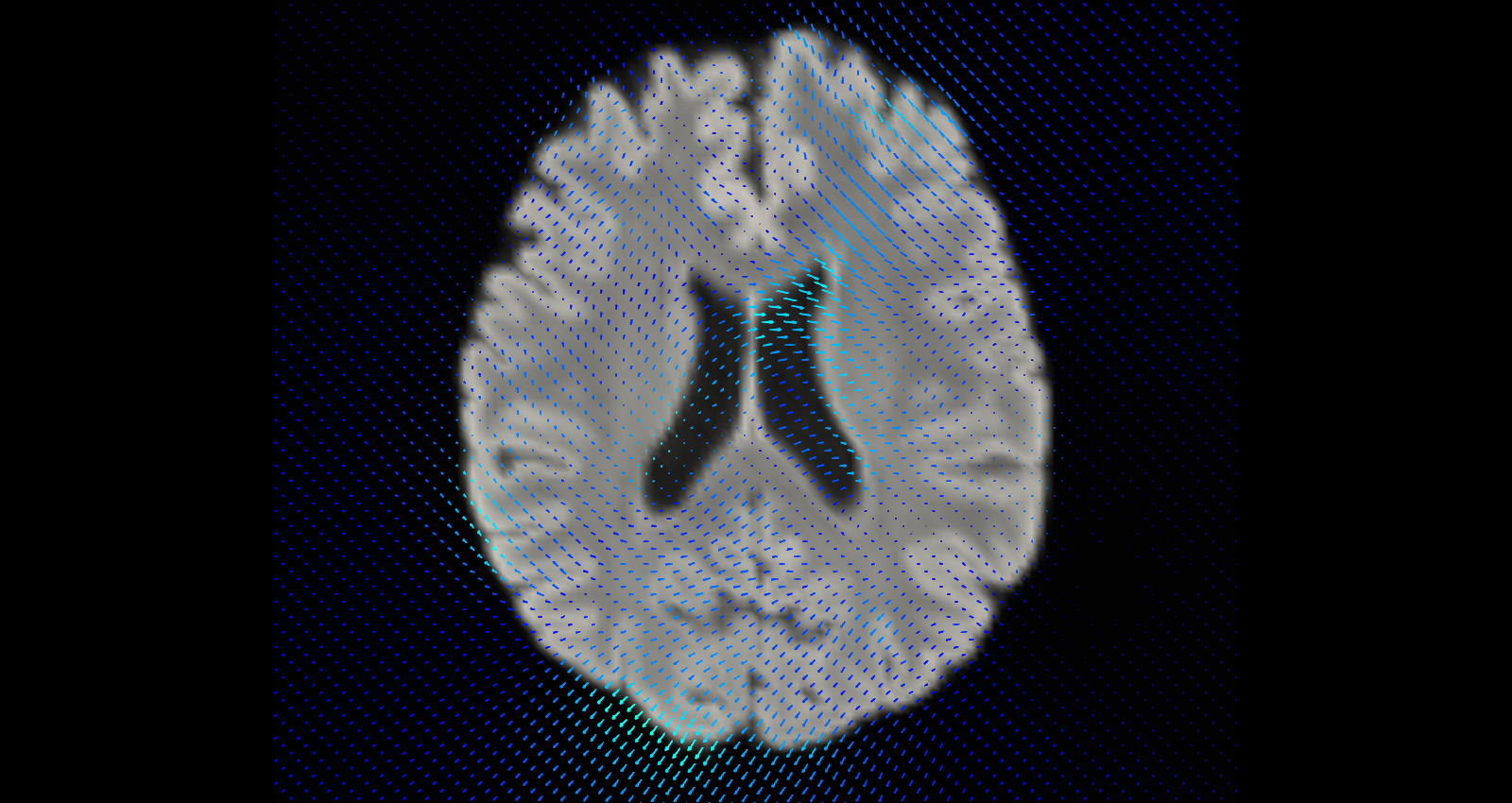}
        \end{subfigure}
        \begin{subfigure}{0.21\textwidth}
            \centering
            \includegraphics[trim={17cm 1.25cm 17cm 1.25cm},clip,width=\textwidth]{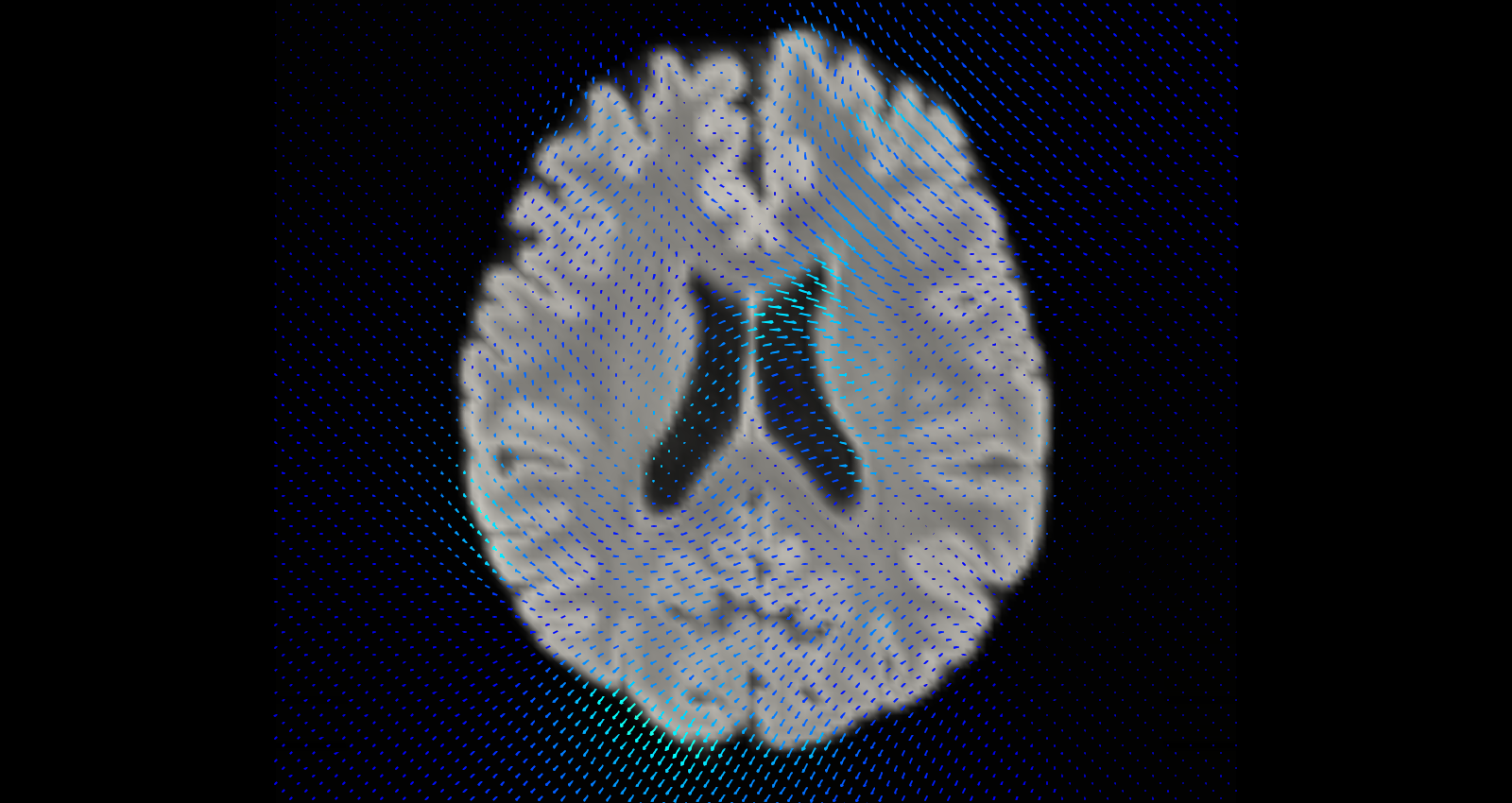}
        \end{subfigure}
        \begin{subfigure}{0.08\textwidth}
            \centering
            \includegraphics[trim={36cm 5cm 12cm 5cm},clip,height=2.5cm]{figures/2/M1/colorbar.png}
        \end{subfigure}
    \endminipage
    \minipage[c]{0.15\linewidth}
        \caption*{\centering\footnotesize Fixed image $F$ and mean displacement}
    \endminipage

    \minipage[c]{0.65\linewidth}
        \begin{subfigure}{0.21\textwidth}
            \centering
            \includegraphics[trim={17cm 1.25cm 17cm 1.25cm},clip,width=\textwidth]{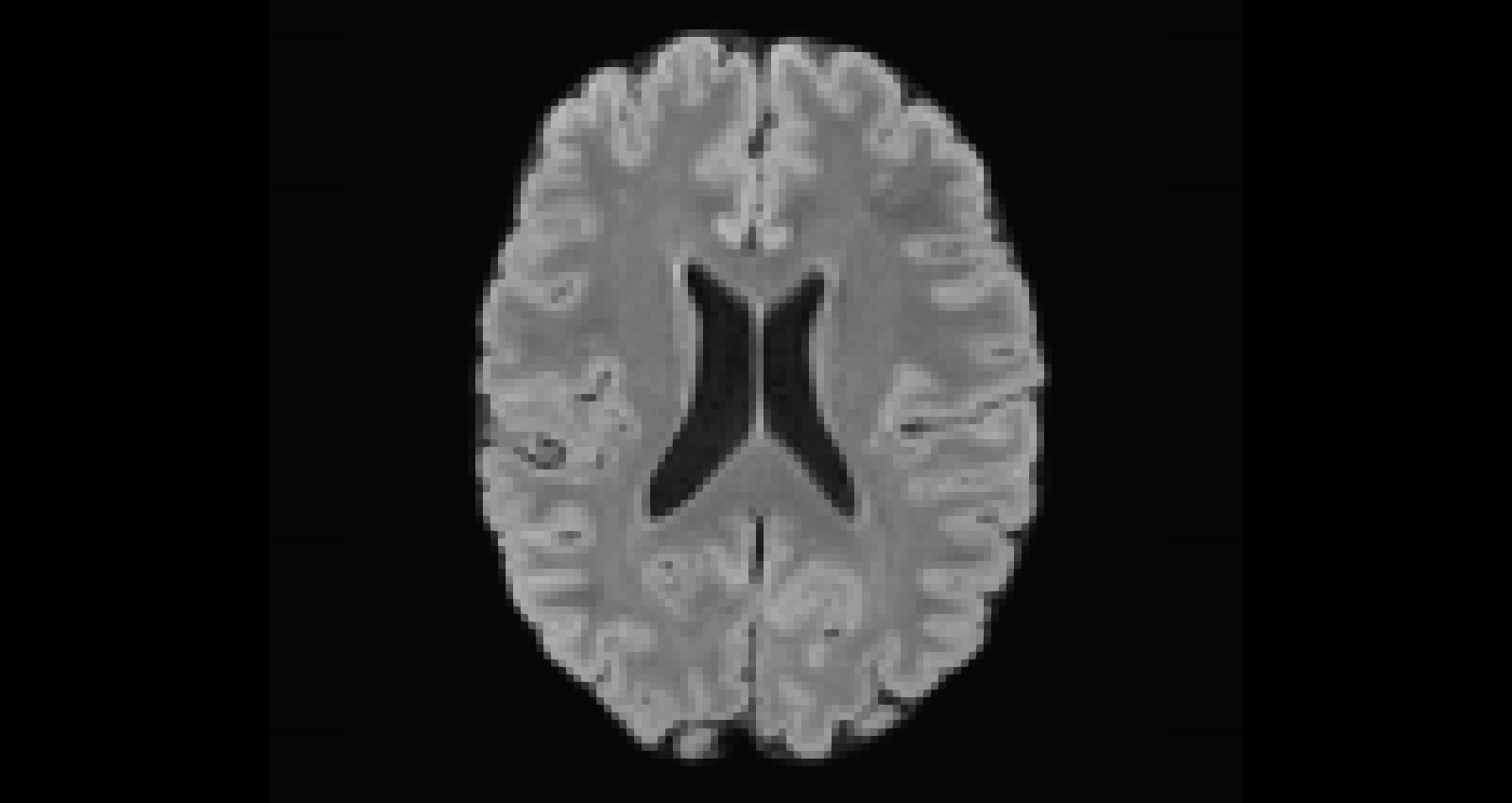}
            \caption*{$M_2$}
        \end{subfigure}
        \begin{subfigure}{0.21\textwidth}
            \centering
            \includegraphics[trim={17cm 1.25cm 17cm 1.25cm},clip,width=\textwidth]{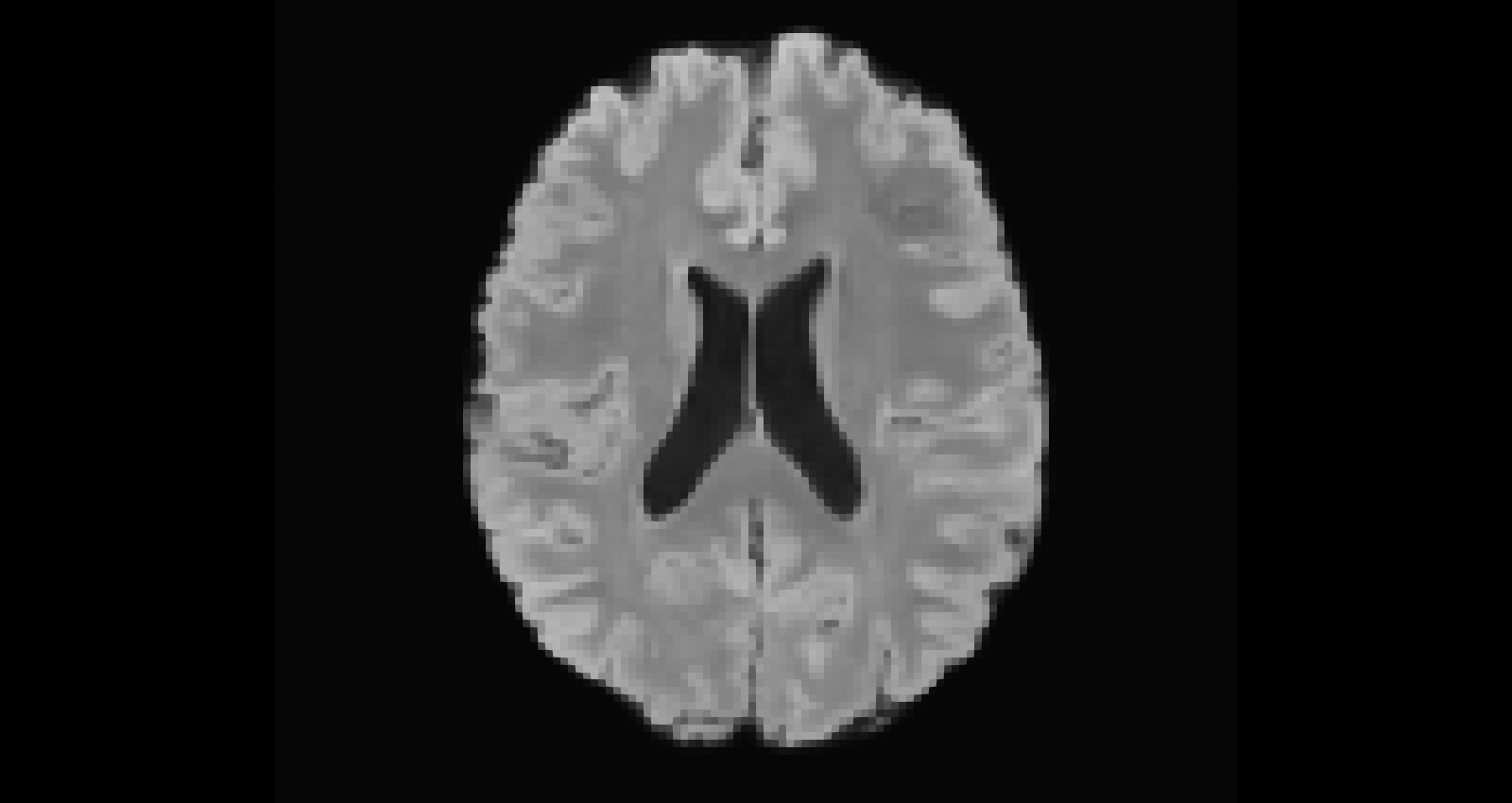}
            \caption{$\lambda_{\text{reg}} = 0.2$}
        \end{subfigure}
        \begin{subfigure}{0.21\textwidth}
            \centering
            \includegraphics[trim={17cm 1.25cm 17cm 1.25cm},clip,width=\textwidth]{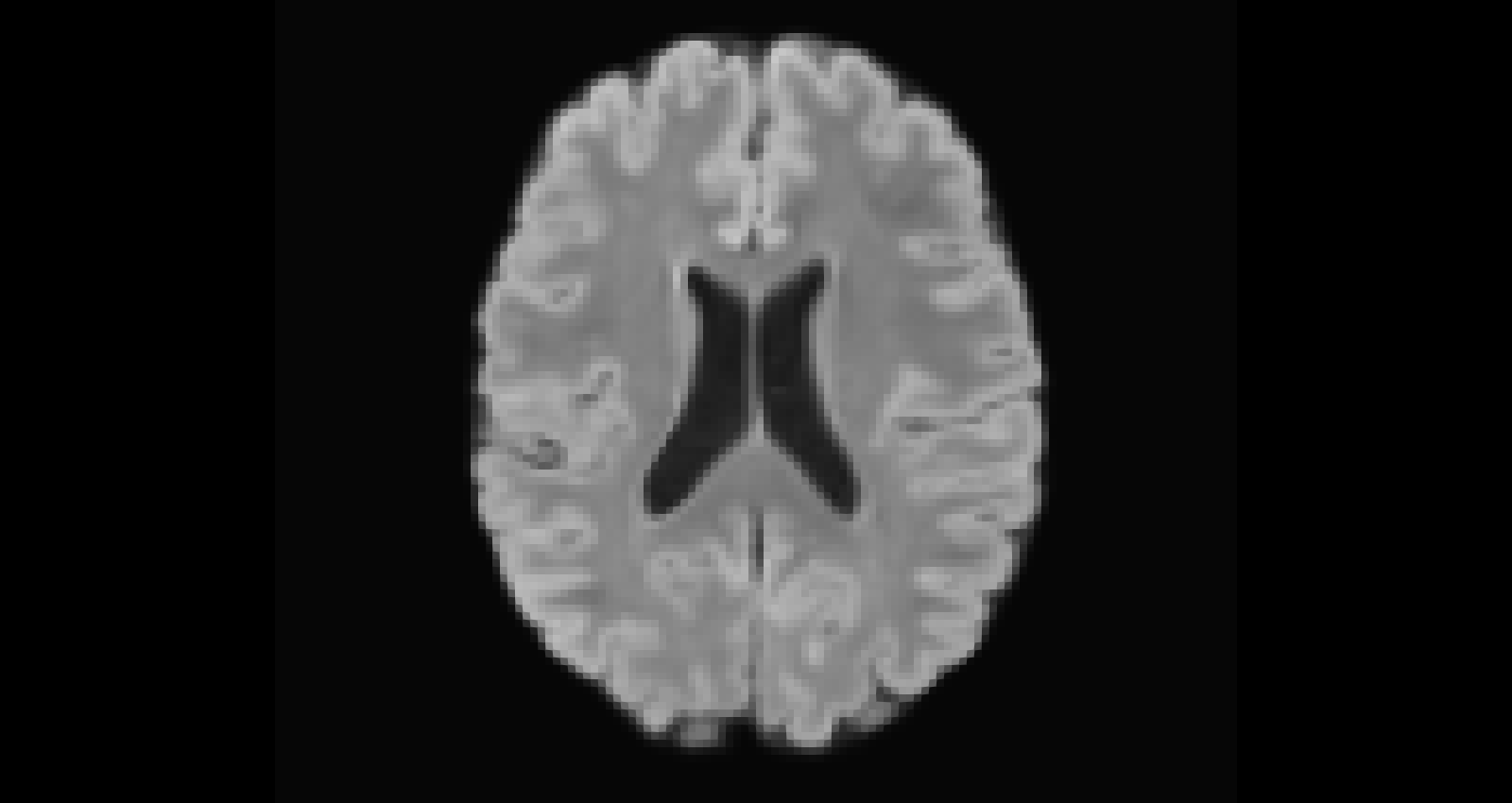}
            \caption{$\lambda_{\text{reg}} = 2.0$}
        \end{subfigure}
        \begin{subfigure}{0.21\textwidth}
            \centering
            \includegraphics[trim={17cm 1.25cm 17cm 1.25cm},clip,width=\textwidth]{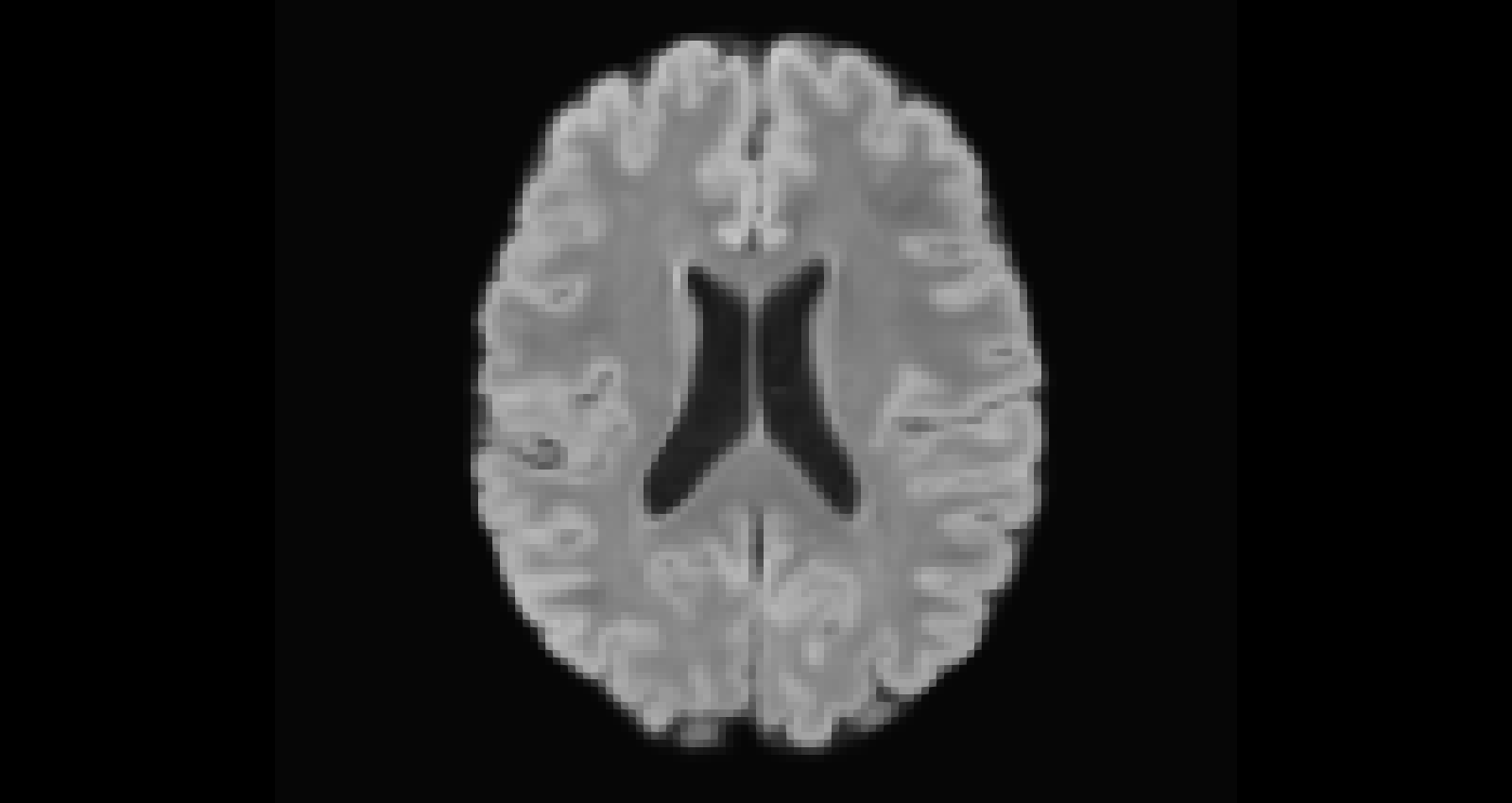}
            \caption{Proposed}
        \end{subfigure}
    \endminipage
    \minipage[c]{0.15\linewidth}
        \caption*{\centering\footnotesize Warped moving image $M_2\left( \mu_w \right)$}
    \endminipage

    \caption{Output of \gls{VI} for two image pairs which require different regularisation strengths. $M_1$ is visibly different to the fixed image and $M_2$ is similar to it. The proposed regularisation is initialised with $\lambda_{\text{init}} = 2.0$. For both image pairs, the alignment is best in case of the fixed regularisation strength $\lambda_{\text{reg}} = 0.2$ but the output transformation is not diffeomorphic. For $M_2$, the proposed learnable regularisation loss is almost identical to $\lambda_{\text{reg}} = 2.0$, which ensures good accuracy and smoothness of the transformation. For $M_1$, the proposed loss outputs somewhat higher accuracy than $\lambda_{\text{reg}} = 2.0$. The figure shows the middle axial slice of 3D images.}
    \label{fig:reg2}
\end{figure}

In \Cref{fig:reg2}, we show the output of \gls{VI} for two pairs of images which require different regularisation strengths for accurate alignment. We choose a fixed image and two moving images $M_1$ and $M_2$, with one visibly different and the other similar to the fixed image. We also analyse the result when using fixed regularisation weights $\lambda_\text{reg} = 0.2$, which leads to non-diffeomorphic transformations, and $\lambda_\text{reg} = 2.0$, which produces smooth transformations but, in case of $M_1$, at the expense of accuracy. The proposed regularisation, initialised with $\lambda_{\text{init}} = 2.0$, helps to prevent oversmoothing.

%---------------------------------------------------------------
\subsection{Uncertainty quantification}
\label{sec:UQ}

We run a number of experiments to evaluate the uncertainty estimates and better understand the differences between uncertainty output by various non-rigid registration methods in practice: \begin{enumerate}
    \item We compare the uncertainty estimates output by \gls{VI}, \gls{SG-MCMC}, and VoxelMorph on inter-subject brain \gls{MRI} data from UK Biobank qualitatively and quantitatively, by calculating the Pearson correlation coefficient between the displacement and label uncertainties;
    \item We compare the uncertainty estimates when the \gls{SG-MCMC} algorithm is initialised to different transformations;
    \item We compare the result when using non-parametric \glspl{SVF} and \glspl{SVF} based on B-splines to parametrise the transformation.
\end{enumerate}

In order to make sampling from \gls{SG-MCMC} efficient, we determine the largest step size that guarantees diffeomorphic transformations as defined in \Cref{sec:setup} and set it to $\tau = 4 \times 10^{-1}$. A single Markov chain requires only \SI{4}{\giga\byte} of memory, so \nochains{} Markov chains are run in parallel, each initialised to a different sample from the approximate variational posterior. We discard the first \nosamplesburnin{} samples from each chain to allow the Markov chains to reach the stationary distribution.

\subsubsection{Comparison of uncertainty estimates output by different models}
\label{sec:exp3}

\textbf{VoxelMorph.} %<*VXM1>
To fill the knowledge gaps on differences between uncertainty estimates produced by non-rigid image registration algorithms, we train the probabilistic VoxelMorph \citep{Dalca2018} in atlas mode on a random $80/20$ split of \numprint{13401} brain \gls{MRI} scans in the UK Biobank\footnote{The official VoxelMorph implementation used in the experiments is available on GitHub: \url{https://github.com/voxelmorph/voxelmorph}.}. We use the same fixed image as in the experiments and exclude the moving images from the training data. To enable a fair comparison, the chosen similarity metric is the \gls{SSD}. %</VXM1>
We also study the differences between uncertainty estimates output by \gls{VI} and \gls{SG-MCMC}, based on \nosamples{} samples output by each model. %<*autocorrelation>
In order to reduce auto-correlation, samples output by \gls{SG-MCMC} are selected at regular intervals from the \nosamplestotal{} samples drawn from each chain.
%</autocorrelation>

%<*VXM2>
Like in \gls{VI}, which we use to initialise the \gls{SG-MCMC} algorithm, the approximate variational posterior of transformation parameters output by VoxelMorph is assumed to be a multivariate normal distribution $q_{\text{VXM}} \left( w \right) \sim \mathcal{N} \left( \mu_{\text{VXM}}, \Sigma_{\text{VXM}} \right)$. The only difference is the covariance matrix $\Sigma_\text{VXM}$, which is assumed to be diagonal rather than diagonal + low-rank. In order to set the model hyperparameters, we analyse the \glspl{ASD} and \glspl{DSC} on subcortical structure segmentations and the Jacobian determinants of sample transformations, with the aim of striking a balance between accuracy and smoothness. The most important hyperparameters are the the fixed regularisation strength parameter, set to $\lambda_{\text{VXM}} = 10.0$, and the initial value of the diagonal covariance matrix, set to $\Sigma_{\text{VXM}} = \text{diag} \left( 0.01 \right)$.
%</VXM2>

\begin{figure}[!htb]
    \centering
    \minipage[c]{0.5\linewidth}
        \begin{subfigure}{0.3\textwidth}
            \centering
            \includegraphics[trim={18cm 3cm 18cm 3cm},clip,width=\textwidth]{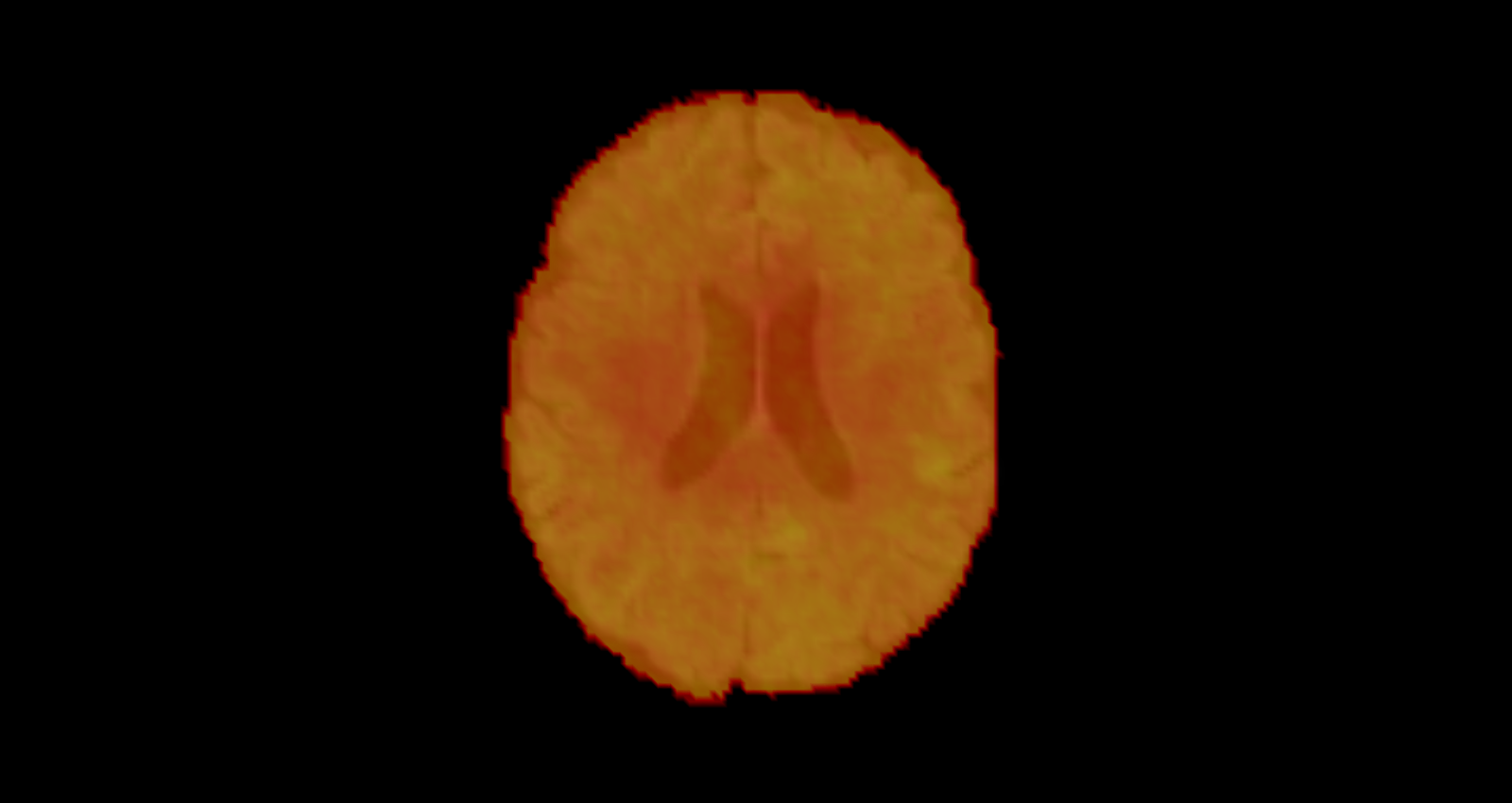}
            \caption*{}
        \end{subfigure}
        \begin{subfigure}{0.3\textwidth}
            \centering
            \includegraphics[trim={18cm 3cm 18cm 3cm},clip,width=\textwidth]{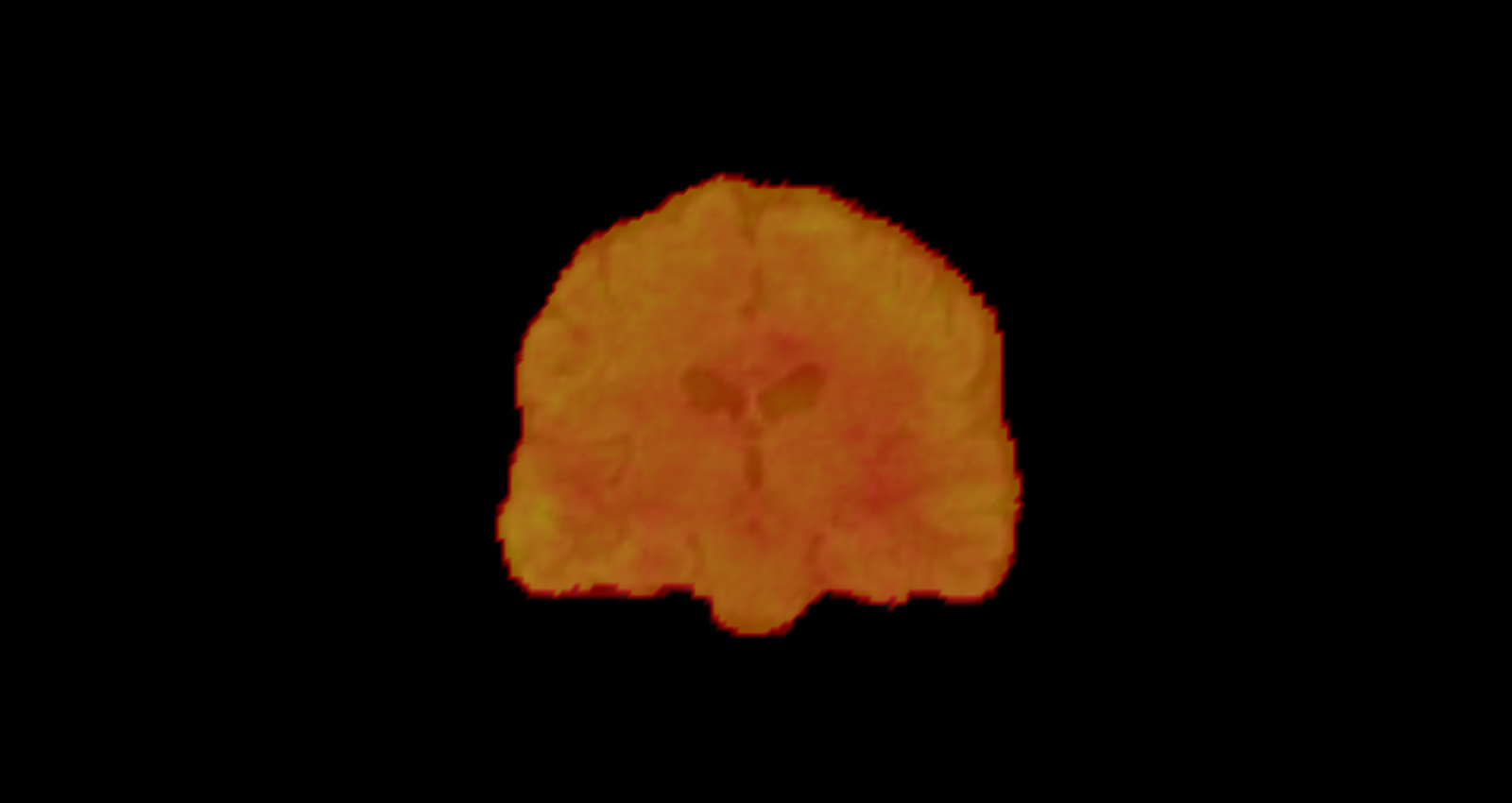}
            \caption*{\gls{VI}}
        \end{subfigure}
        \begin{subfigure}{0.3\textwidth}
            \centering
            \includegraphics[trim={18cm 3cm 18cm 3cm},clip,width=\textwidth]{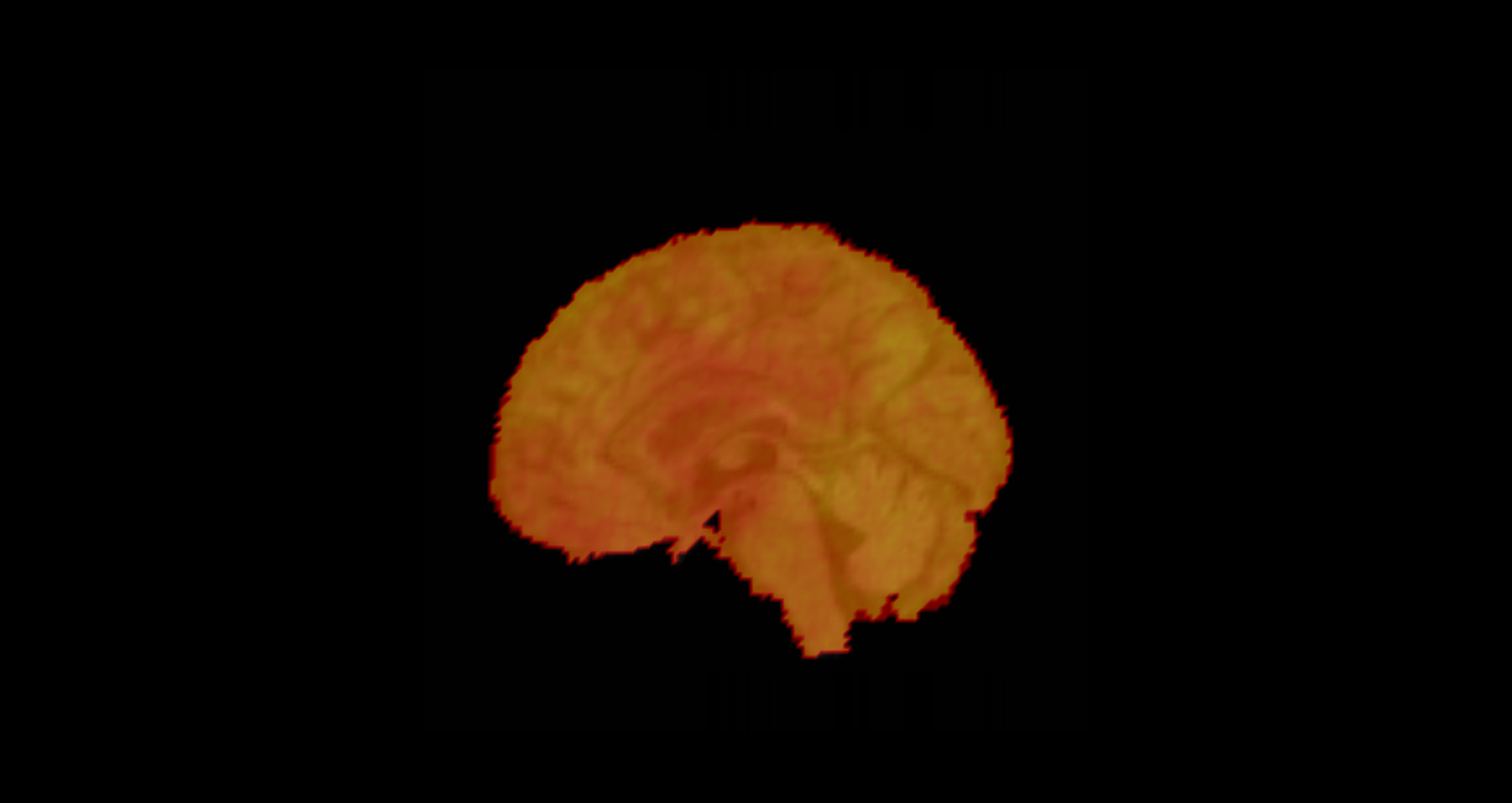}
            \caption*{}
        \end{subfigure}
    \endminipage
    \minipage[c]{0.5\linewidth}
        \begin{subfigure}{0.3\textwidth}
            \centering
            \includegraphics[trim={18cm 3cm 18cm 3cm},clip,width=\textwidth]{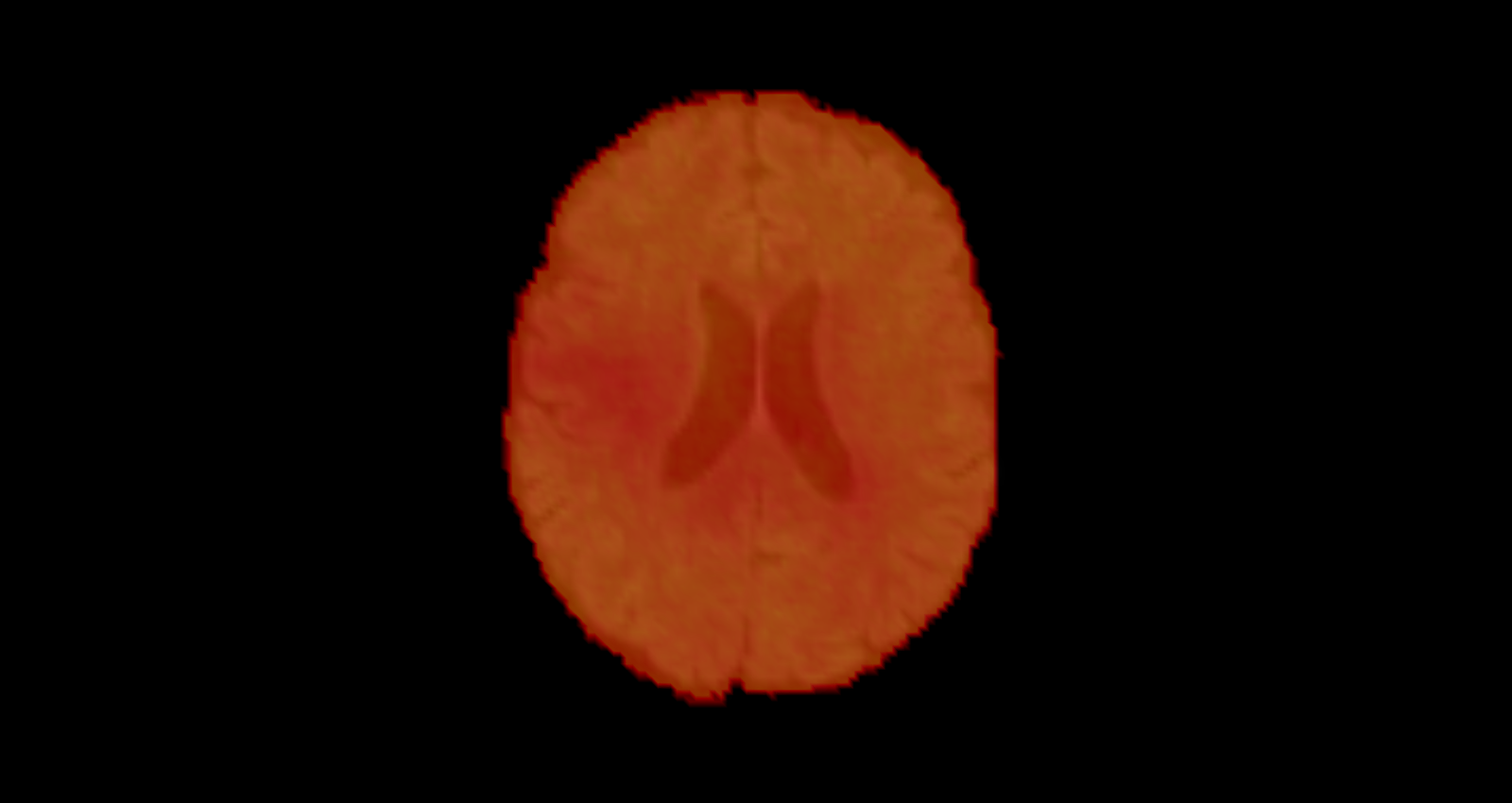}
            \caption*{}
        \end{subfigure}
        \begin{subfigure}{0.3\textwidth}
            \centering
            \includegraphics[trim={18cm 3cm 18cm 3cm},clip,width=\textwidth]{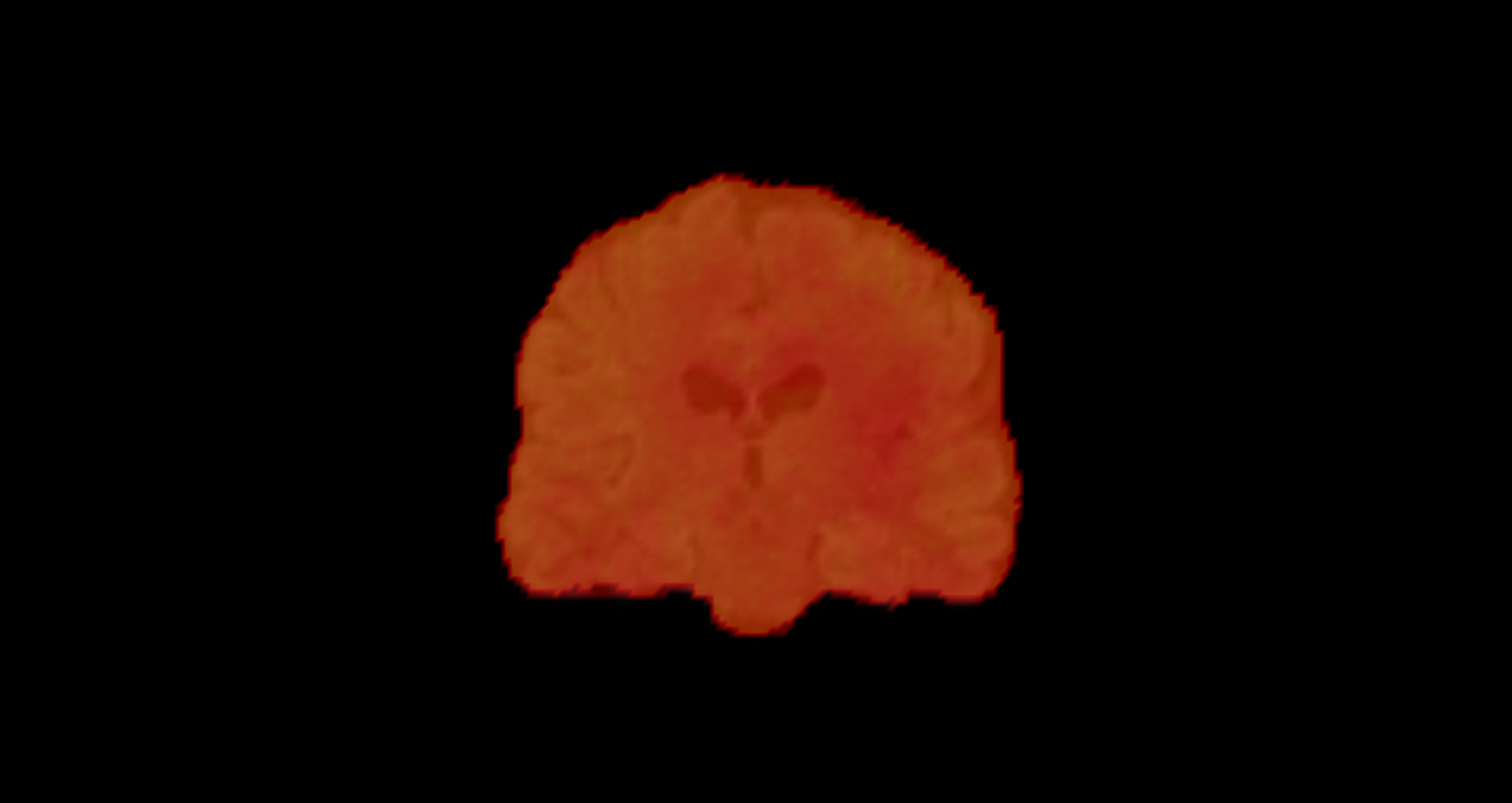}
            \caption*{\gls{SG-MCMC}}
        \end{subfigure}
        \begin{subfigure}{0.3\textwidth}
            \centering
            \includegraphics[trim={18cm 3cm 18cm 3cm},clip,width=\textwidth]{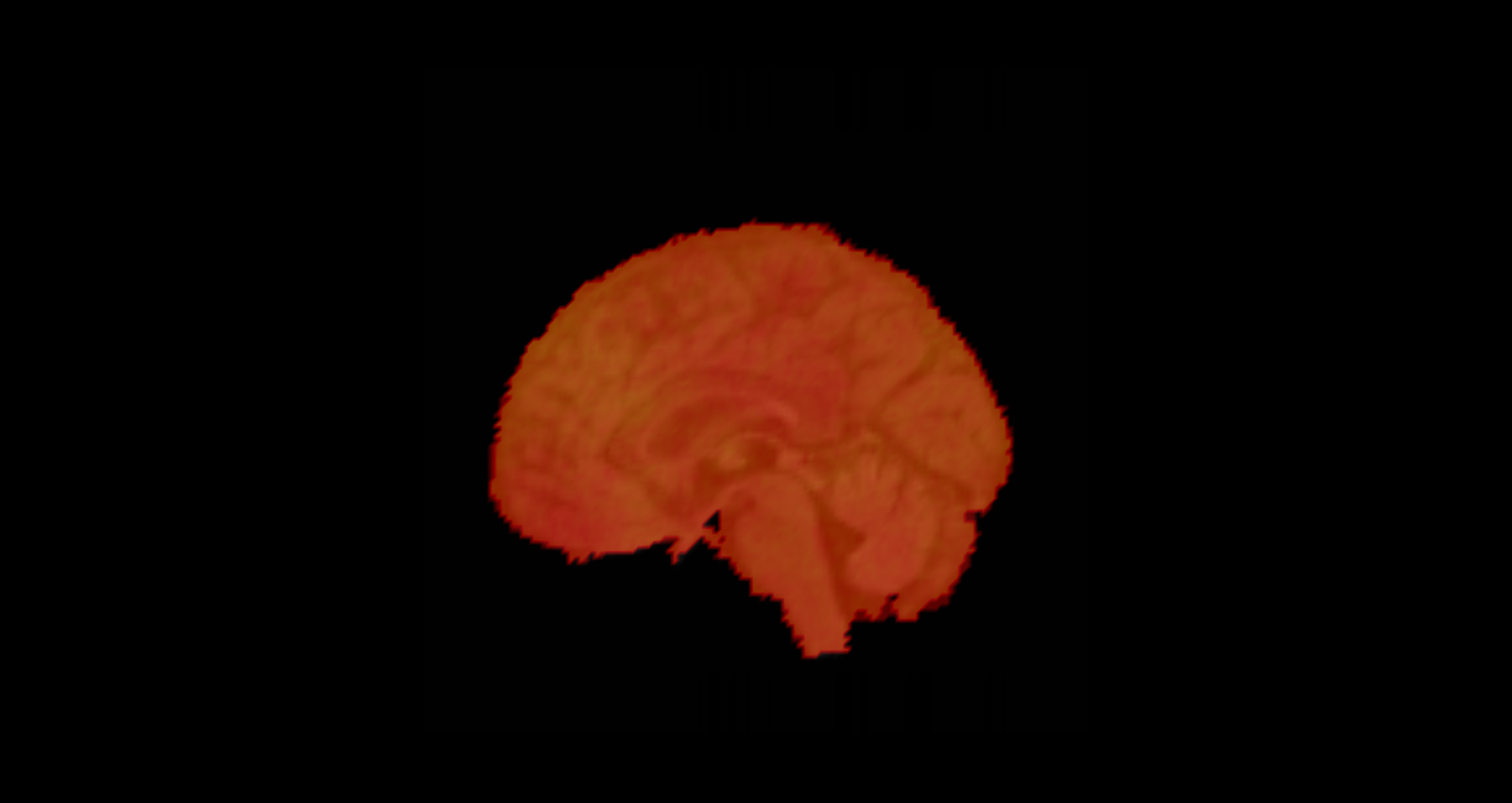}
            \caption*{}
        \end{subfigure}
    \endminipage
    
    \minipage[c]{0.5\linewidth}
        \begin{subfigure}{0.3\textwidth}
            \centering
            \includegraphics[trim={18cm 3cm 18cm 3cm},clip,width=\textwidth]{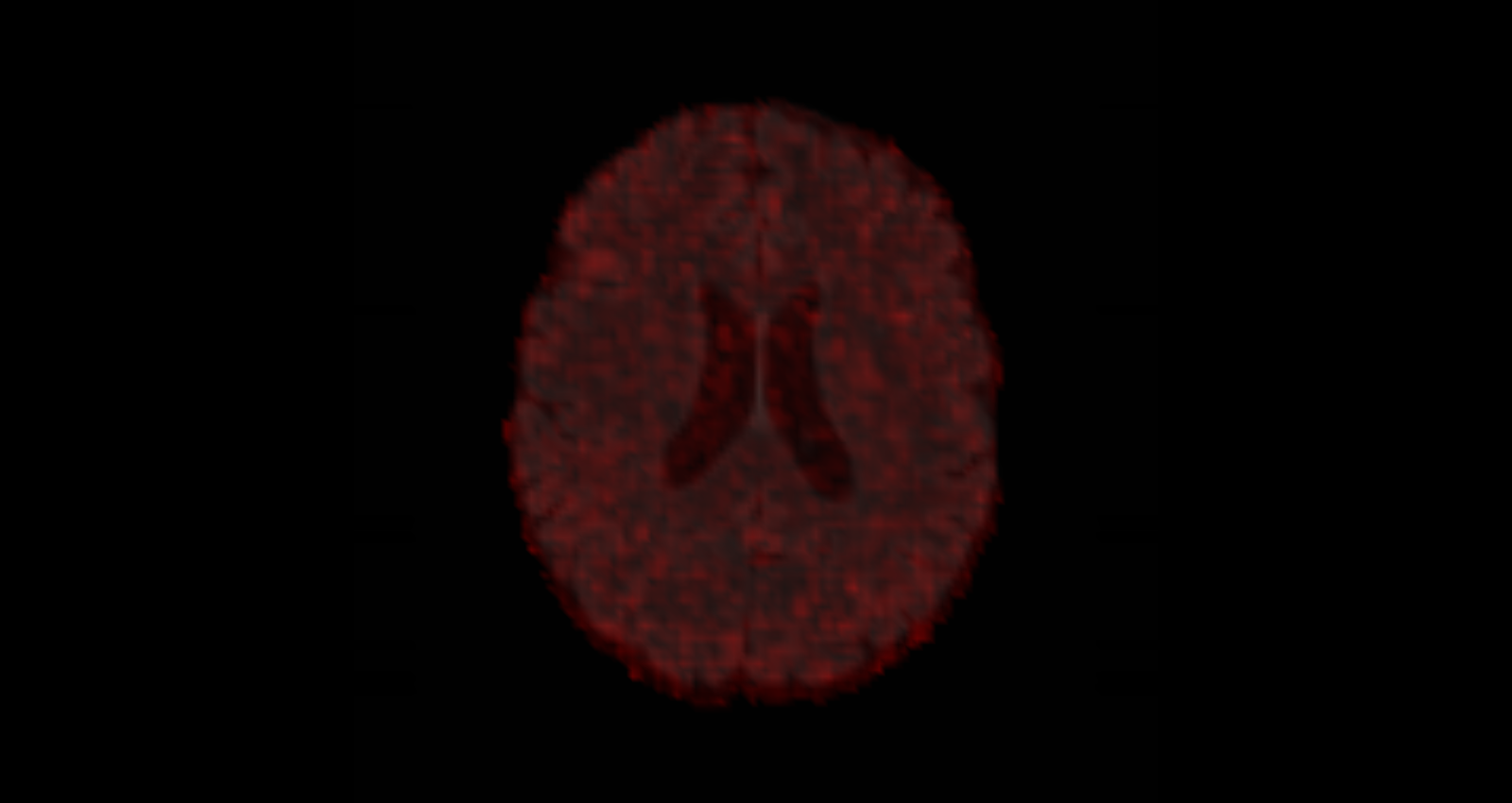}
            \caption*{}
        \end{subfigure}
        \begin{subfigure}{0.3\textwidth}
            \centering
            \includegraphics[trim={18cm 3cm 18cm 3cm},clip,width=\textwidth]{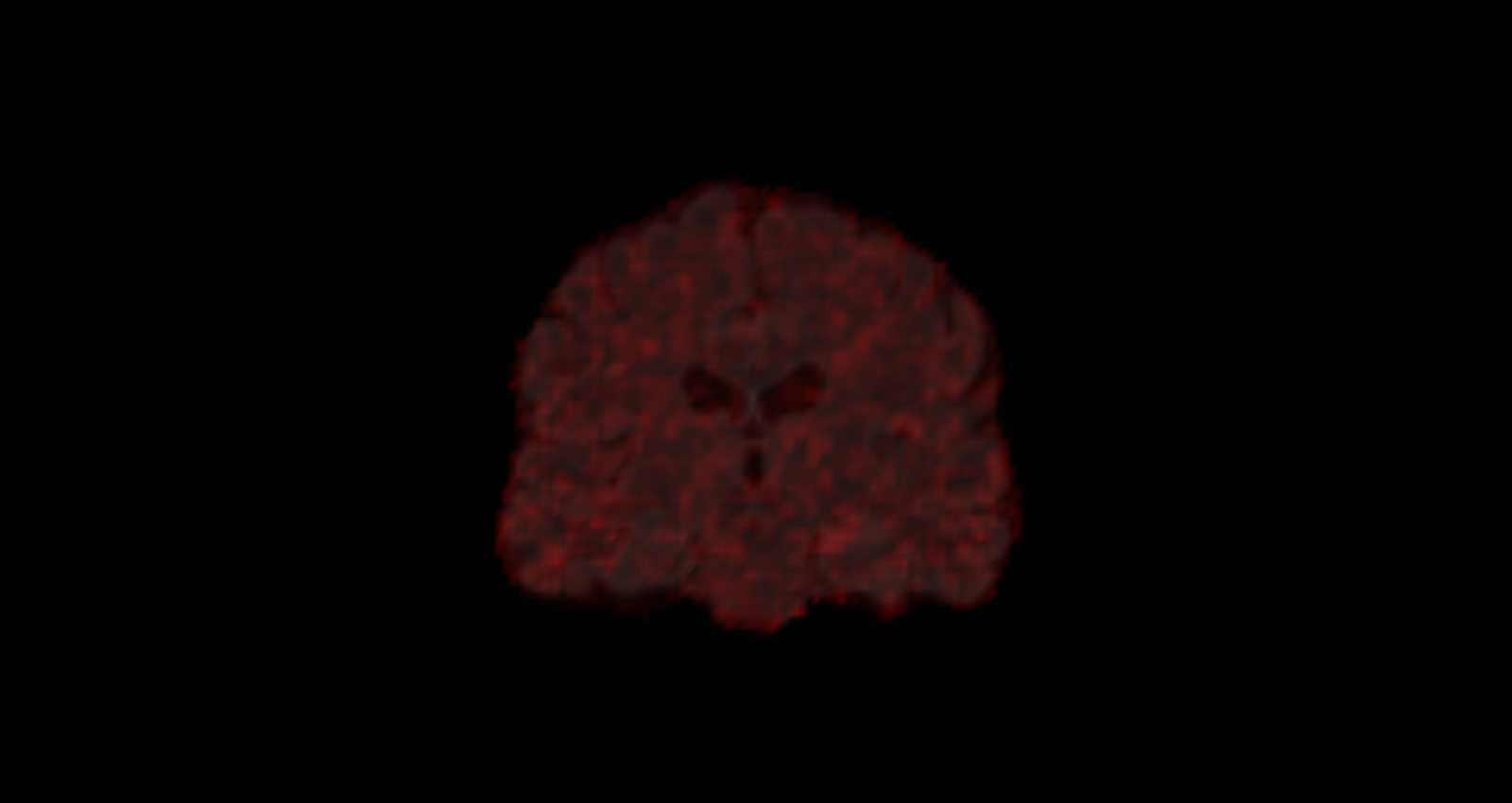}
            \caption*{VoxelMorph}
        \end{subfigure}
        \begin{subfigure}{0.3\textwidth}
            \centering
            \includegraphics[trim={18cm 3cm 18cm 3cm},clip,width=\textwidth]{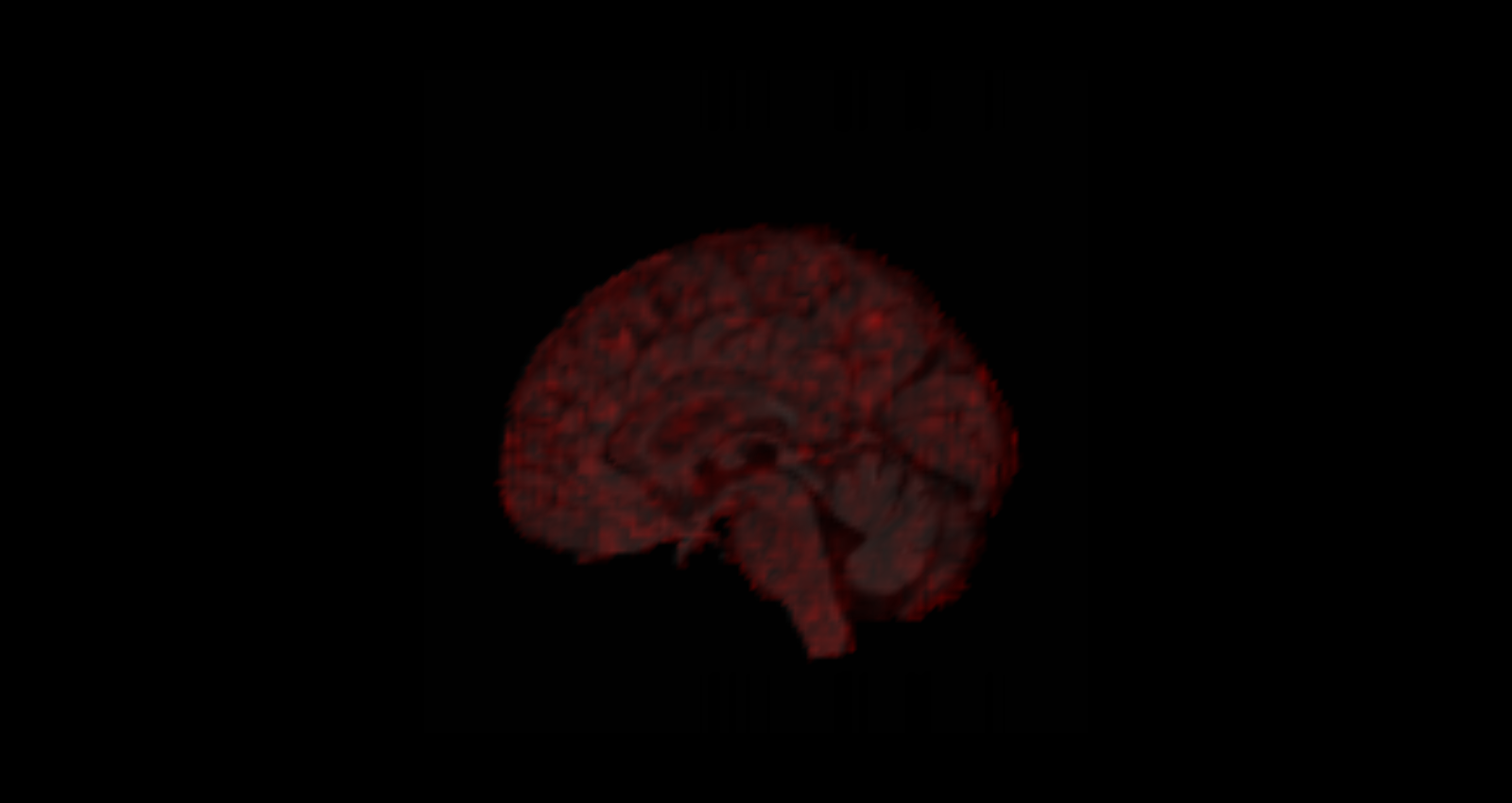}
            \caption*{}
        \end{subfigure}
        \begin{subfigure}{0.02\textwidth}
            \centering
            \includegraphics[trim={36cm 4cm 14cm 4cm},clip,height=3cm]{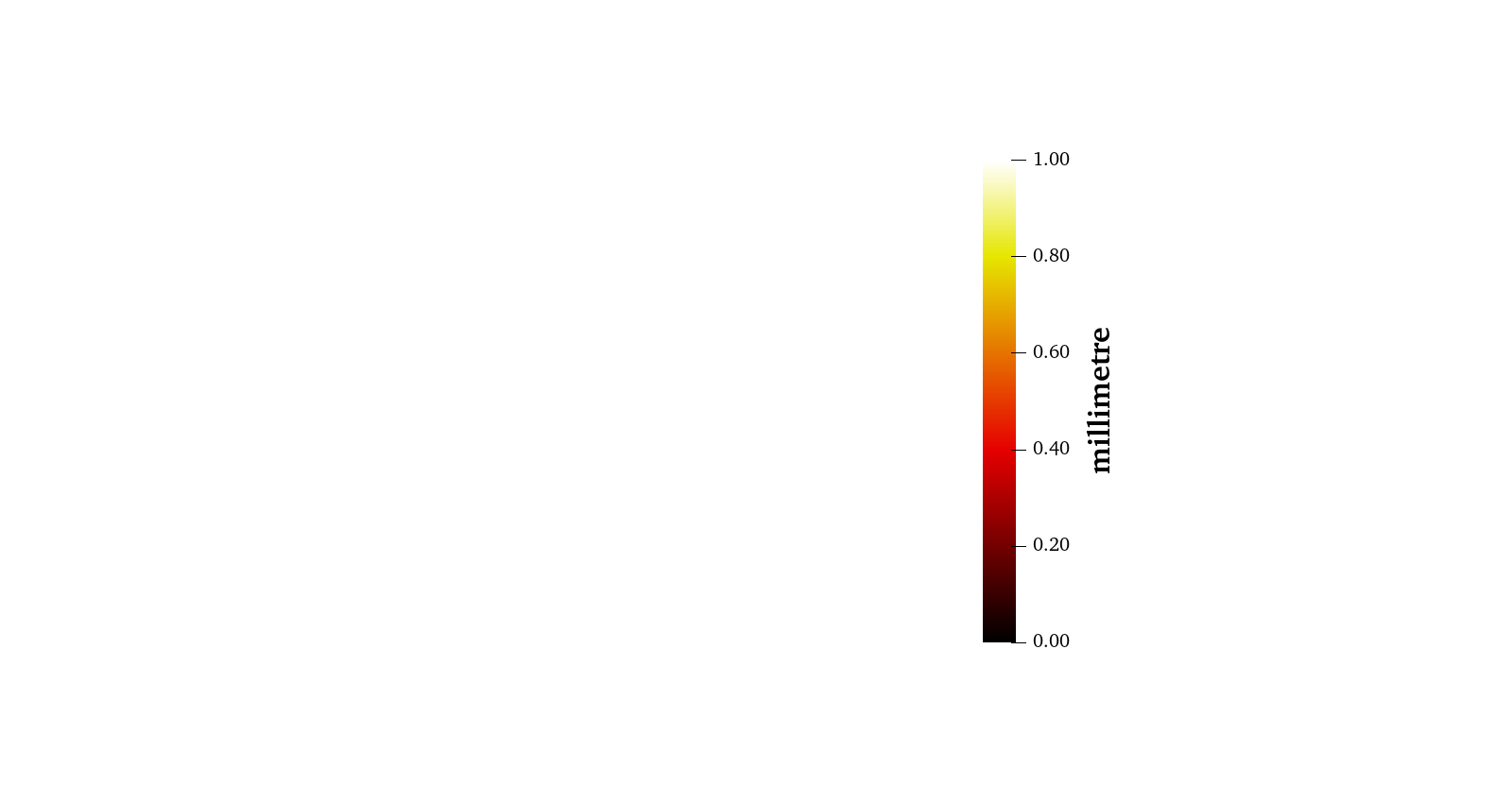}
            \caption*{}
        \end{subfigure}
    \endminipage
    
    \caption{Uncertainty output by the models on the input image pair shown in \Cref{fig:reg}. The standard deviation of the displacement field magnitude is calculated using \nosamples{} samples. %<*fig-caption>
    In case of \gls{SG-MCMC}, samples are selected at regular intervals from the \nosamplestotal{} samples output by each of the two Markov chains, which is needed to prevent autocorrelation between samples. %</fig-caption>
    \gls{SG-MCMC} outputs lower uncertainty than \gls{VI}. The uncertainty estimates output by \gls{VI} and VoxelMorph are very dissimilar, even though the two models assume a similar approximate variational posterior. %<*fig-caption-assumption>
    In case of \gls{SGLD}, visualising standard deviation of the displacement field magnitude is valid under the assumption that the posterior is approximately Gaussian and mono-modal. The standard deviations of displacement field magnitudes sampled from \gls{VI} and \gls{SG-MCMC} are very different, so the visualisation in case of \gls{SG-MCMC} should not be treated as an accurate description of the true uncertainty, but rather as evidence that the true posterior distribution of transformation parameters likely is not Gaussian with a diagonal + low-rank covariance matrix.
    %</fig-caption-assumption>
    }
    \label{fig:UQ}
\end{figure}

\textbf{Qualitative comparison of uncertainty estimates.} The output of \gls{VI}, \gls{SG-MCMC}, and VoxelMorph on a sample image pair is shown in \Cref{fig:UQ}. It should be noted that, due to the fact that direct correspondence between regions is hard to determine, the problem of registering inter-subject brain \gls{MRI} scans is more challenging than problems where non-rigid registration uncertainty had been studied previously, e.g. intra-subject brain \gls{MRI} \citep{Risholm2010, Simpson2012} and intra-subject cardiac \gls{MRI} \citep{LeFolgoc2017}. Nonetheless, the uncertainty estimates output by \gls{VI} and \gls{SG-MCMC} are consistent with previous findings, e.g. higher uncertainty in homogeneous regions \citep{Simpson2012, Dalca2018}. 

Unlike in \cite{LeFolgoc2017}, where Gaussian \glspl{RBF} were used to parametrise the transformation on intra-subject cardiac \gls{MRI} scans, \gls{VI} outputs higher uncertainty than \gls{MCMC}. \gls{SGLD} is known to overestimate the posterior covariance due to non-vanishing learning rates at long times \citep{Mandt2017}, which further suggests that the uncertainty output by \gls{SGLD} might be underestimated. The uncertainty estimates produced by \gls{VI} and \gls{SG-MCMC} are consistent but different in magnitude, while those output by VoxelMorph are noticeably different, with the values much smaller. This indicates the need for further research into calibration of uncertainty estimates for image registration methods based on deep learning.

\begin{figure}[!htb]
    \centering
    \captionof{table}{Mean and standard deviation of \gls{ASD} and \gls{DSC} on small, medium, and large subcortical structures, calculated using \nosamples{} output samples. Large structures include the brain stem and thalamus, medium structures---the caudate, hippocampus, and putamen, and small structures---the accumbens, amygdala, and pallidum. The values for the best performing model are underlined.}
    \label{tab:metrics-table}
    
    \bigskip
    \begin{tabular}{lrrrrrr}
        \toprule
                        & \multicolumn{3}{c}{\thead{average surface distance\\(mm)}}                                    & \multicolumn{3}{c}{Dice score}                                                                    \\
        \textbf{model} & \thead{small\\structures}     & \thead{medium\\structures}    & \thead{large\\structures}     & \thead{small\\structures}         & \thead{medium\\structures}    & \thead{large\\structures}     \\
        \midrule
        \gls{VI}        & $1.12$ $(0.16)$               & $0.97$ $(0.08)$               & $\underline{1.04}$ $(0.26)$   & $0.68$ $(0.07)$                   & $0.79$ $(0.03)$               & $\underline{0.88}$ $(0.02)$   \\
        \gls{SG-MCMC}      & $1.10$ $(0.15)$               & $\underline{0.95}$ $(0.07)$               & $1.05$ $(0.24)$   & $0.68$ $(0.07)$                   & $0.79$ $(0.03)$               & $\underline{0.88}$ $(0.02)$   \\
        VoxelMorph       & $\underline{1.06}$ $(0.14)$   & $0.96$ $(0.09)$   & $1.05$ $(0.11)$               & $\underline{0.70}$ $(0.05)$       & $\underline{0.80}$ $(0.02)$   & $0.87$ $(0.01)$   \\
        \bottomrule
    \end{tabular}
    
    \bigskip
    \includegraphics[width=\textwidth]{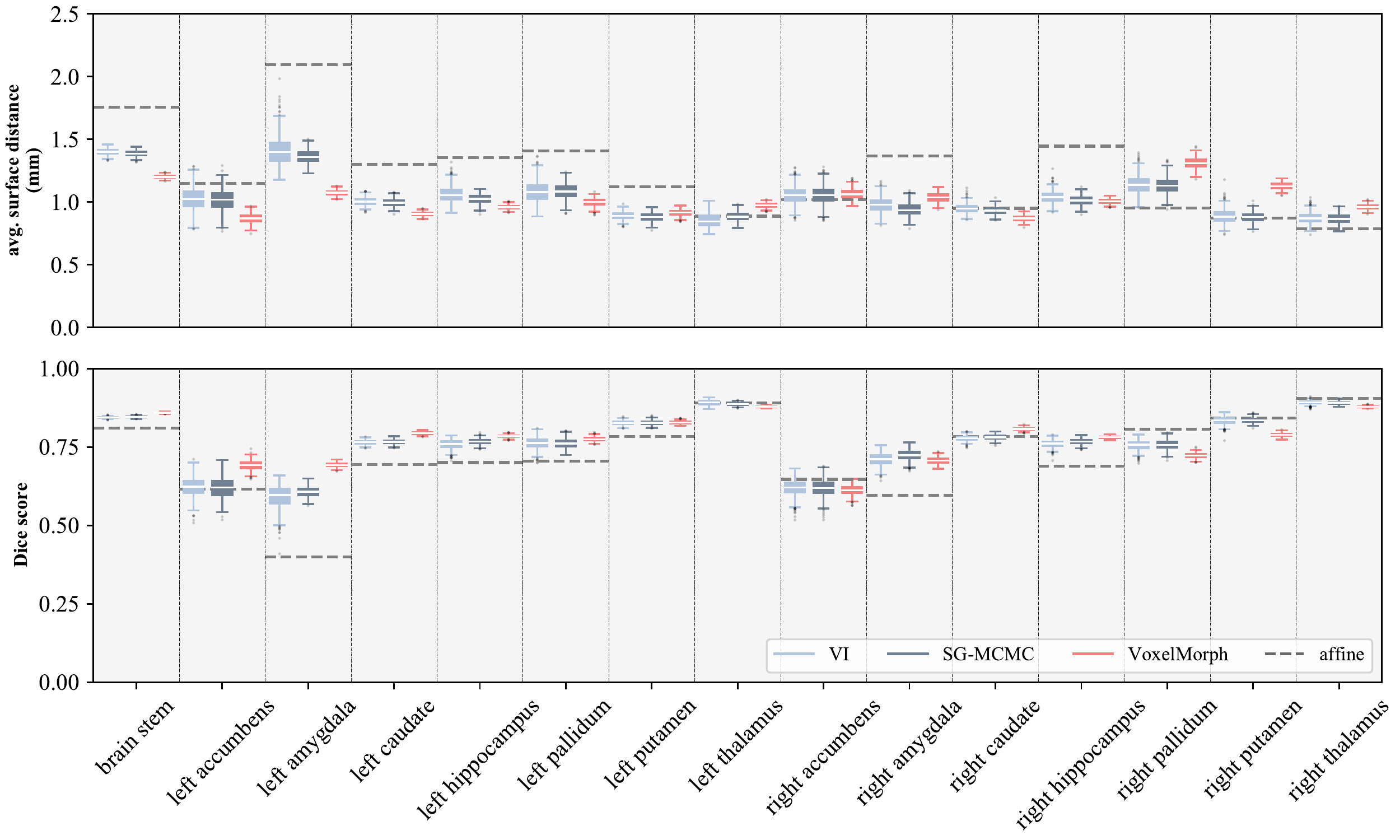}
    \captionof{figure}{\gls{ASD} and \gls{DSC} on each subcortical structure. Dashed lines show the metric values prior to non-rigid registration. The label uncertainty of \gls{VI} and \gls{SG-MCMC} is comparable, despite different transformation uncertainty.}
    \label{fig:metrics-boxplot}
\end{figure}

\textbf{Image registration accuracy.} To evaluate image registration accuracy, we calculate \glspl{ASD} and \glspl{DSC} on the subcortical structure segmentations using the fixed and the moving segmentation warped with transformations sampled from the models. The metric comparison between our model and VoxelMorph on the image pair in \Cref{fig:UQ} is shown in \Cref{tab:metrics-table} for segmentations grouped by volume and in \Cref{fig:metrics-boxplot} for individual segmentations.
The metrics show significant improvement over affine registration on most subcortical structures. The differences between label uncertainties are less pronounced than between transformation uncertainties. \glspl{ASD} and \glspl{DSC} are generally marginally better when using samples from \gls{SG-MCMC} than from the approximate variational posterior. Better accuracy in case of \gls{SGLD} is expected, given restrictive assumptions of the approximate variational model. \gls{VI}, \gls{SG-MCMC}, and VoxelMorph produce similar accuracy, despite different uncertainty estimates.

%<*disp-vs-label-uncertainty>
\textbf{Quantitative comparison of uncertainty estimates.} It is difficult to define ground truth uncertainty with regards to image registration. \cite{Luo2019} suggested that well-calibrated image registration uncertainty estimates need to be informative of anatomical features, which are important in neurosurgery. To compare the uncertainty output by different models quantitatively and evaluate whether the uncertainty estimates are clinically useful, we adopt a method similar to that used by \cite{Luo2019} and calculate the Pearson correlation coefficient $r_{u_d u_l}$ between the displacement uncertainties $u_d$ and label uncertainties $u_l$. Here we define label uncertainty $u_l$ to be the standard deviation of the \gls{DSC} for each subcortical structure, and displacement uncertainty $u_d$ to be the voxel-wise mean of the displacement field standard deviation magnitude within the region that corresponds to a given subcortical structure. %</disp-vs-label-uncertainty>

\begin{table}[!htb]
    \centering
    \caption{Pearson correlation coefficient $r_{u_d u_l}$ between the displacement uncertainties $u_d$ and label uncertainties $u_l$. We define label uncertainty $u_l$ to be the standard deviation of the \gls{DSC} for each subcortical structure, and displacement uncertainty $u_d$ to be the voxel-wise mean of the displacement field standard deviation magnitude within the region which corresponds to a given subcortical structure. The value of the correlation coefficient is highest in case of VoxelMorph but qualitative evidence suggests that the uncertainty estimates output by the model are not well calibrated. The correlation coefficient is also higher for \gls{SG-MCMC} than for \gls{VI}.}
    \label{tab:corr}
    
    \begin{tabular}{lr}
        \toprule
        \textbf{model}      & $r_{u_d u_l}$         \\
        \midrule
        \gls{VI}            & $0.07$                \\
        \gls{SG-MCMC}       & $0.10$                \\
        VoxelMorph          & $0.12$                \\
        \bottomrule
    \end{tabular}
\end{table}

%<*disp-vs-label-uncertainty2>
The correlation coefficient is highest in case of VoxelMorph but qualitative evidence suggests that image registration uncertainty output by the model is not well calibrated. The displacement field uncertainty is more informative of label uncertainty in case of \gls{SG-MCMC} than \gls{VI}, so the uncertainty estimates output by \gls{SG-MCMC} are likely to be more useful for clinical purposes than those output by \gls{VI} or VoxelMorph.
%</disp-vs-label-uncertainty2>

\begin{figure}[htb!]
    \centering
    \captionof{table}{Mean and standard deviation of the number and percentage of voxels where the sampled transformation is not diffeomorphic as defined in \Cref{sec:setup}. The values are based on \nosamples{} samples.}
    \label{tab:diff}
    
    \begin{tabular}{lrr}
        \toprule
        \textbf{model}           & $\vert \det J_{\varphi^{-1}} \vert \leq 0$   & \% ($\times 10^{-6}$) \\
        \midrule
        \gls{VI}                  & $0.00$ $(0.04)$                              & $0.0$ $(0.2)$         \\
        \gls{SG-MCMC}                & $0.00$ $(0.00)$                              & $0.0$ $(0.0)$         \\
        VoxelMorph                 & $0.00$ $(0.00)$                              & $0.0$ $(0.0)$         \\
        \bottomrule
    \end{tabular}
    
    \vspace*{0.5cm}
    \begin{minipage}[c]{0.175\textwidth}
        \centering
        \includegraphics[trim={17cm 4cm 17cm 4cm},clip,width=\textwidth]{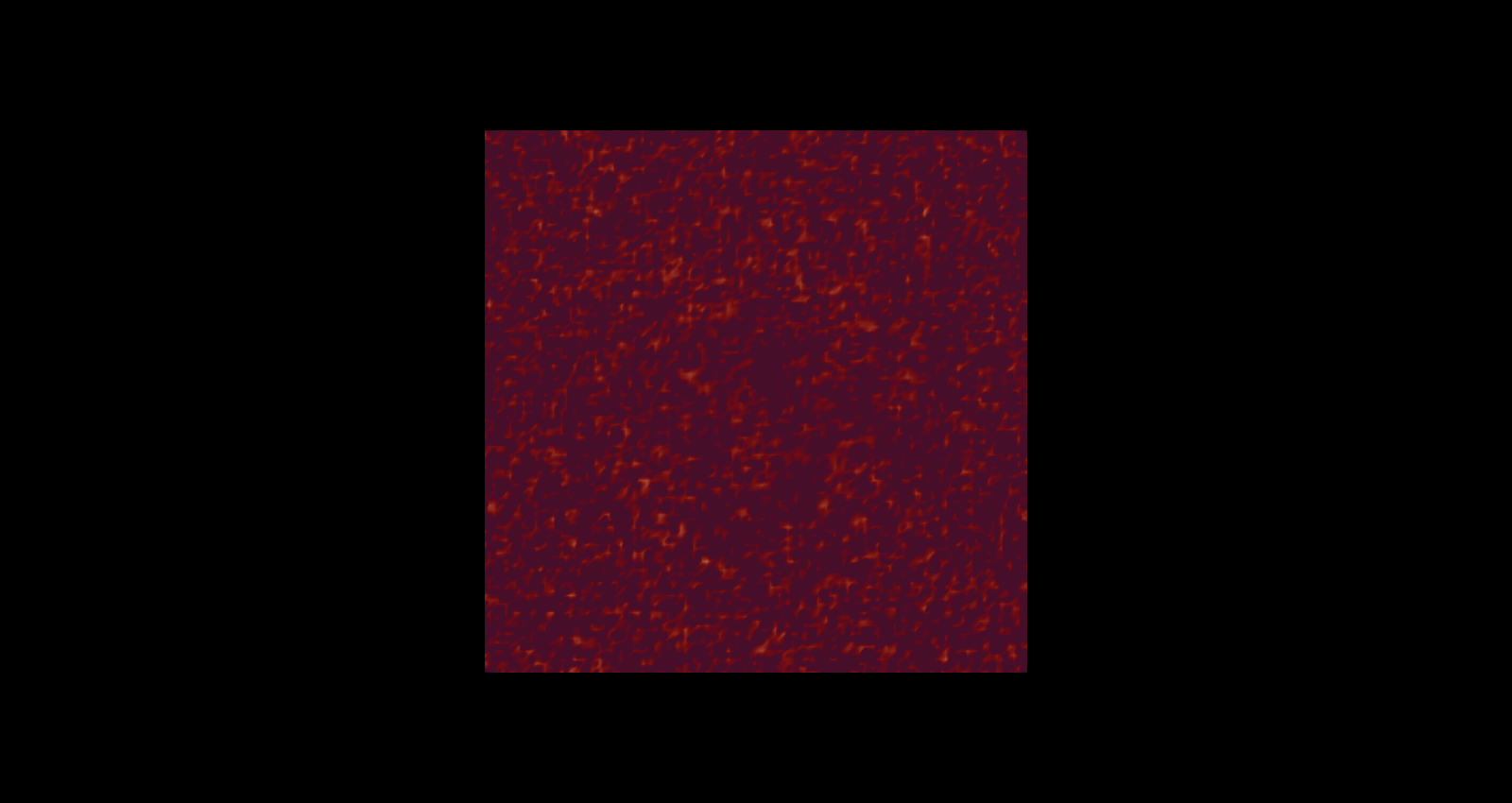}
        \caption*{\gls{VI}}
    \end{minipage}
    \begin{minipage}[c]{0.175\textwidth}
        \centering
        \includegraphics[trim={17cm 4cm 17cm 4cm},clip,width=\textwidth]{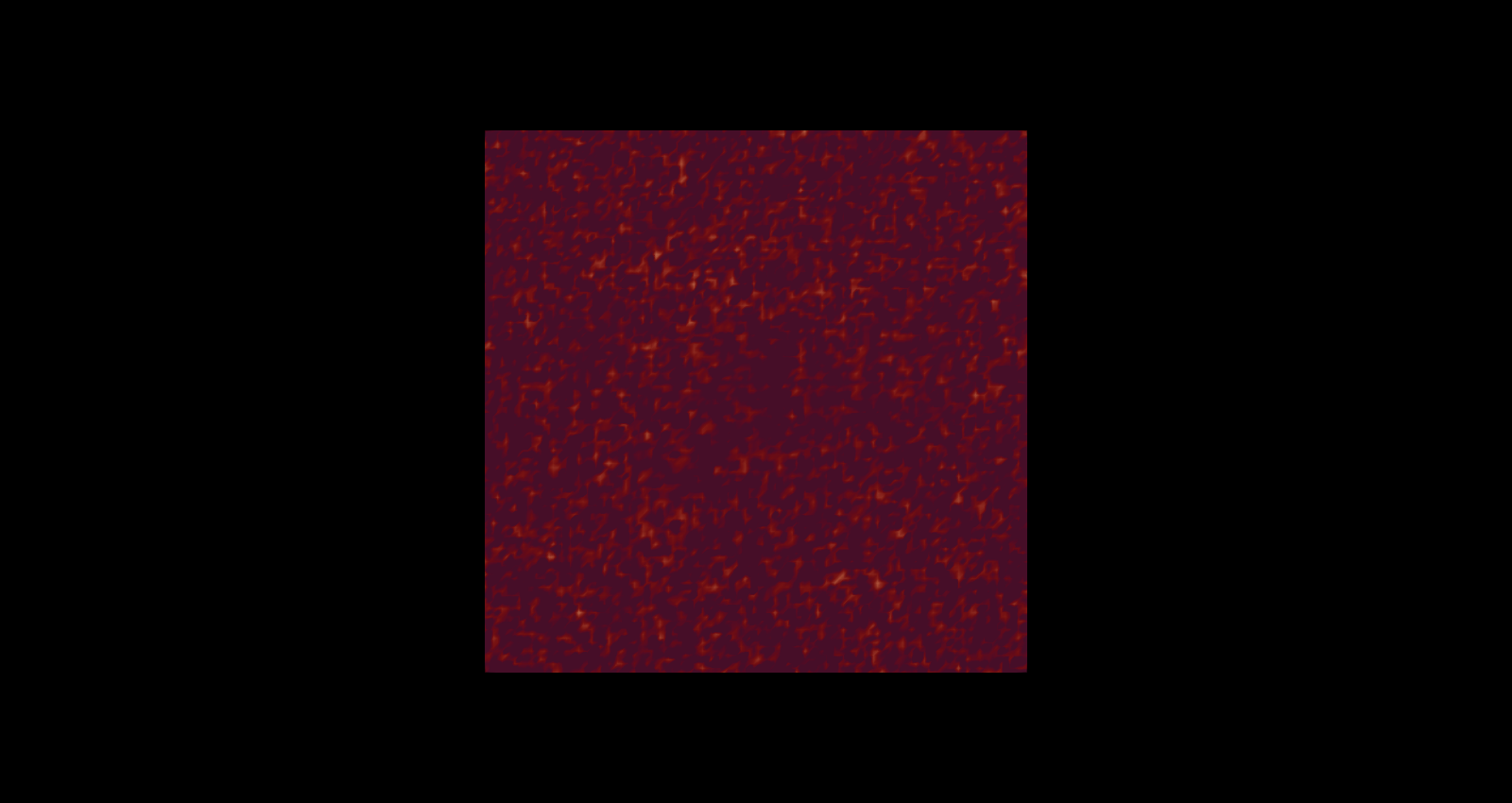}
        \caption*{\gls{MCMC}}
    \end{minipage}
    \begin{minipage}[c]{0.175\textwidth}
        \centering
        \includegraphics[trim={17cm 4cm 17cm 4cm},clip,width=\textwidth]{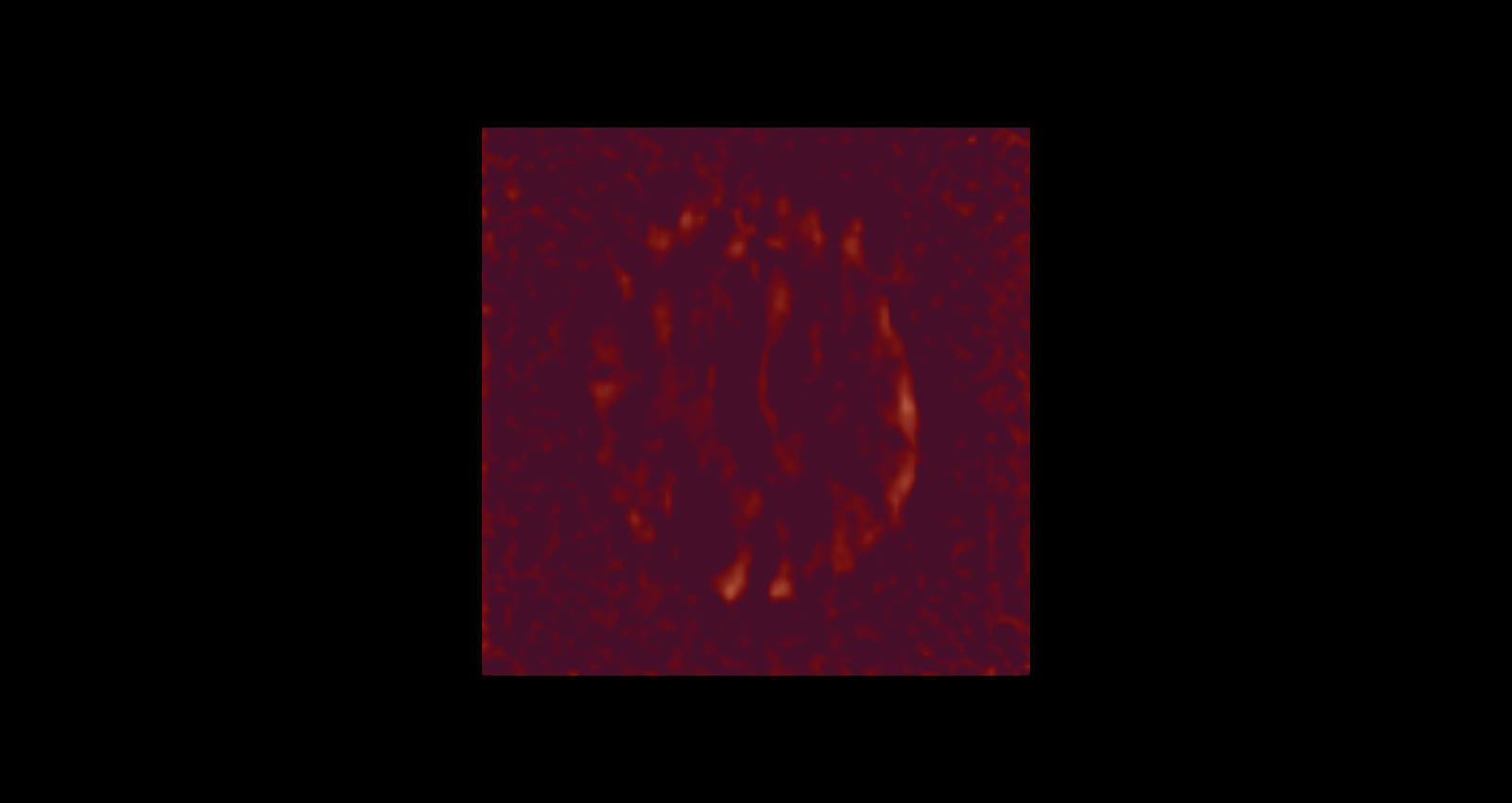}
        \caption*{VoxelMorph}
    \end{minipage}
    \begin{minipage}[c]{0.1\textwidth}
        \centering
        \includegraphics[trim={36cm 4cm 14cm 4cm},clip,height=3.5cm]{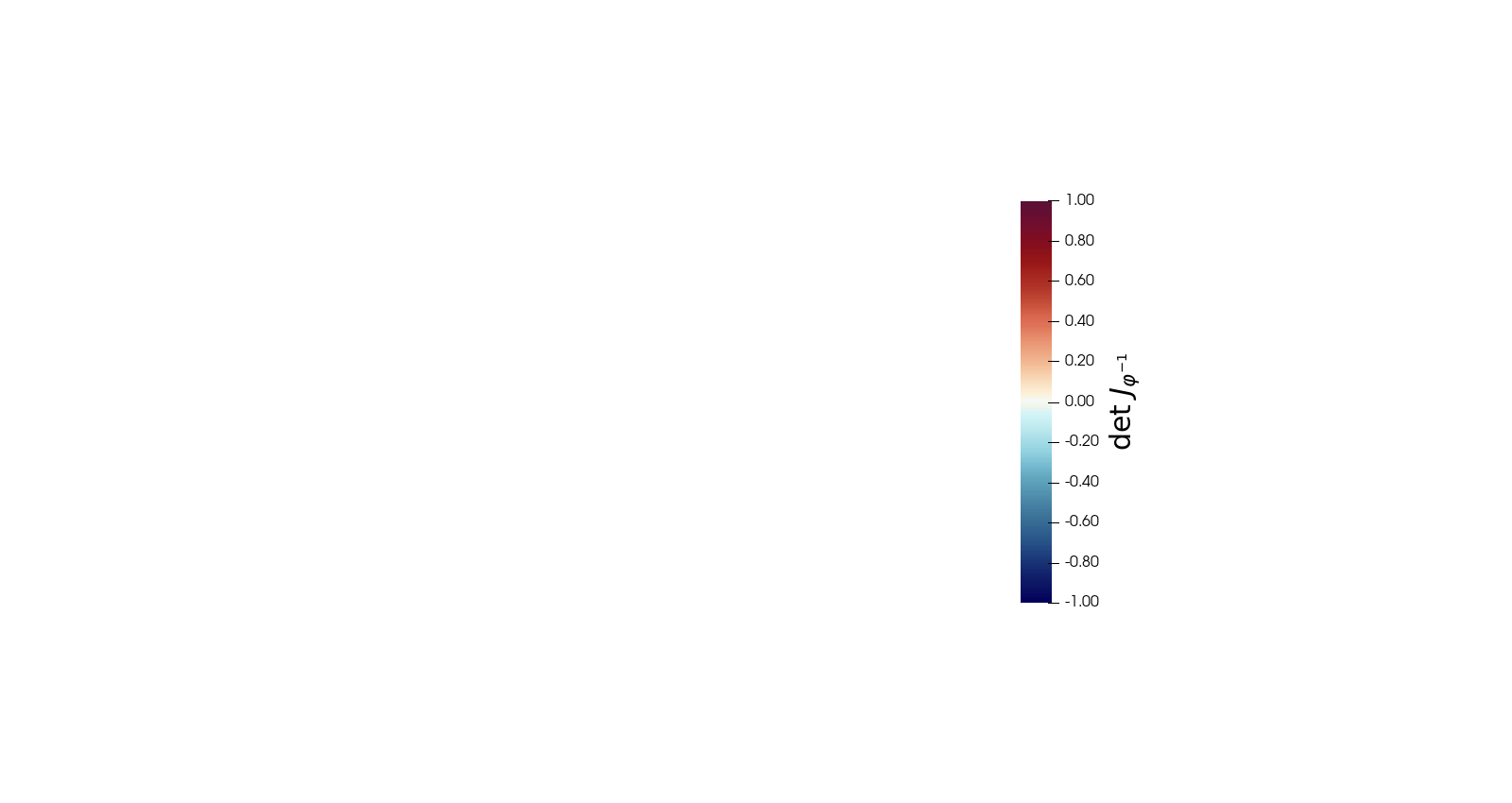}
        \caption*{}
    \end{minipage}
    
    \captionof{figure}{Jacobian determinant of a sample transformation from each model. Even though the output transformations are diffeomorphic, they are not convincingly smooth due to parametrisation of the approximate variational posterior of transformation parameters, whose covariance matrix is diagonal + low-rank in case of \gls{VI} and diagonal in case of VoxelMorph. Middle slice of a 3D image in the axial plane.}
    \label{fig:diff}
\end{figure}

\textbf{Transformation smoothness.} Finally, in order to evaluate the quality of the model output, in \Cref{tab:diff} we report the number of voxels where the sampled transformations are not diffeomorphic. Each model produces transformations where the number of non-positive Jacobian determinants is nearly zero. \gls{SG-MCMC} slightly reduced the number of non-positive Jacobian determinants compared to \gls{VI}. The transformations output by VoxelMorph appear more smooth, which is directly related to the lower transformation uncertainty shown in \Cref{fig:UQ}.

\subsubsection{Comparison of the output of SG-MCMC for different initialisations}
\label{sec:exp4}

To study the potential impact of initialisation on the output uncertainty estimates, we analyse the transformations sampled from \gls{SGLD} run with different initial velocity fields $w_0$. We experiment with a sample $w_0 \sim \mathcal{N} \left( \mu_w, \Sigma_w \right)$ from \gls{VI}, a zero velocity field which corresponds to an identity transformation, and a random velocity field $w_0 \sim \mathcal{N} \left( 0, I_{3N^3} \right)$ sampled from a standard multivariate normal distribution. The first \nosamplesburnin{} samples from each chain are discarded to allow \gls{MCMC} to reach the stationary distribution, and the two Markov chains are run for \nosamplestotal{} transitions each, from which we extract \nosamples{} total samples at regular intervals.

We observed no strong dependence of the result on the initialisation. The uncertainty values are visually identical to those in \Cref{fig:UQ}. The mean voxel-wise discrepancy between the magnitudes of uncertainty estimates is approximately \SI{0.1}{\milli\metre}, while the maximum is approximately \SI{0.2}{\milli\metre}. This suggests that the Markov chains mix well.

\subsubsection{Comparison of the output of SG-MCMC for non-parametric \glsfmtshortpl{SVF} and \glsfmtshortpl{SVF} based on B-splines}
\label{sec:exp5}

One of the common features of recent state-of-the-art models for non-rigid registration that enable uncertainty quantification, e.g.  \cite{Dalca2018} and \cite{Krebs2019}, is an \gls{SVF} transformation parametrisation as well as a diagonal covariance matrix of the approximate variational posterior of transformation parameters, which ignores spatial correlations. We showed in \Cref{sec:exp3} that this assumption can lead to diffeomorphic transformations that are not smooth even in case of image registration that is not based on deep learning. Furthermore, previous work on uncertainty quantification in image registration made the assumption of independence between control points but not between directions and used transformation parametrisations that guaranteed smoothness in an implicit manner, e.g. B-splines \citep{Simpson2012} or Gaussian \glspl{RBF} \citep{LeFolgoc2017}.

To better understand the impact of transformation parametrisations on uncertainty quantification, we analyse the output transformations and uncertainty estimates when using sparse \glspl{SVF} based on cubic B-splines. The control point spacing is set to two or four voxels along each dimension, which gives \SI{3.64}{\milli\metre} or \SI{7.28}{\milli\metre}. The \gls{SGLD} step size is set to $5 \times 10^{-2}$, using the heuristic in \Cref{sec:UQ}.

\begin{figure}[htb!]
    \centering
    \captionsetup{type=figure}
    \begin{minipage}[c]{0.175\textwidth}
        \centering
        \includegraphics[trim={18cm 3cm 18cm 3cm},clip,width=\textwidth]{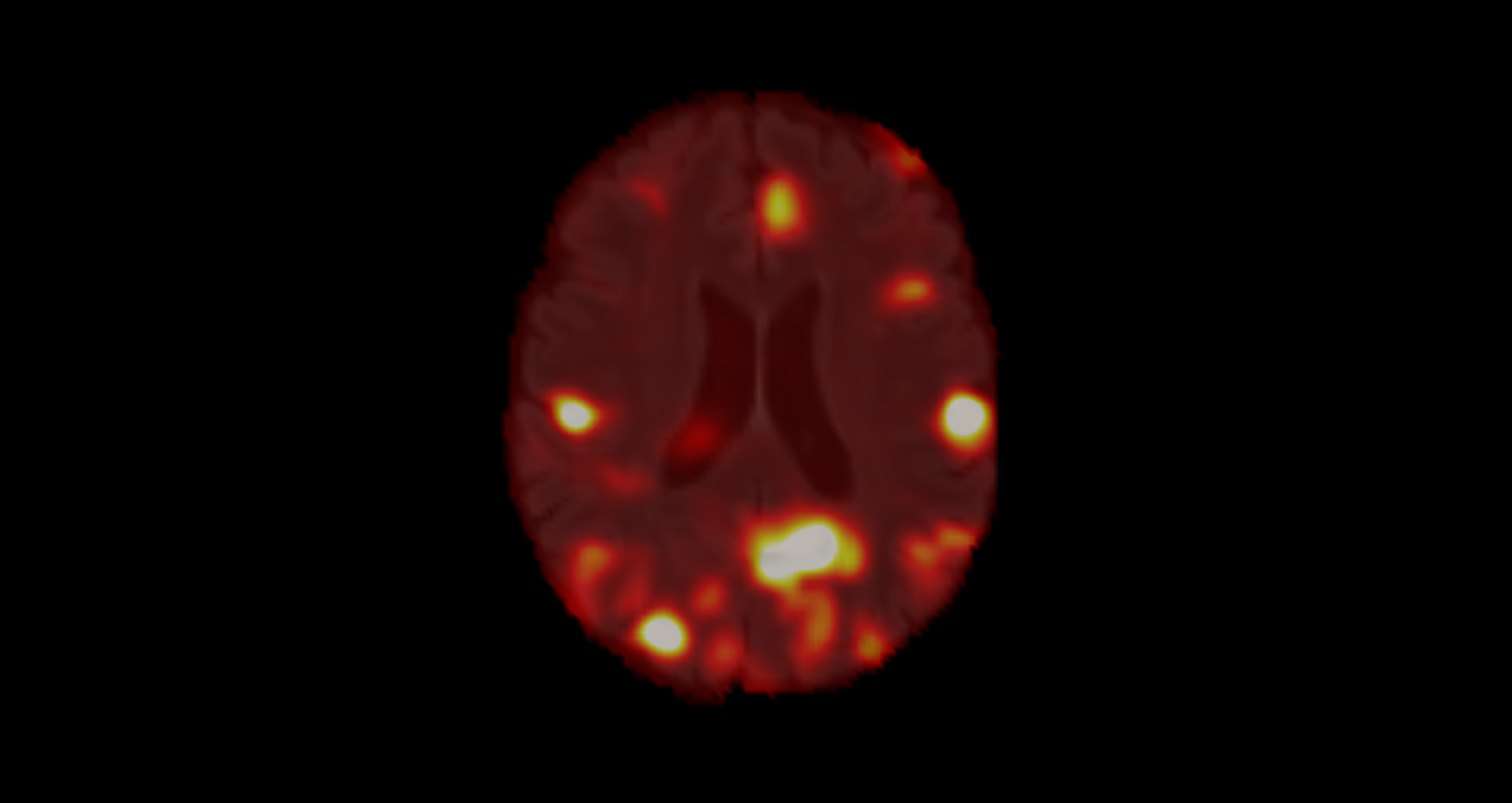}
    \end{minipage}
    \begin{minipage}[c]{0.175\textwidth}
        \centering
        \includegraphics[trim={18cm 3cm 18cm 3cm},clip,width=\textwidth]{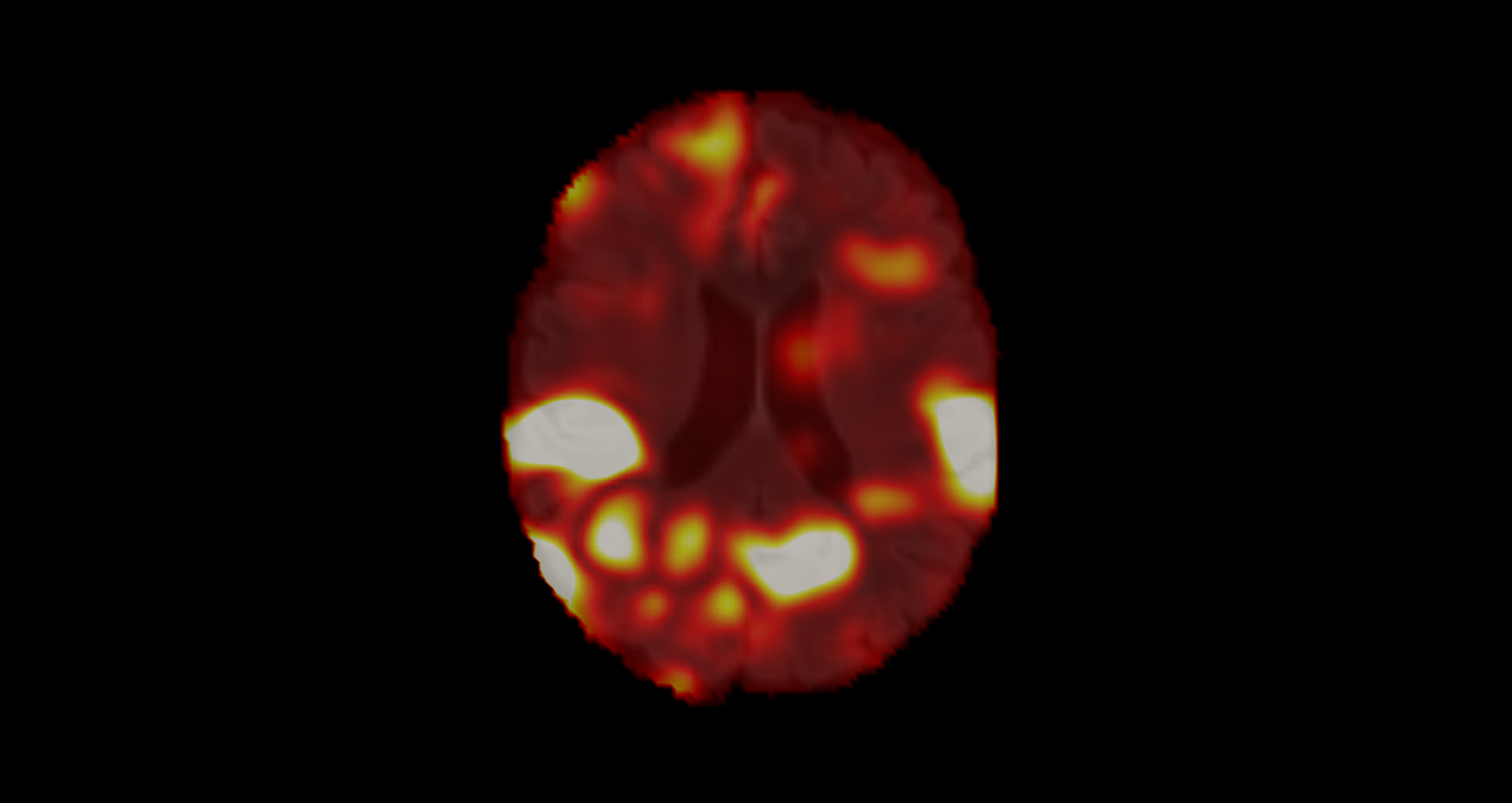}
    \end{minipage}
    \begin{minipage}[c]{0.1\textwidth}
        \centering
        \includegraphics[trim={36cm 4cm 14cm 4cm},clip,height=3cm]{figures/3/colorbar_uncertainty.png}
    \end{minipage}
    
    \begin{minipage}[c]{0.175\textwidth}
        \centering
        \subcaptionbox{$\delta = 2$
        \label{Jac-2}}[\linewidth]{\includegraphics[trim={17cm 4cm 17cm 4cm},clip,width=\textwidth]{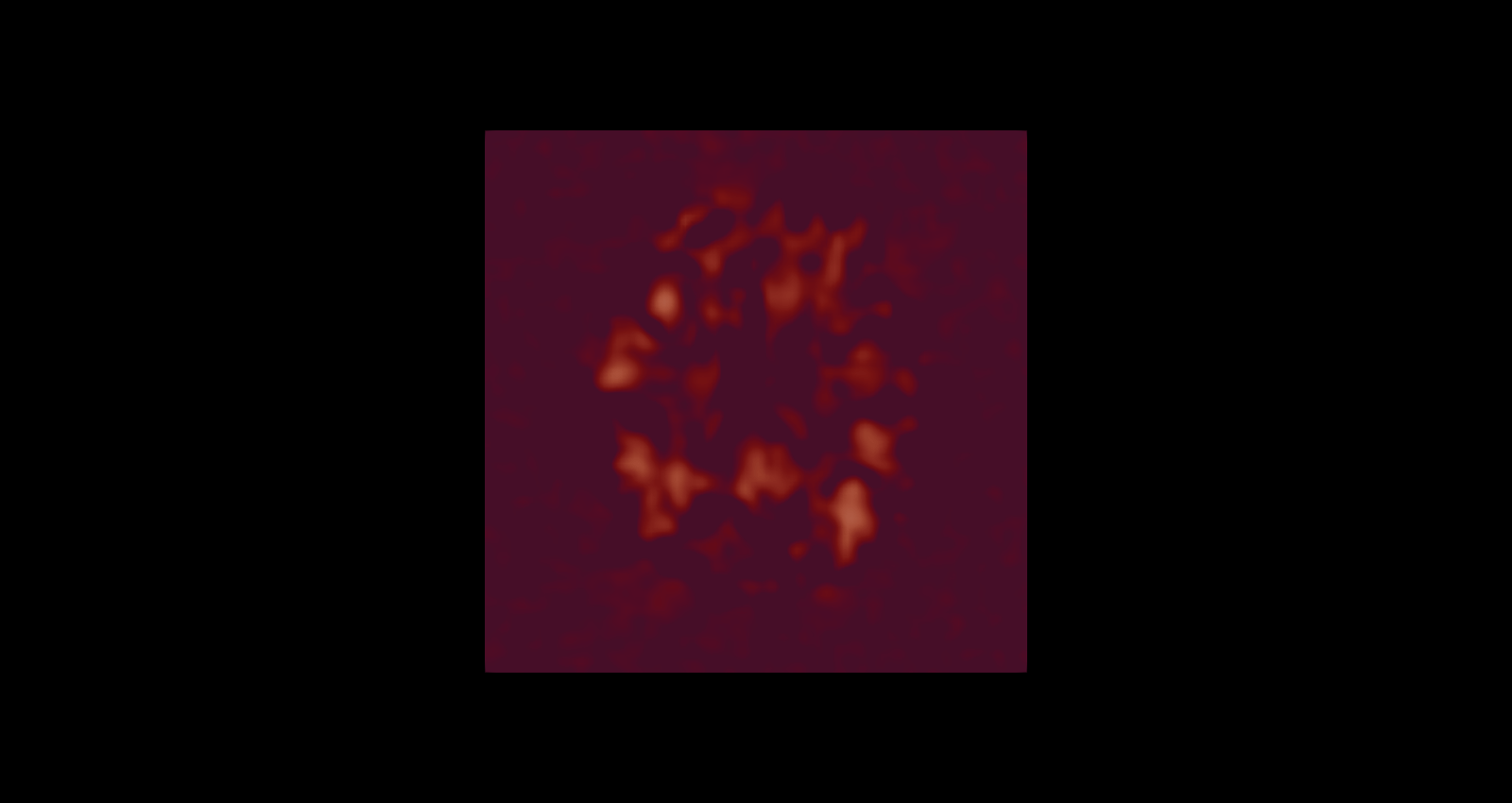}}
    \end{minipage}
    \begin{minipage}[c]{0.175\textwidth}
        \centering
        \subcaptionbox{$\delta = 4$
        \label{Jac-4}}[\linewidth]{\includegraphics[trim={17cm 4cm 17cm 4cm},clip,width=\textwidth]{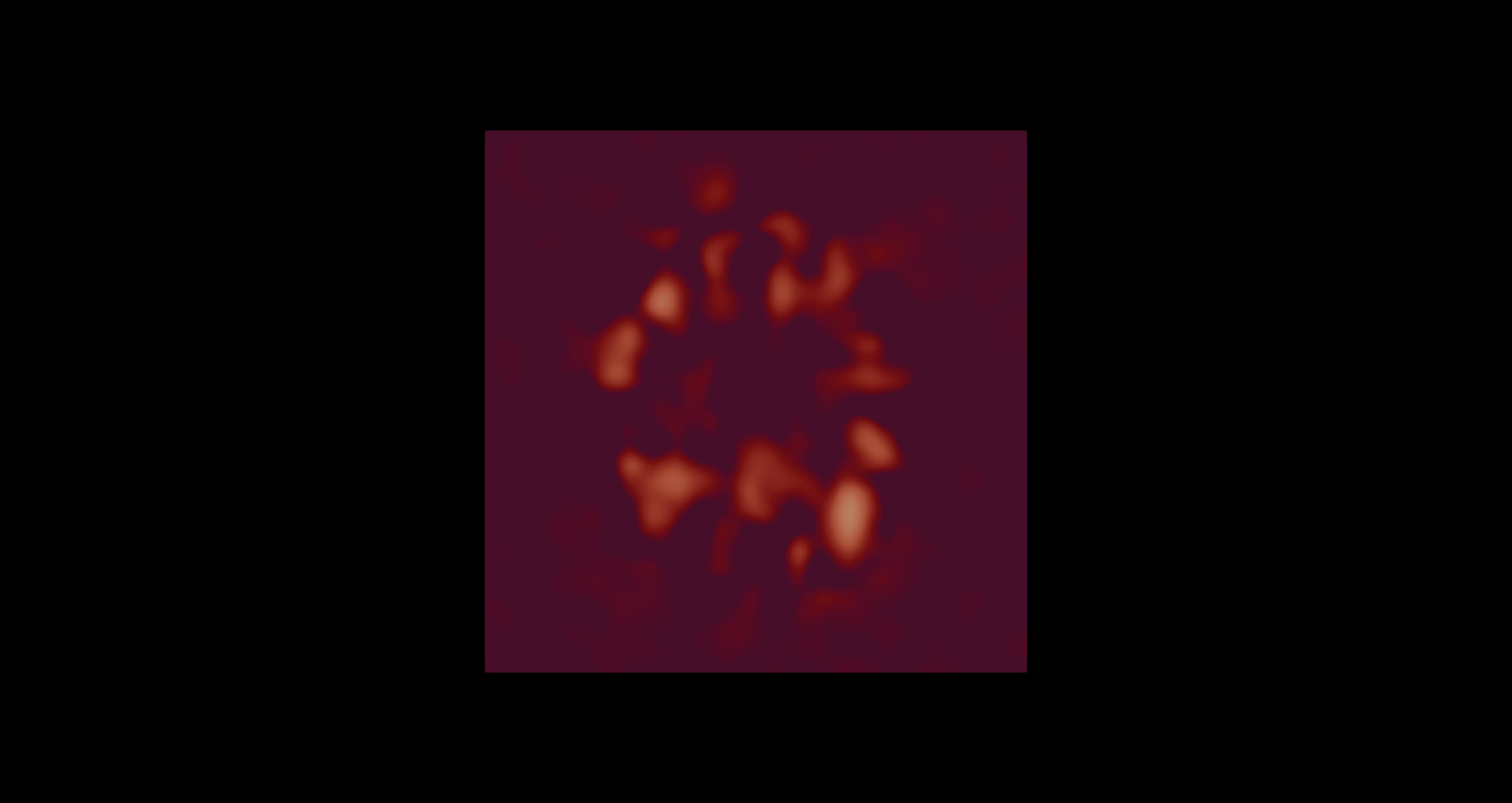}}
    \end{minipage}
    \begin{minipage}[c]{0.1\textwidth}
        \centering
        \subcaptionbox*{\label{Jac2-colorbar}}[\linewidth]{\includegraphics[trim={38cm 4cm 14cm 4cm},clip,height=3cm]{figures/3/colorbar_det_J.png}}
    \end{minipage}
    
    \caption[Jacobian determinant]{Uncertainty and the Jacobian determinant of transformations sampled from the models using \glspl{SVF} based on cubic B-splines. The transformations are visibly smoother than in case of non-parametric \glspl{SVF}. The uncertainty estimates are also shown to be highly dependent on the transformation parametrisation. The figure shows the middle axial slice of 3D images.}
    \label{fig:SVFFD}
\end{figure}

In \Cref{fig:SVFFD}, we show the output uncertainties and sample Jacobian determinants. Using \glspl{SVF} based on B-splines results in smoother transformations. In fact, the implicit smoothness of B-splines helps both with the crude assumption of a diagonal covariance matrix in \gls{VI}, which translates to a diagonal pre-conditioning matrix in \gls{SGLD}, and with computational and memory efficiency. We also observed a positive impact on the image registration accuracy. For both control point spacings, \glspl{SVF} based on B-splines are more accurate than non-parametric \glspl{SVF} on the majority of structures. Despite comparable accuracy, the uncertainty estimates differ.

%---------------------------------------------------------------
\section{Discussion}

%---------------------------------------------------------------
\subsection{Modelling assumptions}
\label{sec:modelling_assumptions}

%<*discussion>
To draw samples from the true posterior of transformation parameters remains a difficult problem even with a large number of simplifying model assumptions. If the true posterior were approximately Gaussian, \gls{VI} would provide a good approximation thereof and the lower uncertainty output by \gls{SG-MCMC} would indicate that the output samples are autocorrelated even with subsampling. However, if the true posterior is not Gaussian, then the posterior output by \gls{VI} is ill-fitting, and the samples output by \gls{SG-MCMC} cover multiple modes near the mode that corresponds to \gls{VI}, which makes a good case for the use of \gls{SGLD}. %</discussion>

In practice, the quality of uncertainty estimates is also sensitive to the validity of model assumptions. These include coinciding image intensities up to the expected spatial noise offsets and ignoring spatial correlations between residuals. The first assumption is valid in case of mono-modal registration but the model can be easily adapted to other settings through use of a different data loss. We manage the second assumption by use of virtual decimation \citep{Groves2011, Simpson2012}, which calculates the effective degrees of freedom of the residual map with stationary covariance by proxy of the correlation between neighbouring voxels in each direction.

We showed how good modelling choices, such as a careful choice of priors and transformation parametrisation can mitigate some of the issues caused by approximations necessary in practical applications. The main problem in inter-subject brain \gls{MRI} registration remains accuracy but the trade-off between the quality of the transformation and registration accuracy can be managed effectively.

%---------------------------------------------------------------
\subsection{Runtime}

The experiments were run on a system with an Intel i9-10920X CPU and a GeForce RTX 3090 GPU. \Cref{tab:efficiency} shows the runtime of the models. \gls{VI} takes approximately \SI{3}{\minute} to register a pair of images and produces \numprint{140} samples/sec, while \gls{SG-MCMC} produces \numprint{25} autocorrelated samples/sec. %<*speed>
The registration time for \gls{SG-MCMC} is absent because it is difficult to pin down the mixing time, so a fair comparison of the relative efficiency of \gls{VI} and \gls{SG-MCMC} can be done only on the basis of the number of samples per second.
%</speed>

Due to lack of publicly available information we cannot directly compare the efficiency of our model to other Bayesian image registration methods. The speed of our model is an order of magnitude better than reported by \cite{LeFolgoc2017}, while also being three- rather than two-dimensional. Thus, the proposed method based on \gls{SG-MCMC} is very efficient given the Bayesian constraint. It is not as efficient as feed-forward neural networks, such as VoxelMorph. However, the proposed model is fully Bayesian and enables asymptotically exact sampling from the posterior of transformation parameters.

\begin{table}[!htb]
    \centering
    \caption{Comparison of \gls{VI}, \gls{SG-MCMC}, and VoxelMorph vis-à-vis computational efficiency. %<*correlation>
    In contrast to \gls{VI} and VoxelMorph, \gls{SG-MCMC} requires burn-in and produces samples which need to be further subsampled in order to avoid autocorrelation. For this reason, in case of \gls{SG-MCMC} we report the sampling speed based on the time needed to draw \numprint{4000} samples. Note that the ratio used to subsample the output of \gls{SG-MCMC} may vary depending on the application (cf. \Cref{sec:SG-MCMC}).
    %</correlation>
    }
    \label{tab:efficiency}
    
    \begin{tabular}{lrrS[table-format=1.1e1]}
        \toprule
        \textbf{model}     & training time     & registration time      & {samples/sec}   \\
        \midrule
        \gls{VI}            & ---               & \SI{3}{\minute}       & \num{1.4e2}     \\
        \gls{SG-MCMC}       & ---               & ---                   & {$< 1.0$}       \\
        VoxelMorph          & \SI{38}{\hour}    & \SI{55}{\ms}          & \num{2.6e1}     \\
        \bottomrule
    \end{tabular}
\end{table}

%---------------------------------------------------------------
\section{Conclusion}

In this paper we presented a new Bayesian model for three-dimensional medical image registration. The proposed regularisation loss allows to adjust regularisation strength to the data when using an \gls{SVF} transformation parametrisation, which involves a very large number of degrees of freedom. Sampling from the posterior distribution via \gls{SG-MCMC} makes it possible to quantify registration uncertainty even for large images. The computational efficiency and theoretical guarantees regarding samples output by \gls{SG-MCMC} make our model an attractive alternative for uncertainty quantification in non-rigid image registration compared to methods based on \gls{VI}.

%---------------------------------------------------------------
\acks{This research used UK Biobank resources under the application number 12579. Daniel Grzech is funded by the EPSRC CDT for Medical Imaging EP/L015226/1 and Lo\"{i}c Le Folgoc by EP/P023509/1. We are very grateful to Prof. Wells and the Reviewers for their feedback during the review process.}

%---------------------------------------------------------------
\ethics{The work follows appropriate ethical standards in conducting research and writing the manuscript, following all applicable laws and regulations regarding treatment of animals or human subjects.}

%---------------------------------------------------------------
\coi{We declare we do not have conflicts of interest.}

%---------------------------------------------------------------
\bibliography{references}

%---------------------------------------------------------------
\newpage
\appendix
\setcounter{section}{0}
\renewcommand{\thesection}{\Alph{section}}

\section{}
\label{app:reg}

In this appendix we show how to calculate the mean of the logarithm of the gamma distribution \citep{whuber2018}.

Let $X$ be a random variable which follows the gamma distribution with the shape and rate parameters $\alpha$ and $\beta$, i.e. $X \sim \Gamma \left( \alpha, \beta \right)$. We are interested in the expected value of $Y = \log X $. If we assume that $\beta = 1$, then the \gls{PDF} of $X$ is given by:
\begin{equation}
    f_X \left( x \right) = \frac{1}{\Gamma \left( \alpha \right)} x^\alpha e^{-x} \frac{\dif x}{x}
\end{equation}
Note that $\Gamma \left( \alpha \right)$ is a constant and the integral of $f_X$ must equal $1$, so we have:
\begin{equation}
    \Gamma \left( \alpha \right) = \int_{\mathbb{R}_+} x^\alpha e^{-x} \frac{\dif x}{x}
\end{equation}

Let $x = \exp y$. This means that $\frac{\dif x}{x} = \dif y$, so the \gls{PDF} of $Y$ is given by:
\begin{equation}
    f_Y \left( y \right) = \frac{1}{\Gamma\left( \alpha \right)} e^{\alpha y - e^y} \dif y
\end{equation}
Again, because $\Gamma \left( \alpha \right)$ is a constant and the integral of the \gls{PDF} of $Y$ must equal $1$, we have:
\begin{equation}
    \Gamma \left( \alpha \right) = \int_\mathbb{R} e^{\alpha y - e^y} \dif y
\end{equation}

Now, using Feynman's trick of differentiating under the integral sign, we see that:
\begin{alignat}{6}
    \mathbb{E} \left( Y \right) &= &&\int_\mathbb{R} y f_Y \left( y \right) &&\dif y &&= \frac{1}{\Gamma \left( \alpha \right)} &&\int_\mathbb{R} \Gamma \left( \alpha \right) y f_Y \left( y \right) &&\dif y \\
    &= \frac{1}{\Gamma \left( \alpha \right)} &&\int_\mathbb{R} \frac{d}{d\alpha} e^{\alpha y - e^{y}} &&\dif y &&= \frac{1}{\Gamma \left( \alpha \right)} \frac{d}{d\alpha} &&\int_\mathbb{R} e^{\alpha y - e^y} &&\dif y \\
    &= \frac{1}{\Gamma \left( \alpha \right)} &&\frac{d}{d\alpha} \Gamma \left( \alpha \right) && &&= \frac{d}{d\alpha} \log \Gamma \left( \alpha \right) \\
    &= \psi \left( \alpha \right)
\end{alignat}
where $\psi$ is the digamma function. Finally, the rate parameter $\beta$ shifts the logarithm by $-\log \beta$. Therefore, the expected value of $\log X$ is given by:
\begin{equation}
    \mathbb{E} \left[ \log X \right] = \psi \left( \alpha \right) - \log \beta
\end{equation}

\section{}
\label{app:KL}

In this appendix we show how to calculate the \gls{KL-divergence} term between the approximate variational posterior $q \left( w \right) \sim \mathcal{N} \left( \mu_w, \Sigma_w \right)$ and the prior $p \left( w \right)$, as well as the entropy $H(q)$ of the approximate variational posterior, both of which are needed to maximise the \gls{ELBO} in \Cref{eq:ELBO}.

The \gls{KL-divergence} can be evaluated in terms of the entropy $H(q)$ of the approximate variatonal posterior and the regularisation energy $\mathcal{E}_{\text{reg}}$:
\begin{equation}
    \kld{q}{p} = \int_w q_w \left(w \right) \log q \left( w \right) \dif w - \int_w q \left(w \right) \log p \left( w \right) \dif w = -H(q) + \big \langle \mathcal{E}_{\text{reg}} \big \rangle_q
\end{equation}
where $\big \langle \cdot \big \rangle$ denotes the expected value.

The entropy $H(q)$ of the approximate variational posterior is calculated as follows:
\begin{alignat}{2}
    H(q) = -\int_w q \left( w \right) \log q \left( w \right) \dif w = \frac{1}{2} \log \det \left( \Sigma_w \right) + \frac{1}{2} \big \langle \left( w - \mu_w \right)^\intercal \Sigma^{-1}_w \left(w - \mu_w \right) \big \rangle_q + \text{const.}
\end{alignat}
The first term on the left on the RHS is calculated using the matrix determinant lemma:
\begin{equation}
    \det \left( \Sigma_w \right) = \det \left( \text{diag} \left( \sigma^2_w \right) + u_w u_w^\intercal \right) = \left( 1 + u_w^\intercal \text{diag} \left( \sigma_w^{-2} \right) u_w \right) \times \det \left( \text{diag} \left( \sigma^2_w \right) \right)
\end{equation}
To evaluate the other term, and in particular the precision matrix $\Sigma^{-1}_w$, we can use the Sherman-Morrison formula \citep{Sherman1950}, which states that:
\begin{equation}
    \Sigma^{-1}_w = \left( \text{diag} \left( \sigma^2_w \right) + u_w u_w^\intercal \right)^{-1} = \text{diag} \left( \sigma^{-2}_w \right) - \frac{\text{diag} \left( \sigma^{-2}_w \right) u_w u_w^\intercal \text{diag} \left( \sigma^{-2}_w \right)}{1 + u_w^\intercal \text{diag} \left( \sigma^{-2}_w \right) u_w}
\end{equation}

\end{document}